\documentclass{article}

\PassOptionsToPackage{numbers, compress}{natbib}

     \usepackage[final]{neurips_2022}

\usepackage[utf8]{inputenc} 
\usepackage[T1]{fontenc}    
\usepackage{hyperref}       
\usepackage{url}            
\usepackage{booktabs}       
\usepackage{amsfonts}       
\usepackage{nicefrac}       
\usepackage{microtype}      
\usepackage{xcolor}         

\usepackage{graphicx}
\usepackage{subfig}
\usepackage{wrapfig}

\usepackage{multirow}
\usepackage{float}
\usepackage{bm}

\usepackage{algorithm}
\usepackage{algorithmic}

\usepackage{amsmath}

\newtheorem{definition}{Definition}

\allowdisplaybreaks[4]
\usepackage{amssymb}
\usepackage{mathtools}

\usepackage[para]{footmisc}

\definecolor{amber(sae/ece)}{rgb}{1.0, 0.49, 0.0}
\definecolor{cadmiumorange}{rgb}{0.93, 0.53, 0.18}
\definecolor{azure(colorwheel)}{rgb}{0.0, 0.5, 1.0}

\title{Pareto Set Learning for \\ Expensive Multi-Objective Optimization}

\DeclareMathOperator*{\argmin}{arg\,min}

\def\ddefloop#1{\ifx\ddefloop#1\else\ddef{#1}\expandafter\ddefloop\fi}

\def\ddef#1{\expandafter\def\csname v#1\endcsname{\ensuremath{\boldsymbol{#1}}}}
\ddefloop ABCDEFGHIJKLMNOPQRSTUVWXYZabcdefghijklmnopqrstuvwxyz\ddefloop

\def\ddef#1{\expandafter\def\csname v#1\endcsname{\ensuremath{\boldsymbol{\csname #1\endcsname}}}}
\ddefloop {alpha}{beta}{gamma}{delta}{epsilon}{varepsilon}{zeta}{eta}{theta}{vartheta}{iota}{kappa}{lambda}{mu}{nu}{xi}{pi}{varpi}{rho}{varrho}{sigma}{varsigma}{tau}{upsilon}{phi}{varphi}{chi}{psi}{omega}{Gamma}{Delta}{Theta}{Lambda}{Xi}{Pi}{Sigma}{varSigma}{Upsilon}{Phi}{Psi}{Omega}{ell}\ddefloop

\def\ddef#1{\expandafter\def\csname bb#1\endcsname{\ensuremath{\mathbb{#1}}}}
\ddefloop ABCDEFGHIJKLMNOPQRSTUVWXYZ\ddefloop

\author{
  Xi Lin, Zhiyuan Yang, Xiaoyuan Zhang, Qingfu Zhang \\
  Department of Computer Science, City University of Hong Kong\\
  \texttt{\{xi.lin, zhiyuyang4-c, xzhang2523-c\}@my.cityu.edu.hk, qingfu.zhang@cityu.edu.hk} \\
}

\begin{document}

\maketitle

\begin{abstract}

Expensive multi-objective optimization problems can be found in many real-world applications, where their objective function evaluations involve expensive computations or physical experiments. It is desirable to obtain an approximate Pareto front with a limited evaluation budget. Multi-objective Bayesian optimization (MOBO) has been widely used for finding a finite set of Pareto optimal solutions. However, it is well-known that the whole Pareto set is on a continuous manifold and can contain infinite solutions. The structural properties of the Pareto set are not well exploited in existing MOBO methods, and the finite-set approximation may not contain the most preferred solution(s) for decision-makers. This paper develops a novel learning-based method to approximate the whole Pareto set for MOBO, which generalizes the decomposition-based multi-objective optimization algorithm (MOEA/D) from finite populations to models. We design a simple and powerful acquisition search method based on the learned Pareto set, which naturally supports batch evaluation. In addition, with our proposed model, decision-makers can readily explore any trade-off area in the approximate Pareto set for flexible decision-making. This work represents the first attempt to model the Pareto set for expensive multi-objective optimization. Experimental results on different synthetic and real-world problems demonstrate the effectiveness of our proposed method. 

\end{abstract}

\section{Introduction}
\label{sec_introduction}

\begin{wrapfigure}{R}{0.45\linewidth}
    \vspace{-0.2in}
    \centering
    \subfloat[ParEGO\label{fig_pml_3objs_a}]{\includegraphics[width = 0.5\linewidth]{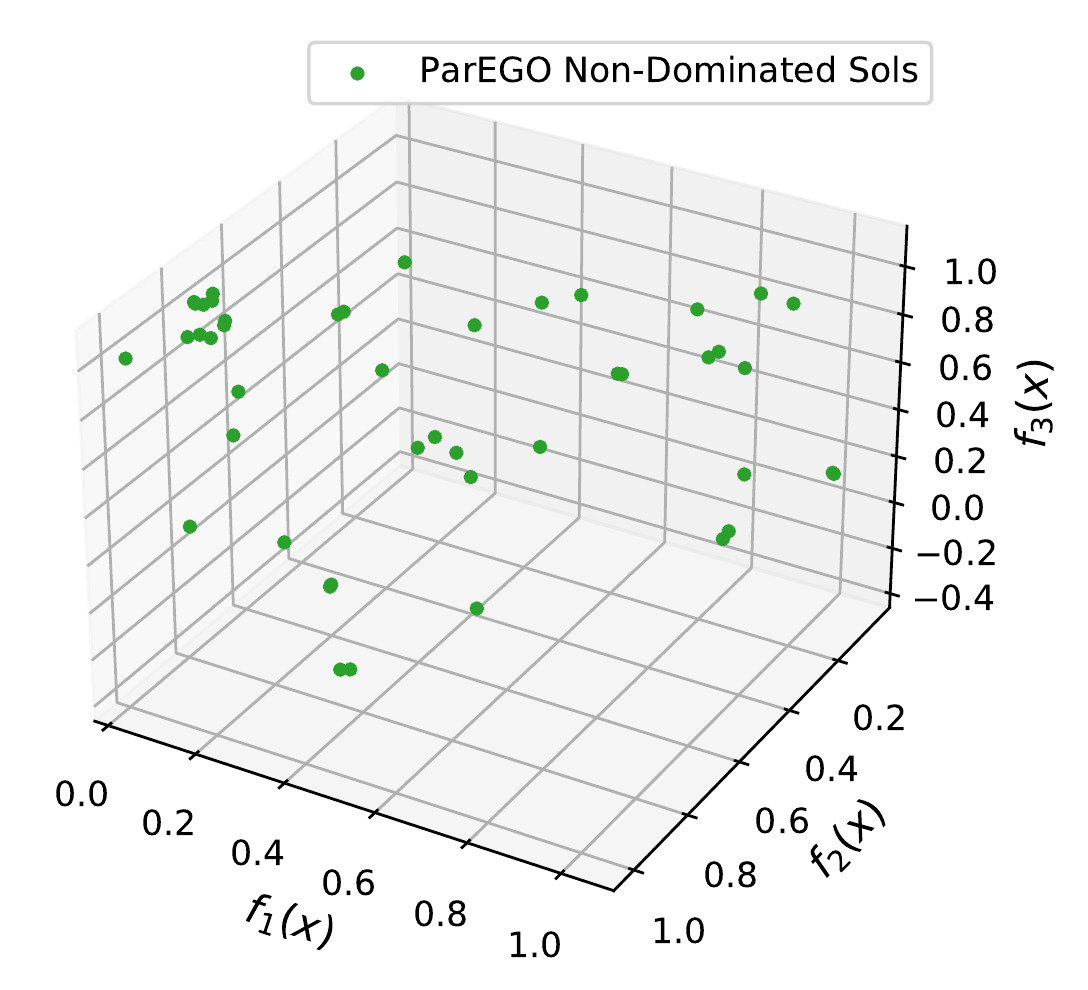}}
    \hfill
    \subfloat[Pareto Set Learning\label{fig_pml_3objs_b}]{\includegraphics[width = 0.5\linewidth]{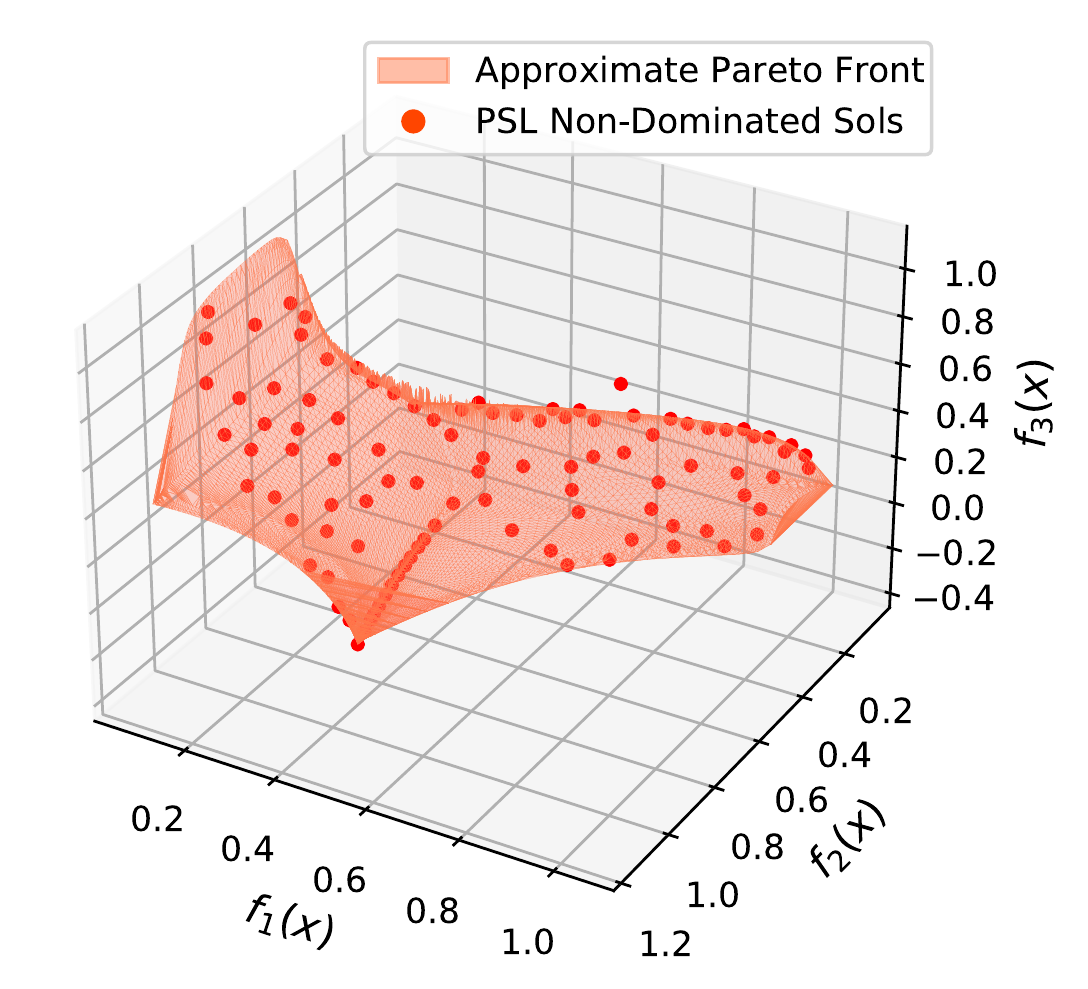}}
    \caption{\textbf{Pareto Set Learning} can approximate the whole Pareto set, and let decision-makers easily explore any trade-off among objectives to choose their preferred solutions.}
    \label{fig_pml_3objs}
    \vspace{-0.1in}
\end{wrapfigure}

Many real-world applications involve optimizing multiple costly-to-evaluate and potentially competing objectives, such as finding strong yet ductile material~\citep{jablonka2021bias}, building a neural network with high accuracy and low latency~\citep{eriksson2021latency}, and improving the quality while minimizing total charge in particle accelerator tuning~\citep{roussel2021multiobjective}. Very often, these objectives conflict each other and cannot be optimized simultaneously by a single solution. Instead, there is a set of solutions with different optimal trade-offs among the objectives, called the Pareto set. For a Pareto optimal solution, none of its objective values can be further improved without deteriorating others. In addition, the evaluation of each solution could require time-consuming computation or costly physical experiments, and thus a large number of evaluations are unbearable. Different multi-objective Bayesian optimization (MOBO) algorithms~\citep{khan2002multi,laumanns2002bayesian,keane2006statistical}, typically directly generalized from the single-objective Bayesian optimization (BO)~\citep{mockus1989bayesian, jones2001taxonomy, brochu2010tutorial, shahriari2016taking, frazier2018tutorial}, have been proposed to find a small set of approximate Pareto solutions with a small amount of objective function evaluation budget. 

The finite set approximation has some undesirable drawbacks. For a nontrivial multi-objective optimization problem, the Pareto set is on a continuous manifold and has infinite solutions with different optimal trade-offs among the objectives~\citep{miettinen1998nonlinear}. This Pareto set structure could be helpful to better select candidate solutions for expensive evaluation, and hence accelerate the optimization process of MOBO. In addition, a small set of solutions may not contain the one(s) that exactly satisfy the decision-maker's preferences. Finding the most preferred trade-off solution(s) could require several rounds of interaction with the decision-makers. These approaches would be extremely time-consuming, especially when the optimization modeler and the decision-maker are not in the same team, which is common in real-world MOBO applications~\citep{malkomes2021beyond}.

This paper proposes a novel Pareto set learning (PSL) method to approximate the whole Pareto set for expensive multi-objective optimization problems with a limited evaluation budget. Our proposed method can accelerate the multi-objective Bayesian optimization process, and also provide decision-makers with more useful information to support flexible decision-making. To the best of our knowledge, this is the first attempt to learn the whole Pareto set for expensive multi-objective optimization. Our main contributions include:

\begin{itemize}
    \item We propose a novel set model to map any trade-off preference to its corresponding Pareto solution, along with a surrogate model-based method to approximate the whole Pareto set with a limited evaluation budget.
    \item We develop a lightweight yet powerful batch acquisition search method for efficient MOBO, which can outperform other MOBO approaches in terms of both performance and computational cost. We demonstrate that the learned Pareto set can support flexible user-involved decision-making.  
    \item We test our proposed method on both synthetic benchmarks and real-world application problems. The results validate the efficiency and usefulness of PSL.
\end{itemize}

\section{Related work}
\label{sec_background}

\textbf{Bayesian Optimization.} Surrogate model-based methods have been widely used and studied for expensive optimization~\citep{kushner1964new, jones1998efficient, pelikan1999boa, snoek2012practical}. These methods iteratively build a surrogate model to approximate the black-box objective function, and uses an acquisition function to search for the optimal solution. Much effort has been made on various design issues in Bayesian optimization, such as acquisition functions~\citep{srinivas2009gaussian}, high-dimensional optimization~\citep{wang2013bayesian, wang2018batched}, batch evaluation~\citep{desautels2014parallelizing}, scalable optimization~\citep{snoek2015scalable, eriksson2019scalable}, and theoretical analysis~\citep{kawaguchi2015bayesian}. Most work for BO are on single-objective optimization. We refer readers to \citet{garnett2022bayesian} for a comprehensive introduction.  

\textbf{Multi-Objective Bayesian Optimization.} MOBO extends single-objective Bayesian optimization for solving expensive multi-objective optimization problems. Although the Pareto set could contain infinite solutions, the MOBO methods typically focus on finding a single or a finite set of solutions. The scalarization-based algorithms, such as ParEGO~\citep{knowles2006parego} and TS-TCH~\citep{paria2020flexible} iteratively scalarize the multi-objective problem into single-objective ones with random preferences, and then apply single-objective BO to solve them. MOEA/D-EGO~\citep{zhang2010expensive} adopts the MOEA/D framework~\cite{zhang2007moea} to solve a set of surrogate scalarized subproblems simultaneously. SMS-EGO~\citep{ponweiser2008multiobjective} and PAL~\citep{zuluaga2013active, zuluaga2016varepsilon} generalize the upper confidence bound (UCB) to multi-objective optimization. \citet{emmerich2006single} and \citet{emmerich2008computation} propose the probability of improvement (PI) and expected improvement (EI) for multi-objective hypervolume. Entropy search methods~\citep{hennig2012entropy,hernandez2014predictive, hoffman2015output, wang2017max} have also been studied in multi-objective optimization~\citep{hernandez2016predictive,belakaria2019max,suzuki2020multi}. \citet{bradford2018efficient} and \citet{belakaria2020uncertainty} consider Thompson sampling and uncertainty maximization for multi-objective optimization, respectively. Different algorithms can be hybridized to achieve better performances~\cite{steponavivce2017dynamic}.

Different new developments have been recently proposed to handle the issues of diverse batch evaluation~\citep{lukovic2020diversity}, efficient hypervolume improvement calculation~\citep{daulton2020differentiable}, noisy optimization~\citep{daulton2021parallel,daulton2022robust}, high-dimensional optimization~\cite{daulton2022multi}, and decision criteria beyond Pareto optimality~\citep{malkomes2021beyond}. Some attempts have been made to incorporate the decision-maker's preference into MOBO~\citep{abdolshah2019multi, paria2020flexible, astudillo2020multi}. They typically need the decision-maker's preference before or during optimization, which may not always be available in real-world applications. All these MOBO methods aim to provide a finite set of approximate Pareto optimal solutions to decision-makers.

\textbf{Structure Learning.} In addition to the surrogate objective model, some methods have been proposed to explore the problem structure for Bayesian optimization. \citet{sener2020learning} learn the geometric structure of the problem in an online manner to accelerate optimization. \citet{wang2020learning} and \citet{zhao2022multi} apply Monte Carlo tree search (MCTS) to divide the search space for efficient modeling and searching. Novel latent space modelings~\citep{gomez2018automatic, tripp2020sample} have been proposed for optimization problems with complex solution representations. 

\textbf{From Population to Pareto Set Learning.} For the last several decades, most multi-objective optimization methods have focused on finding a single or a finite set of Pareto optimal solutions (e.g., a population) to approximate the Pareto set~\citep{miettinen1998nonlinear,ehrgott2005multicriteria}. A few attempts have been made to approximate the whole Pareto set with simple mathematical models~\citep{rakowska1991tracing,hillermeier2001generalized,zhang2008rm,giagkiozis2014pareto}. \citet{pirotta2015multi} and \citet{parisi2016multi} have proposed to conduct Pareto manifold approximation for multi-objective reinforcement learning. Recently, different approaches have also been investigated to incorporate the preference information into deep neural networks for image style transfer~\citep{shoshan2019dynamic,dosovitskiy2019you}, multi-task learning with finite solutions~\cite{sener2018multi,lin2019pareto,mahapatramulti2020multi,ma2020efficient} or approximate Pareto front~\citep{lin2020controllable, navon2020learning,ruchte2021scalable}, reinforcement learning~\cite{yang2019generalized,abdolmaleki2020distributional,abdolmaleki2021multi}, and neural combinatorial optimization~\cite{lin2022pareto}. In this work, we generalize the decomposition-based multi-objective optimization algorithm (MOEA/D)~\citep{zhang2007moea}, and propose to learn a set model which maps all valid trade-off preferences to the Pareto set for efficient MOBO.

\begin{figure}[t]
\centering
\subfloat[(Weakly) Pareto Sols.]{\includegraphics[width = 0.25\linewidth]{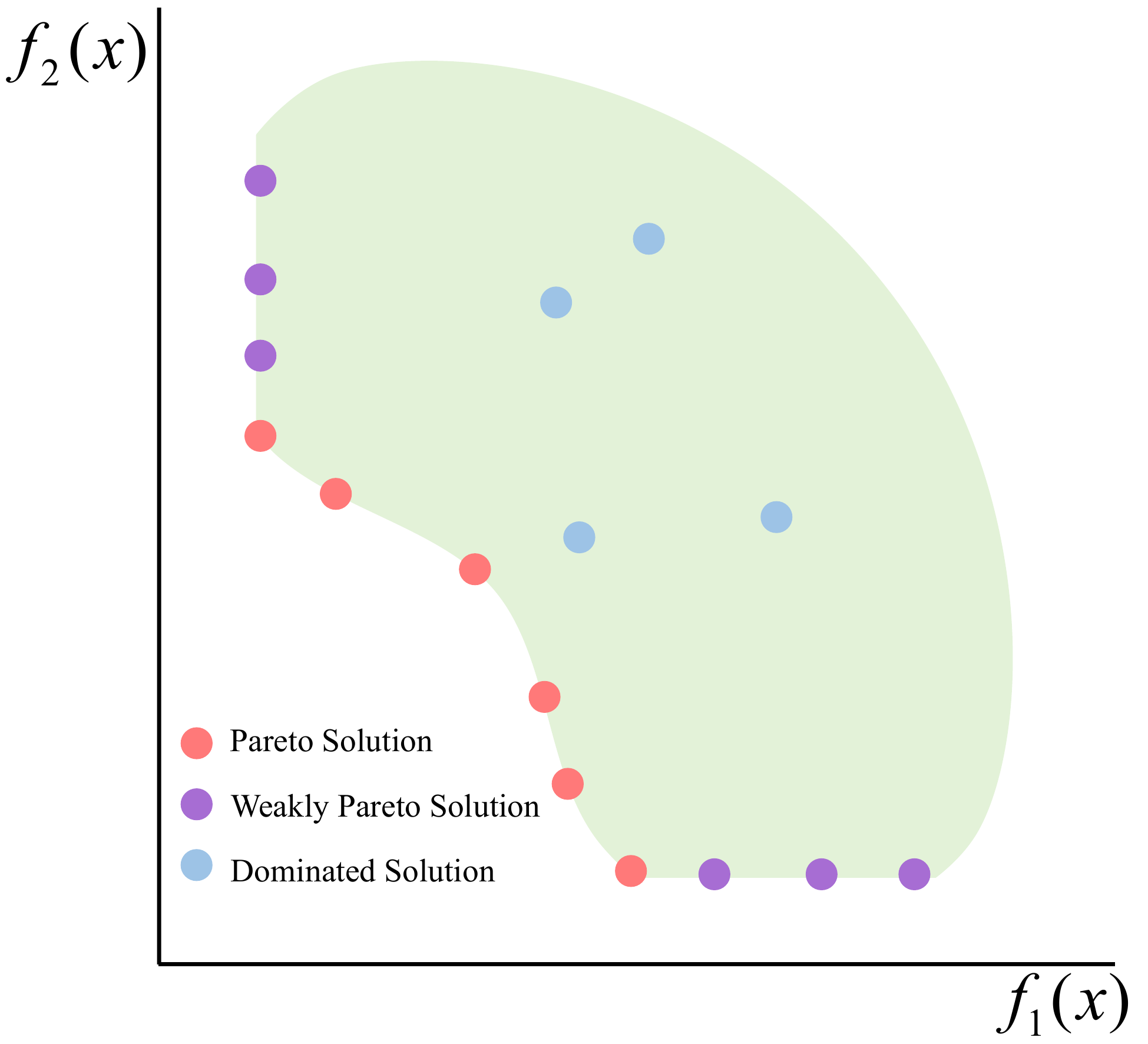}}\hfill
\subfloat[Weakly Pareto Front]{\includegraphics[width = 0.25\linewidth]{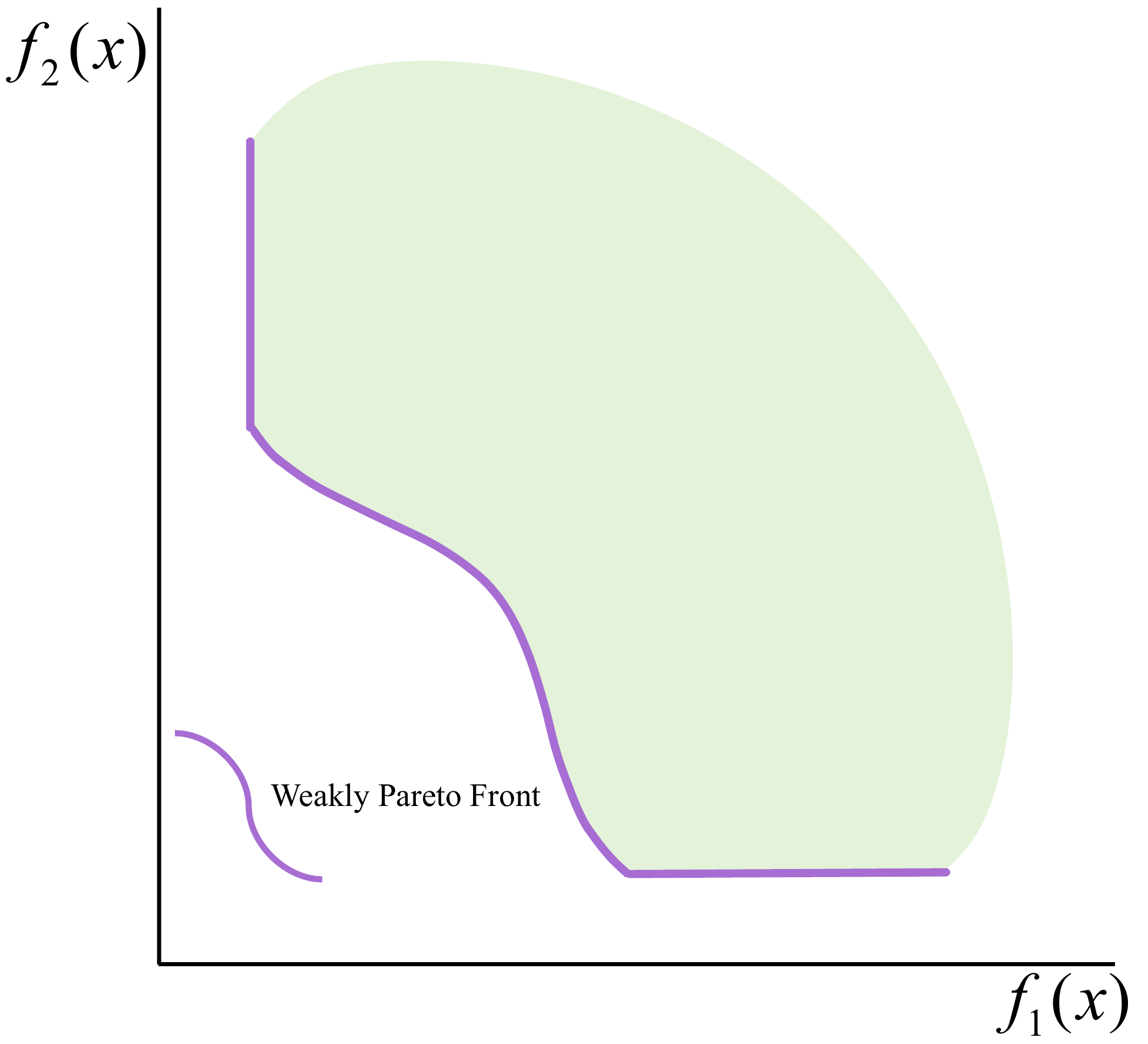}}\hfill
\subfloat[Pareto Front]{\includegraphics[width = 0.25\linewidth]{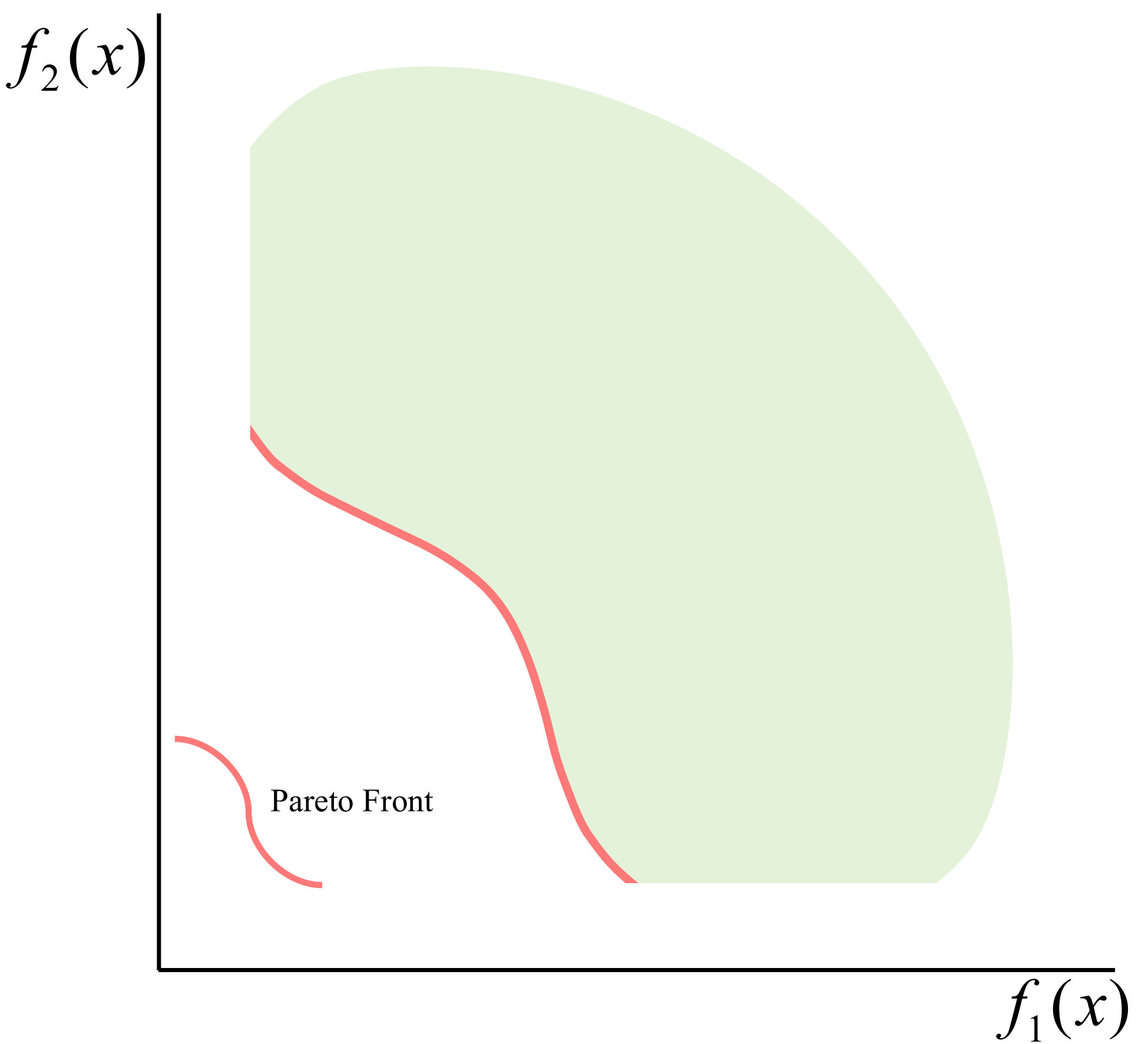}}\hfill
\subfloat[PSL for MOBO]{\includegraphics[width = 0.25\linewidth]{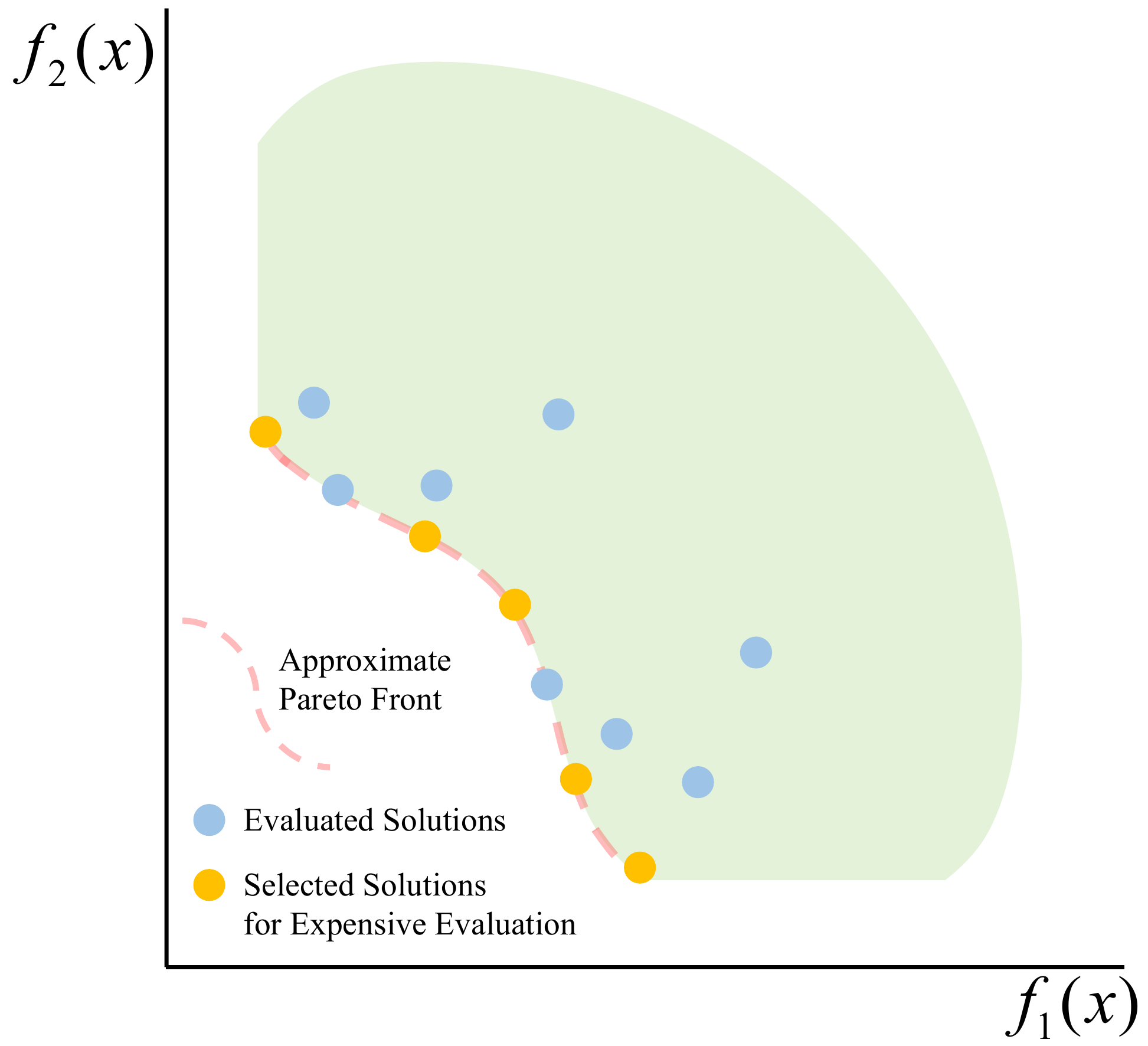}}
\caption{\textbf{(Weakly) Pareto solutions and (Weakly) Pareto front.} \textbf{(a)} Examples of Pareto solutions, weakly Pareto solutions, and dominated solutions. The Pareto solutions are also weakly Pareto optimal. The weakly Pareto optimal but not Pareto optimal solutions (e.g., purple points) are dominated but not strictly dominated by at least one Pareto solution. \textbf{(b)} The weakly Pareto front $\vf(\mathcal{M}_{\text{weak}})$ is the image of all weakly Pareto optimal solutions $\mathcal{M}_{\text{weak}}$ in the objective space. \textbf{(c)} The Pareto front $\vf(\mathcal{M}_{\text{ps}})$ is the image of all Pareto optimal solution $\mathcal{M}_{\text{ps}}$ (Pareto set) in the objective space. \textbf{(d)} Our proposed method approximates the whole Pareto set and uses it to select solutions for expensive evaluation, which can improve the search efficiency of multi-objective Bayesian optimization.}
\label{fig_weakly_pareto_front}
\vspace{-0.2in}
\end{figure}

\section{Expensive multi-objective optimization}
\label{sec_problem}

We consider the following expensive continuous multi-objective optimization problem:
\begin{align}
\min_{\vx \in \mathcal{X}} \ \vf(\vx) = (f_1(\vx),f_2(\vx),\cdots, f_m(\vx)),
\label{eq_mop}
\end{align}
where $\vx$ is a solution in the decision space $\mathcal{X} \subset \bbR^n$, $\vf: \mathcal{X} \rightarrow \bbR^m$ is an $m$-dimensional vector-valued objective function, and the evaluation is expensive for all individual objectives $f_i(\vx)$, $i=1,\ldots, m$. For a non-trivial problem, no single solution can optimize all objectives at the same time, and we have to make a trade-off among them. We have the following definitions for multi-objective optimization:  

\begin{definition}[Pareto Dominance and Strict Dominance]
\textit{Let $\vx^{a},\vx^{b} \in \mathcal{X}$, $\vx^{a}$ is said to dominate $\vx^{b}$, denoted as $\vx^{a} \prec \vx^{b}$, if and only if $f_i(\vx^{a}) \leq f_i(\vx^b), \forall i \in \{1,...,m\}$ and $\exists j \in \{1,...,m\}$ such that $f_j(\vx^{a}) < f_j(\vx^{(b)})$. In addition, $\vx^{a}$ is said to strictly dominate $\vx^{b}$ ($\vx^{a} \prec_{\text{strict}} \vx^{b}$), if and only if $f_i(\vx^{a}) < f_i(\vx^b), \forall i \in \{1,...,m\}$.}
\end{definition}

\begin{definition}[Pareto Optimality]
\textit{A solution $\vx^{\ast} \in \mathcal{X}$ is Pareto optimal if there is no $\hat \vx \in \mathcal{X}$ such that $\hat \vx \prec \vx^{\ast}$. A solution $\vx^{\prime} \in \mathcal{X}$ is weakly Pareto optimal if there is no $\hat \vx \in \mathcal{X}$ such that $\hat \vx \prec_{\text{strict}} \vx^{\prime}$.}
\end{definition}

\begin{definition}[Pareto Set/Front]
\textit{The set of all Pareto optimal solutions $\mathcal{M}_{\text{ps}} \subseteq \mathcal{X}$ is called the Pareto set, and its image in the objective space $\vf(\mathcal{M}_{\text{ps}}) = \{\vf(\vx)|\vx \in \mathcal{M}_{\text{ps}}\}$ is called the Pareto front. Similarly, we can define the weakly Pareto set $\mathcal{M}_{\text{weak}}$ and weakly Pareto front $\vf(\mathcal{M}_{\text{weak}})$.}
\end{definition}

The strict dominance relation is stronger than the Pareto dominance since it requires strictly better values for all objectives. Therefore, the set of weakly Pareto optimal solutions $\mathcal{M}_{\text{weak}}$ (e.g., the solutions that are \emph{not} strictly dominated) would be larger than $\mathcal{M}_{\text{ps}}$, and it is straightforward to see $\mathcal{M}_{\text{ps}} \subseteq \mathcal{M}_{\text{weak}}$. The illustration of (weakly) Pareto solution and Pareto front is shown in Figure~\ref{fig_weakly_pareto_front}. 

Each Pareto solution $\vx \in \mathcal{M}_{\text{ps}}$ represents a different optimal trade-off among the objectives for problem~(\ref{eq_mop}). Under mild conditions, the Pareto set $\mathcal{M}_{\text{ps}}$ and Pareto front $\vf(\mathcal{M}_{\text{ps}})$ are both on $(m-1)$-dimensional manifold in the decision space $\mathcal{X} \in \bbR^n$ and objective space $\bbR^m$, respectively~\citep{hillermeier2001generalized, zhang2008rm}. The number of Pareto solutions could be infinite (i.e. $|\mathcal{M}_{\text{ps}}| = \infty$).

\textbf{Bayesian Optimization (BO)} is a model-based method for solving expensive black-box optimization problems. Given a set of already-evaluated solutions $\{\vX,\vy\}$, BO builds surrogate models (e.g., Gaussian process) for each objective, and defines acquisition function(s) to leverage the surrogate objective values for navigating the search space. Only promising solutions will be selected for expensive evaluation. We refer interesting reader to \cite{garnett2022bayesian} for a detailed introduction.

\begin{wrapfigure}{R}{0.5\linewidth}
    \centering
    \vspace{-0.15in}
    \subfloat[Preference to PS\label{model_and_connection_a}]{\includegraphics[width = 0.5\linewidth]{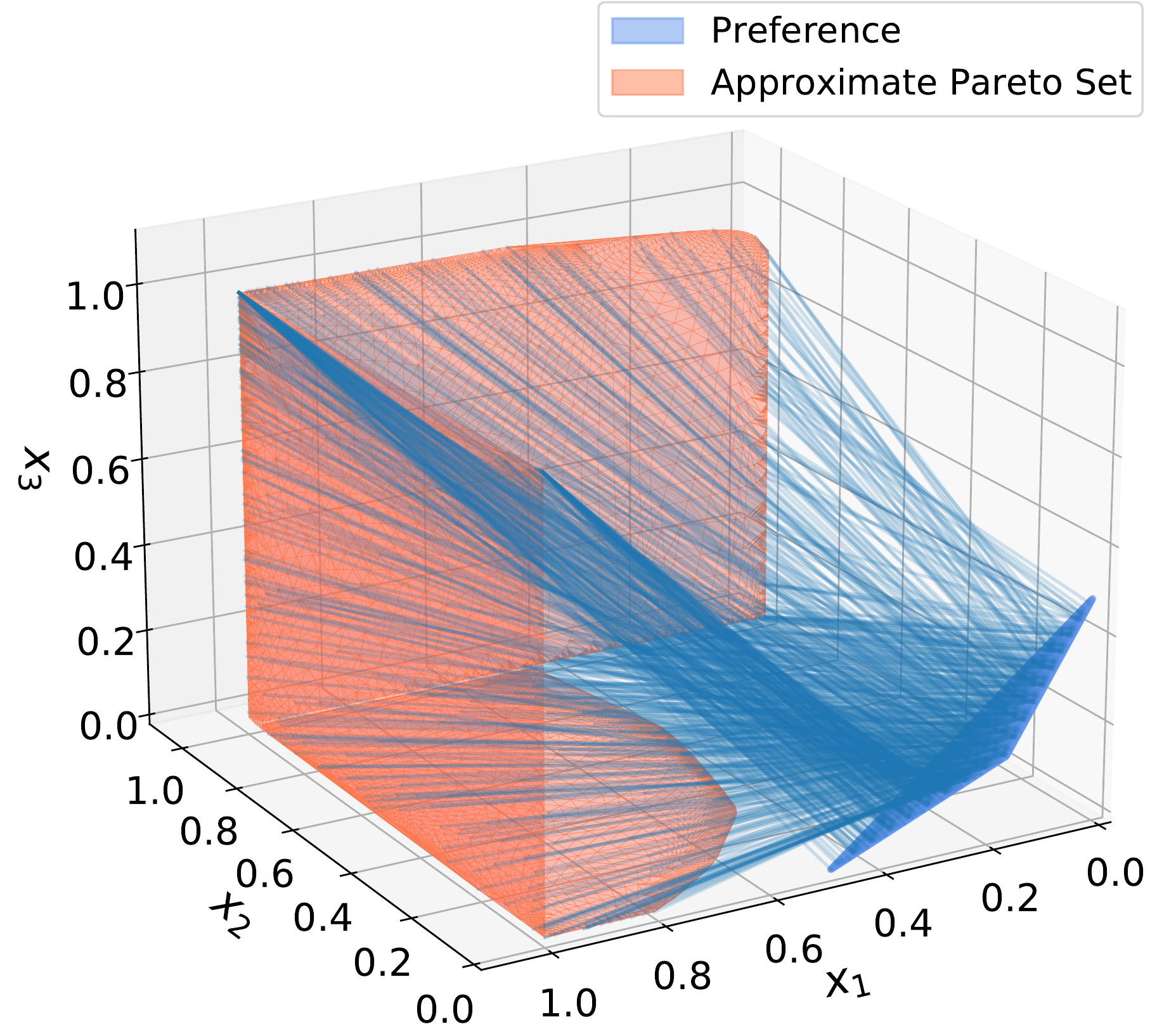}}
    \hfill
    \subfloat[Preference to PF\label{model_and_connection_c}]{\includegraphics[width = 0.5\linewidth]{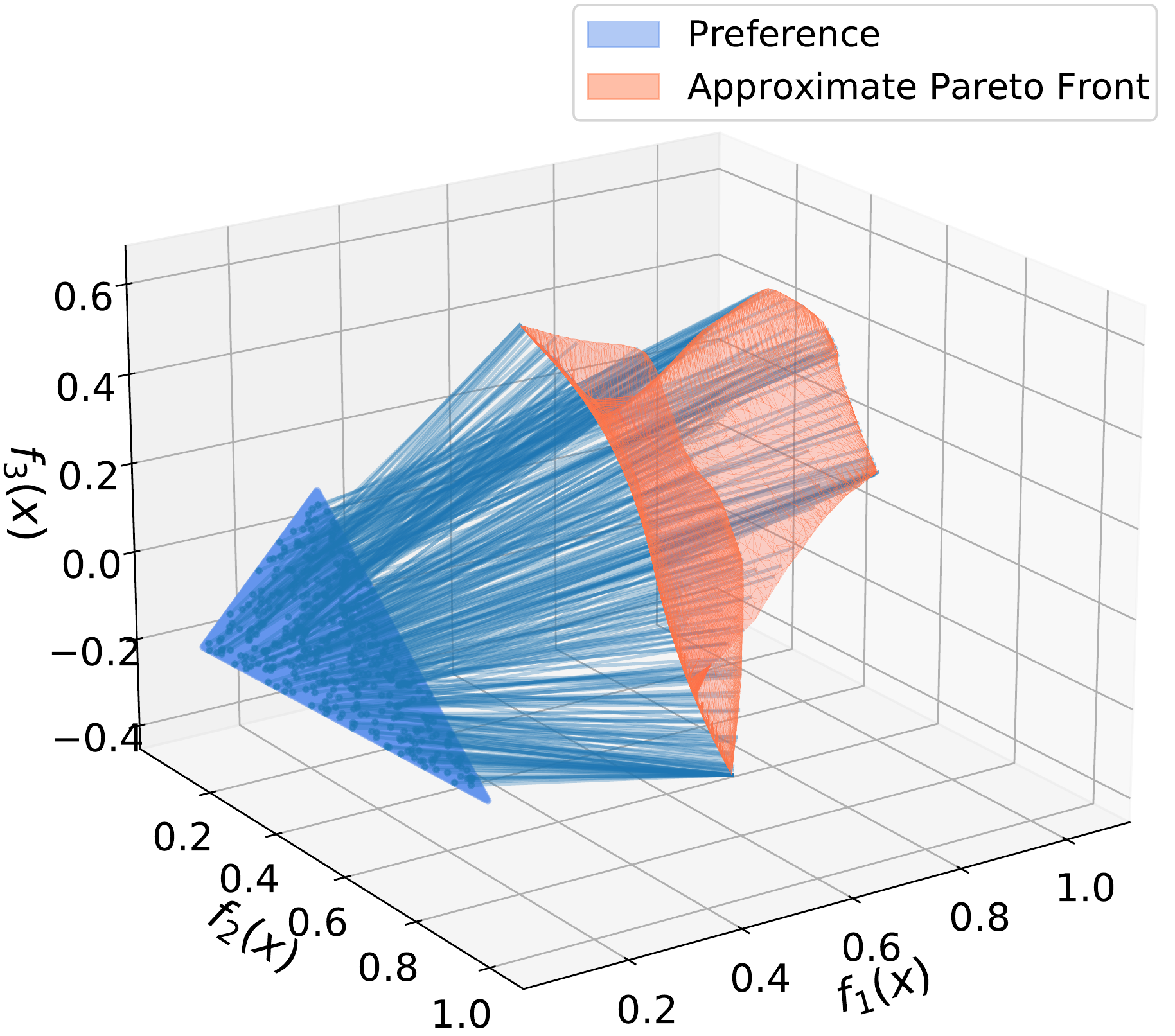}}
    \caption{\textbf{Mapping from Preferences to Approximate Pareto Set/Front:} Our proposed PSL method learns the connection from a set of valid (infinite) trade-off preference $\Lambda = \{\vlambda \in \bbR^m_{+}|\sum \lambda_i = 1\}$ to \textbf{(a)} the approximate Pareto set $\mathcal{M}_{\text{psl}}$ and hence \textbf{(b)} the corresponding predicted Pareto front $\hat \vf(\mathcal{M}_{\text{psl}})$. The whole Pareto set (front) can be easily explored by adjusting the preference. The preference simplex have been resized and rotated for better visualization.} 
    \label{model_and_connection}
    \vspace{-0.1in}
\end{wrapfigure}

\textbf{Pareto Set Learning.} The current MOBO methods aim to find a small set of finite solutions $\mathcal{S} = \{\bar \vx^{(1)},\bar \vx^{(2)},\cdots, \bar \vx^{(|\mathcal{S}|)}\}$ to approximate the Pareto set $\mathcal{M}_{\text{ps}}$. In addition to the evaluated solutions $\mathcal{S}$, our proposed Pareto set learning (PSL) method also learns an estimated Pareto set $\mathcal{M}_{\text{psl}}$ with the predicted Pareto front $\hat \vf(\mathcal{M}_{\text{psl}})$ to approximate the Pareto set $\mathcal{M}_{\text{ps}}$ and Pareto front $\vf(\mathcal{M}_{\text{ps}})$. The whole approximate Pareto set can be easily explored by adjusting the trade-off preference as illustrated in Figure~\ref{model_and_connection}. With the learned Pareto set, we also develop an efficient batched solution selection approach for efficient MOBO, which will be introduced in the next section.

\section{Pareto Set Learning for MOBO}
\label{sec_pmtl}

\subsection{Pareto set model}

As pointed out in Section~\ref{sec_problem}, the Pareto set can contain infinite solutions with different trade-offs. In addition, there is no complete order among the Pareto solutions. A Pareto set model for MOBO should be powerful enough to approximate the whole Pareto set, and convenient enough to easily explore any trade-off solutions. In this work, we propose to build a set model that maps any trade-off preferences to their corresponding Pareto solutions with scalarization. 

\textbf{Scalarization.}
The scalarization method provides a natural connection from a set of preferences $\Lambda = \{\vlambda \in \bbR^m_{+}|\sum \lambda_i = 1\}$ among the $m$ objectives to the Pareto set $\mathcal{M}_{\text{ps}}$. The most simple and straightforward approach is the weight-sum scalarization:
\begin{eqnarray}
\min_{\vx \in \mathcal{X}} g_{\text{ws}}(\vx|\vlambda) =  \min_{\vx \in \mathcal{X}} \textstyle\sum_{i=1}^{m}\lambda_i f_i(\vx).
\label{eq_ws_decomposition}
\end{eqnarray}
However, this method can only find the convex hull of Pareto front $\vf(\mathcal{M}_{\text{ps}})$~\citep{boyd2004convex, ehrgott2005multicriteria}. In this work, we use the following weighted Tchebycheff approach: 
\begin{eqnarray}
\min_{\vx \in \mathcal{X}} g_{\text{tch}}(\vx|\vlambda) =  \min_{\vx \in \mathcal{X}} \max_{1 \leq i \leq m} \{ \lambda_i (f_i(\vx) - (z^*_i - \varepsilon)) \},
\label{eq_tch_decomposition}
\end{eqnarray}
where $\vz^* = (z^*_1,\cdots, z^*_m)$ is the ideal vector for the objective vector $\vf(\vx)$ (i.e. lower-bound for minimization problem), $\varepsilon > 0$ is a small positive scalar, and $u_i = (z^*_i - \varepsilon)$ is called an (unachievable) utopia value for the $i$-th objective $f_i(\vx)$. This scalarization method has a promising property: 

\textbf{Theorem 1 (\citet{choo1983proper}).} \textit{A feasible solution $\vx \in \mathcal{X}$ is weakly Pareto optimal if and only if there is a weight vector $\vlambda > 0$ such that $\vx$ is an optimal solution of the problem~(\ref{eq_tch_decomposition}).}

In other words, all Pareto solutions $\vx \in \mathcal{M}_{\text{ps}}$ can be found by solving the Tchebycheff scalarized subproblem~(\ref{eq_tch_decomposition}) with a specific (but unknown) trade-off preference $\vlambda$. We let $\mathcal{M}_{\text{tch}}$ be the solution set for problem (\ref{eq_tch_decomposition}) with all valid preferences $\Lambda$ and have $\mathcal{M}_{\text{ps}} \subseteq \mathcal{M}_{\text{weak}} = \mathcal{M}_{\text{tch}}$. The weakly Pareto optimal but not Pareto optimal solutions ($\mathcal{M}_{\text{weak}} \setminus \mathcal{M}_{\text{ps}}$) are dominated (but not strictly dominated) by some Pareto solutions, and are usually not desirable for decision-making. They can be further avoided by the augmented Tchebycheff approach~\citep{steuer1983interactive,kaliszewski1987modified}. In this work, we use the following scalarization:
\begin{eqnarray}
g_{\text{tch\_aug}}(\vx|\vlambda) = \max_{1 \leq i \leq m} \{ \lambda_i (f_i(\vx) - (z^*_i - \varepsilon)) \}   + \rho \sum_{i=1}^{m} \lambda_i f_i(\vx), \quad \forall \vlambda \in \Lambda,  
\label{eq_tch_scalarization_augmentation}
\end{eqnarray}
where $\rho$ is a sufficiently small positive scalar depends on the problem and current solution location. This form of augmentation has also been used in ParEGO~\citep{knowles2006parego}. With the augmentation term, the weakly dominated solutions will have larger scalarized values than the corresponding Pareto solutions in (\ref{eq_tch_scalarization_augmentation}), and will ultimately be eliminated with the optimization process (e.g., $\mathcal{M}_{\text{tch\_aug}} = \mathcal{M}_{\text{ps}}$). In this work, we simply set $\rho = 0.001$, dynamically update $z^*_i$ as the current best value for each objective and let $\boldsymbol{\varepsilon} = 0.1|\boldsymbol{z}^*|$. This setting is robust for all problems we considered. The traditional methods focus on solving the scalarization problem~(\ref{eq_tch_scalarization_augmentation}) with a finite set of different preferences $\vlambda$ in a sequential~\cite{knowles2006parego} or collaborative manner~\cite{zhang2007moea,zhang2010expensive}.

\textbf{Set Model.}
With augmented Tchebycheff scalarization, we propose to build a set model for mapping preferences to their solutions:
\begin{eqnarray}
\vx(\vlambda) = h_{\vtheta}(\vlambda),
\label{eq_pml_model}
\end{eqnarray}
where $\vlambda$ is any valid preference in $\Lambda = \{\vlambda \in \bbR^m_{+}|\sum \lambda_i = 1\}$, $\vx(\vlambda) \in \mathcal{X}$ is its corresponding Pareto solution, and $h_{\vtheta}(\vlambda)$ is the Pareto set model with parameter $\vtheta$. The input preference $\vlambda$ has ($m-1$) degree of freedom, and the output solution set $\mathcal{M}_{\text{psl}} = \{\vx = h_{\vtheta}(\vlambda) | \vlambda \in \Lambda \}$ is on an ($m-1$)-dimensional manifold in $\mathcal{X} \in \bbR^n$. In other words, the set model maps the ($m-1$)-dimensional regular preference simplex $\Lambda$ to the ($m-1$)-dimensional solution set $\mathcal{M}_{\text{psl}}$ with complicated structure.

\begin{figure*}[t]
    \centering
    \includegraphics[width= 1 \linewidth]{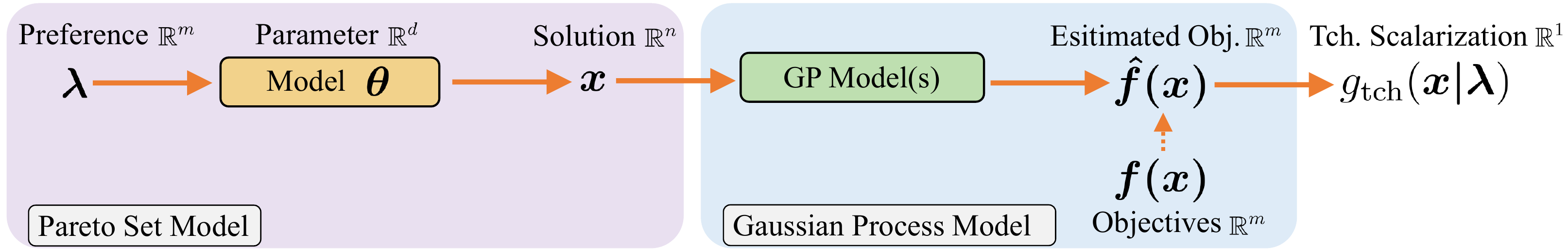}
    \caption{\textbf{Pareto Set Learning for Multi-Objective Bayesian Optimization:} \textbf{(a)} The Pareto set model learns a parameterized mapping from any valid preference $\vlambda \in \Lambda = \{\vlambda \in \bbR^m_{+}|\sum \lambda_i = 1\}$ to its corresponding solution $\vx(\vlambda) \in \bbR^n$. \textbf{(b)} We build independent Gaussian process models for each objective function. With these surrogate models, the set model can be efficiently trained to approximate the Pareto set. \textbf{(c)} In this work, we use the augmented Tchebycheff scalarization to connect each preference to its corresponding Pareto solution.}
    \label{fig_model_PSL}
\end{figure*}

We want to find the optimal parameters $\vtheta^*$ such that the generated set $\mathcal{M}_{\text{psl}}$ matches the solution set for augmented Tchebycheff scalarization $\mathcal{M}_{\text{tch\_aug}}=\{ \vx^*(\vlambda) |\vlambda \in \Lambda  \}$, where
\begin{equation}
\vx^*(\vlambda) = h_{\vtheta^*}(\vlambda) = \argmin_{\vx \in \mathcal{X}} g_{\text{tch\_aug}}(\vx|\vlambda), \forall \vlambda \in \Lambda. 
\label{eq_pml_model_tch}
\end{equation}
The learned mapping is illustrated in Figure~\ref{model_and_connection}. Once the connection is learned, we can explore the whole approximate Pareto set/front by simply adjusting the preferences among objectives. We use an MLP neural network as the set model for all MOBO problems, which is good at capturing complicated problem structures~\citep{sener2020learning}. The model details can be found in Appendix~\ref{sec_supp_model}. 

\subsection{Pareto Set Learning with Gaussian Process}

Since the evaluation of $\vf(\vx(\vlambda)) = \vf(h_{\vtheta}(\vlambda))$ is expensive, we use the surrogate model-based approach to learn the Pareto set model $h_{\vtheta}(\vlambda)$ as shown in Figure~\ref{fig_model_PSL}. Our method is orthogonal to the choice of surrogate models, and we build independent Gaussian process models for each objective as in other MOBO methods~\citep{daulton2020differentiable, lukovic2020diversity}.   

\textbf{Gaussian Process Model.}
A single-objective Gaussian process~\citep{rasmussen2006gaussian} has a prior distribution defined on the function space:
\begin{eqnarray}
f(\vx) \sim GP(\mu(\vx),k(\vx,\vx)),
\label{eq_gaussian_process}
\end{eqnarray}
where $\mu: \mathcal{X} \rightarrow \bbR$ is the mean function and $k: \mathcal{X} \times \mathcal{X} \rightarrow \bbR$ is the covariance kernel function. With $n$ evaluated solutions $\{\vX, \vy\} = \{[\vx^{(i)}],[f(\vx^{(i)}]|i = 1,\ldots,n)\}$, the posterior distribution can be updated by maximizing the marginal likelihood based on the data. For a new solution $\vx^{n+1}$, the posterior mean and variance are:
\begin{align}
\hat{\mu}(\vx^{(n+1)}) = \mu(\vx^{(n+1)}) + \vk^T\vK^{-1}\vy, \quad
\hat{\sigma}^2(\vx^{(n+1)}) = k(\vx^{(n+1)},\vx^{(n+1)}) -  \vk^T \vK^{-1}\vk, 
\label{eq_gaussian_process_prediction}
\end{align}
where $\vk = k(\vx,\vX)$ is the kernel vector, $\vK = k(\vX, \vX)$ is the kernel matrix, Mat\'{e}rn 5/2 kernel are used for all models in this work. For $m$ independent GP models, we let $\hat \vmu(\vx) = [\hat \mu_1(\vx), \cdots,\hat \mu_m(\vx)]$ and $\hat \vsigma^2 (\vx) = [\hat \sigma^2_1(\vx), \cdots,\hat \sigma^2_m(\vx)]$ be the predicted mean and variance for the objective vector. Suppose we have a learned Pareto set $\mathcal{M}_{\text{psl}}$, the GP models give us both predicted value $\hat \vmu(\mathcal{M}_{\text{psl}}) = \{ \hat \vmu(\vx) | \vx \in \mathcal{M}_{\text{psl}}\}$ and uncertainty $\hat \vsigma^2(\mathcal{M}_{\text{psl}}) =  \{ \hat \vsigma^2(\vx) | \vx \in \mathcal{M}_{\text{psl}}\}$ for the whole approximate Pareto set. 

\begin{wrapfigure}{R}{0.5\linewidth}
    \vspace{-0.3in}
    \begin{minipage}{0.5\textwidth}
    \begin{algorithm}[H]
    	\caption{PSL with GP Models}
    	\label{alg_pml_gp}
     	\begin{algorithmic}[1]
        	\STATE \textbf{Input:} Model $\vx(\vlambda) = \vh_{\theta}(\vlambda)$
        	\STATE Initialize the parameters $\vtheta_0$ 
    		\FOR{$t = 1$ to $T$}
    		   \STATE Sample preferences $\{\vlambda_k\}_{k=1}^K \sim \Lambda$
    		   \STATE Update $\vtheta_t$ with gradient descent in (\ref{eq_mc_sgd})
    		\ENDFOR	
    		\STATE \textbf{Output:} Model $\vx(\vlambda) = h_{\vtheta_T}(\vlambda)$
     	\end{algorithmic}
    \end{algorithm}
    \end{minipage}
\end{wrapfigure}

\textbf{Pareto Set Learning.}
Now we propose an efficient algorithm to find the optimal parameter $\vtheta^*$ for the Pareto set model $h_{\vtheta}(\vlambda)$. The optimal solution set $\mathcal{M}_{\text{tch\_aug}}$ for augmented Tchebycheff scalarization~(\ref{eq_tch_scalarization_augmentation}) is unknown, hence we need to optimize all solutions generated by our model $\vx(\vlambda) = h_{\vtheta}(\vlambda)$ with respect to their corresponding augmented Tchebycheff scalarization subproblems for all valid preferences:
\begin{eqnarray}
 \vtheta^* = \argmin_{\vtheta} \bbE_{\vlambda \sim \Lambda}
 g_{\text{tch\_aug}}(\vx = h_{\vtheta}(\vlambda)|\vlambda).
 \label{eq_expectation}
\end{eqnarray}
If the model is perfectly learned, the obtained approximate Pareto set $\mathcal{M}_{\text{psl}} = \{\vx = h_{\vtheta}(\vlambda)| \vlambda \in \Lambda\}$ should be the same as  $\mathcal{M}_{\text{tch\_aug}}$. However, it is difficult to directly optimize (\ref{eq_expectation}) due to the expectation over infinite preferences ($|\vLambda| = \infty$). We use Monte Carlo sampling and gradient descent to iteratively learn the model with the surrogate model:
\begin{eqnarray}
 \vtheta_{t+1} = \vtheta_{t} - \eta \sum_{k=1}^{K} \nabla_{\vtheta} \hat g_{\text{tch\_aug}}(\vx = h_{\vtheta}(\vlambda_k)|\vlambda_k),
 \label{eq_mc_sgd}
\end{eqnarray}
where we randomly sample $K = 10$ different valid preferences $\{\vlambda_1, \cdots, \vlambda_K\} \sim \Lambda$ at each iteration in this work. Here $\hat g_{\text{tch\_aug}}(\cdot)$ is the augmented Tchebycheff scalarization with predicted objective values:
\begin{equation}
\hat g_{\text{tch\_aug}}(\vx|\vlambda) = \max_{1 \leq i \leq m} \{ \lambda_i (\hat f_i(\vx) - (z^*_i - \varepsilon)) \}  + \rho \sum_{i=1}^{m} \lambda_i \hat f_i(\vx).  
\label{eq_tch_decomposition_gp}
\end{equation}
One design issue left is how to set the surrogate objective $\hat \vf(\vx)$. If we only want to obtain the current predictive Pareto front, it is straightforward to use the posterior mean as the surrogate value. The approximate Pareto front under the posterior mean could provide valuable information to decision-makers. However, for Bayesian optimization, we have to take the uncertainty into account to balance exploitation and exploration. Many widely-used criteria, such as expected improvement (EI)~\cite{movckus1975bayesian} and upper confidence bound (UCB)~\cite{srinivas2009gaussian}, could be a more reasonable choice. In this work, we use the lower confidence bound (LCB) for minimization problems.
\begin{equation}
\hat \vf(\vx) = \hat \vmu(\vx) - \beta \hat \vsigma(\vx).
\label{eq_ucb}
\end{equation}
We simply set $\beta = \frac{1}{2}$ and discuss the performance with other surrogate values in Appendix~\ref{subsec_supp_acq}.

The expensive objective function $\vf(\vx)$ is usually black-box and non-differentiable, but we can easily obtain the gradients for the Gaussian process and the set model. Indeed, gradient-based methods have been widely used for optimizing the acquisition function in both BO~\citep{wu2016parallel,wilson2018maximizing} and MOBO~\citep{daulton2020differentiable, lukovic2020diversity}. The max operator in Tchebycheff scalarization is technically only subdifferentiable, but it is known to have good subgradients~\citep{wilson2017reparameterization} for surrogate optimization and can preserve convexity if the objectives $\{\hat f_i(\vx)\}_{i=1}^{m}$ are all convex~\citep{boyd2004convex}. 

The Pareto set learning algorithm with Gaussian process models is summarized in \textbf{Algorithm~\ref{alg_pml_gp}}. We find that the simple random initialization and gradient descent are enough to learn a good Pareto set approximation. The overparameterized neural network could be beneficial to overcome potential non-convexity~\citep{lopez2018easing}.

\subsection{Batched selection on approximate Pareto set}

\begin{minipage}[t]{0.55\textwidth}
  \centering
  \vspace{0pt}  
  \begin{algorithm}[H]
	\caption{MOBO with PSL}
	\label{alg_mobo_pml}
 	\begin{algorithmic}[1]
    	\STATE \textbf{Input:} Black-box vector-valued function $\vf(\vx)$
    	\STATE Initial Samples $\{\vX_0,\vy_0\}$
		\FOR{$t = 1$ to $T$}
		   \STATE Train GPs based on $\{\vX_{t-1},\vy_{t-1}\}$
		   \STATE Learn set model $\vh_{\vtheta_t}(\vlambda)$ with GPs (\textbf{Alg. 1})
		   \STATE Select $\{\vx^{(b)}\}_{b=1}^{B}$ with the set model (\textbf{Alg. 3})
		   \STATE $\vX_t \leftarrow \vX_{t-1} \cup \{\vx^{(b)}\}_{b=1}^{B}$, \\ $\vy_t \leftarrow \vy_{t-1} \cup \vf(\{\vx^{(b)}\}_{b=1}^{B})$
		\ENDFOR	
		\STATE \textbf{Output:} $\{\vX_{t},\vy_{t}\}$ and final set model $h_{\vtheta_T}(\vlambda)$
 	\end{algorithmic}
\end{algorithm}
\end{minipage}
\hfill
\begin{minipage}[t]{0.41\textwidth}
  \centering
  \vspace{0pt}  
\begin{algorithm}[H]
	\caption{Batch Selection with PSL}
	\label{alg_pml_batch_selection}
 	\begin{algorithmic}[1]
    	\STATE \textbf{Input:} Model $\vx(\vlambda) = \vh_{\theta}(\vlambda)$, Batch Size $B$
        \STATE Sample $P$ preferences \\ $\{\vlambda^{(p)}\}_{p=1}^P \sim \Lambda$
        \STATE Generate solutions \\ $\vX = \{\vx(\vlambda^{(p)})\}_{p=1}^{P}$ on $\mathcal{M}_{\text{psl}}$
        \STATE Find subset $\{\vx^{(b)}\}_{b=1}^{B} \subset \vX$ that has the highest $\text{HVI}(\hat \vf(\{\vx^{(b)}\}_{b=1}^{B}))$
        \STATE \textbf{Output:} Batch solutions $\{\vx^{(b)}\}_{b=1}^{B}$
 	\end{algorithmic}
\end{algorithm}
\end{minipage}
\vspace{10pt} 

In this subsection, we propose a lightweight yet efficient batched acquisition search for MOBO with the learned Pareto set model. The algorithm framework is shown in \textbf{Algorithm~\ref{alg_mobo_pml}}. The crucial difference with other MOBO approaches is that we build a set model at each iteration for batched solution selection as shown in \textbf{Algorithm~\ref{alg_pml_gp}} and \textbf{Algorithm~\ref{alg_pml_batch_selection}}. The batched selection procedure contains two closely related steps:

\textbf{Batch Sampling on Approximate Pareto Set.} Our model naturally supports generating an arbitrary number of solutions in batch. If the decision-maker's preferences are available, we can use preference-based sampling in this step. In this work, without any prior knowledge, we uniformly sample $P$ valid preferences $\{\vlambda^{(p)}\}_{p=1}^P$, and generate the corresponding solutions $\vX = \{\vx(\vlambda^{(p)})\}_{p=1}^{P}$ on the approximate Pareto set $\mathcal{M}_{\text{psl}}$. 

\textbf{Batch Selection.} At each iteration of MOBO, we typically select a small number $B$ (e.g., $5$) of solutions $\vX_B = \{\vx^{(b)}\}_{b=1}^{B}$ from the sampled solutions $\vX$ for expensive evaluations. To take all already evaluated solutions $\{\vX_{t-1}, \vy_{t-1}\}$ into consideration, we use the hypervolume~\citep{zitzler2007hypervolume} as the selection criteria. The hypervolume $\text{HV}(\vy) = \textbf{Vol}(\vS)$ measures the volume of $\vS$ dominated by a set $\vy$ in the objective space:
\looseness=-1
\begin{eqnarray}
\vS = \{r \in \mathbb{R}^m \mid \exists y \in \vy \mbox{ such that } y \prec r \prec r^*\},
\end{eqnarray}
where $r^*$ is a reference point that dominated by all $y \in \vy$. The hypervolume improvement (HVI) of a set $\vX_{B}$ with respect to the already evaluated solutions $\{\vX_{t-1}, \vy_{t-1}\}$ can be defined as:
\begin{eqnarray}
      \text{HVI}(\hat \vf(\vX_{B})) = \text{HV}(\vy_{t-1} \cup \hat \vf(\vX_{B})) - \text{HV}(\vy_{t-1}),
 \label{eq_hvi}
\end{eqnarray}
where $\vX_{B} = \{\vx^{(b)}\}_{b=1}^{B}$ are selected solutions, and $\hat \vf(\vX_{B})$ are the surrogate values. In this work, we mainly use the LCB~(\ref{eq_ucb}) as the surrogate value for Bayesian optimization, and provide an ablation study of different surrogate values in Appendix~\ref{subsec_supp_acq}.

A better trade-off set will have a larger hypervolume, and the true Pareto set always has the largest one. We want to select a set of $\vX_{B}$ such that their corresponding objective values
$\hat \vf(\vX_{B})$ maximize $\text{HVI}(\hat \vf(\vX_{B}))$. It would be computationally expensive to jointly optimize a set of solutions to exactly maximize the hypervolume improvement (\ref{eq_hvi}), and therefore sequential greedy selection is typically used~\cite{daulton2020differentiable}. In this work, we select the set $\vX_{B}$ in a sequential greedy manner from $\vX$ where $|\vX| = P = 1,000$ for all problems. More details can be found in Appendix~\ref{subsec_supp_batch}.

\section{Experiments}
\label{sec_experiment}

In this section, we  compare the proposed PSL method with other MOBO approaches on the performance of evaluated solutions. We also analyze the quality of the learned Pareto set model, which other methods cannot produce.

\textbf{Baseline Algorithms.} We consider several widely-used MOBO methods and two model-free approaches as baselines. The implementations of NSGA-II~\citep{deb2002fast}, MOEA/D-EGO~\citep{zhang2010expensive}, TSEMO~\citep{bradford2018efficient}, USeMO-EI~\citep{belakaria2020uncertainty}, DGEMO~\citep{lukovic2020diversity} are from DGEMO's open-source codebase\footnote{\url{https://github.com/yunshengtian/DGEMO}} based on pymoo\footnote{\url{https://pymoo.org/problems/index.html}}~\citep{pymoo}. The implementations of scrambled Sobol sequence, qParEGO~\citep{knowles2006parego}, TS-TCH~\citep{paria2020flexible}, qEHVI~\citep{daulton2020differentiable} and qNEHVI~\citep{daulton2021parallel} and from BoTorch\footnote{\url{https://github.com/pytorch/botorch}}~\cite{balandat2020botorch}. We implement the proposed PSL\footnote{\url{https://github.com/Xi-L/PSL-MOBO}} in Pytorch~\citep{paszke2019pytorch}.

\textbf{Benchmarks and Real-World Problems.}
The algorithms are first compared on six newly proposed synthetic test instances (see Appendix~\ref{subsec_supp_benchmark}), as well as the widely-used VLMOP1-3~\citep{van1999multiobjective} and DTLZ2~\citep{deb2002scalable} benchmark problems. Then we also conduct experiments on $5$ different real-world multi-objective engineering design problems (RE)~\citep{tanabe2020easy}, including 1) four bar truss design~\citep{cheng1999generalized}; 2) pressure vessel design~\citep{kramer1994augmented}; 3) disk brake design~\citep{ray2002swarm}; 4) gear train design~\citep{deb2006innovization} and 5) rocket injector design~\citep{vaidyanathan2003cfd}. Details of these problems can be found in Appendix~\ref{sec_supp_problem}.

\textbf{Experiment Setting.} For each experiment, we randomly generate $10$ initial solutions for expensive evaluations, and then conduct MOBO with $20$ batched evaluations with batch size $5$. Therefore, there are total $110$ expensive evaluations. For an experiment, all algorithms are independently run $10$ times. We use the hypervolume indicator~\citep{zitzler2007hypervolume} as the metric to compare the quality of evaluated solutions chosen by different MOBO algorithms, which is monotonic to the Pareto dominance relation. The ground truth Pareto front will always have the best (highest) hypervolume.

\subsection{Experimental results and analysis}

\begin{table}[ht]
\vspace{-0.2in}
\small
\centering
\caption{Algorithm runtime per iteration (in seconds).}
\label{table_run_time}
\tabcolsep=0.12cm
\begin{tabular}{lccccccc}
\toprule
Problem          & \#objs      & MOEA/D-EGO & TSEMO & \multicolumn{1}{l}{USeMO-EI} & \multicolumn{1}{l}{DGEMO} & qEHVI & \multicolumn{1}{l}{PSL(Ours): Model + Selection} \\ \midrule
F1    & 2      & 40.95      & 4.82  & 6.12                         & 61.48                     & 36.71 & 5.26 + 1.33 = 6.59   \\
DTLZ2   & 3      & 71.83      & 7.28  & 8.76                         & 83.57                     & 75.92 &7.02 + 1.59 = 8.61   \\ \bottomrule
\end{tabular}
\end{table}

\begin{figure*}[ht]
\captionsetup[subfigure]{font=scriptsize,labelfont=scriptsize}
\centering
\subfloat[F1]{\includegraphics[width = 0.33\linewidth]{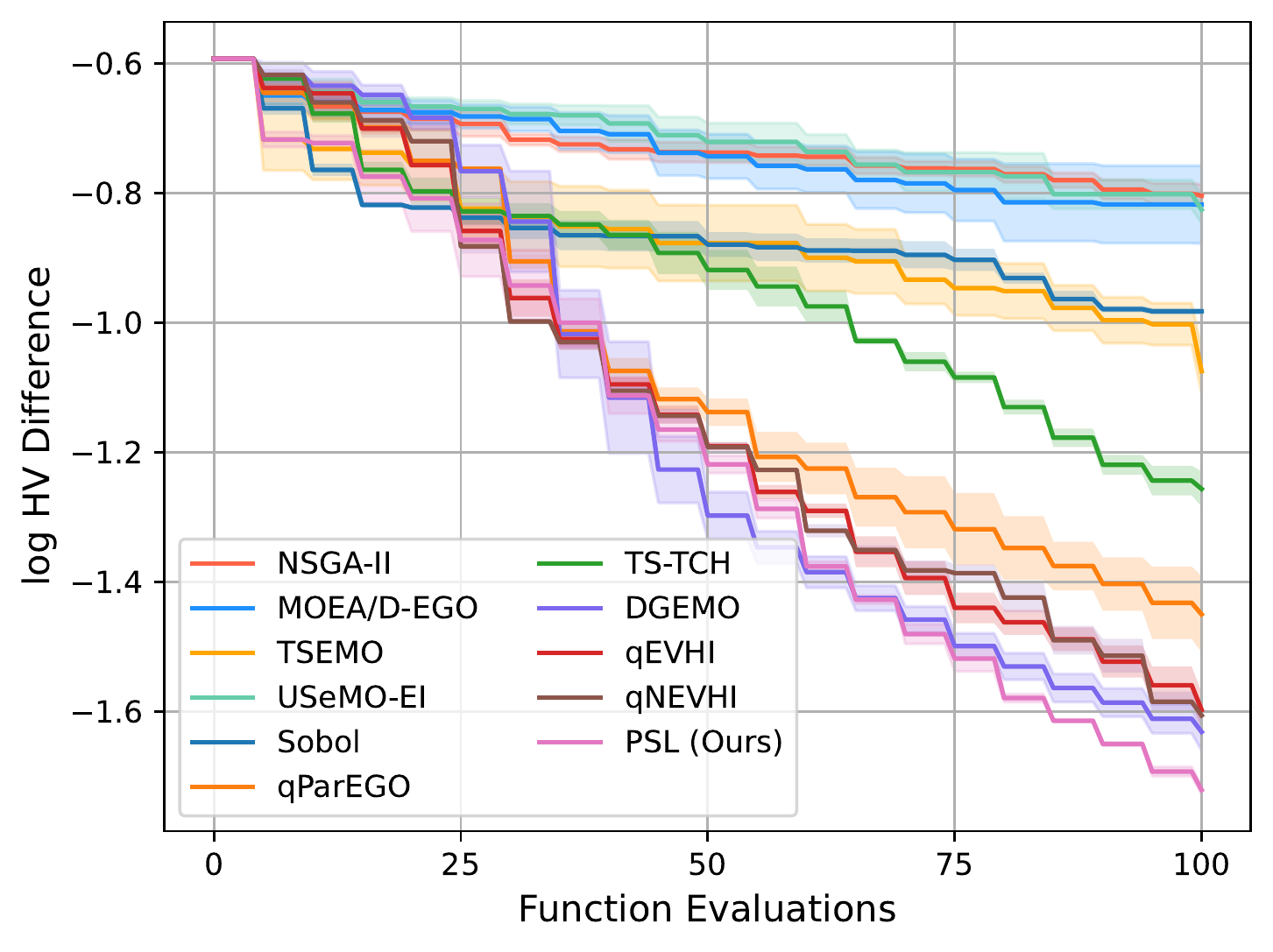}}\hfill
\subfloat[F2]{\includegraphics[width = 0.33\linewidth]{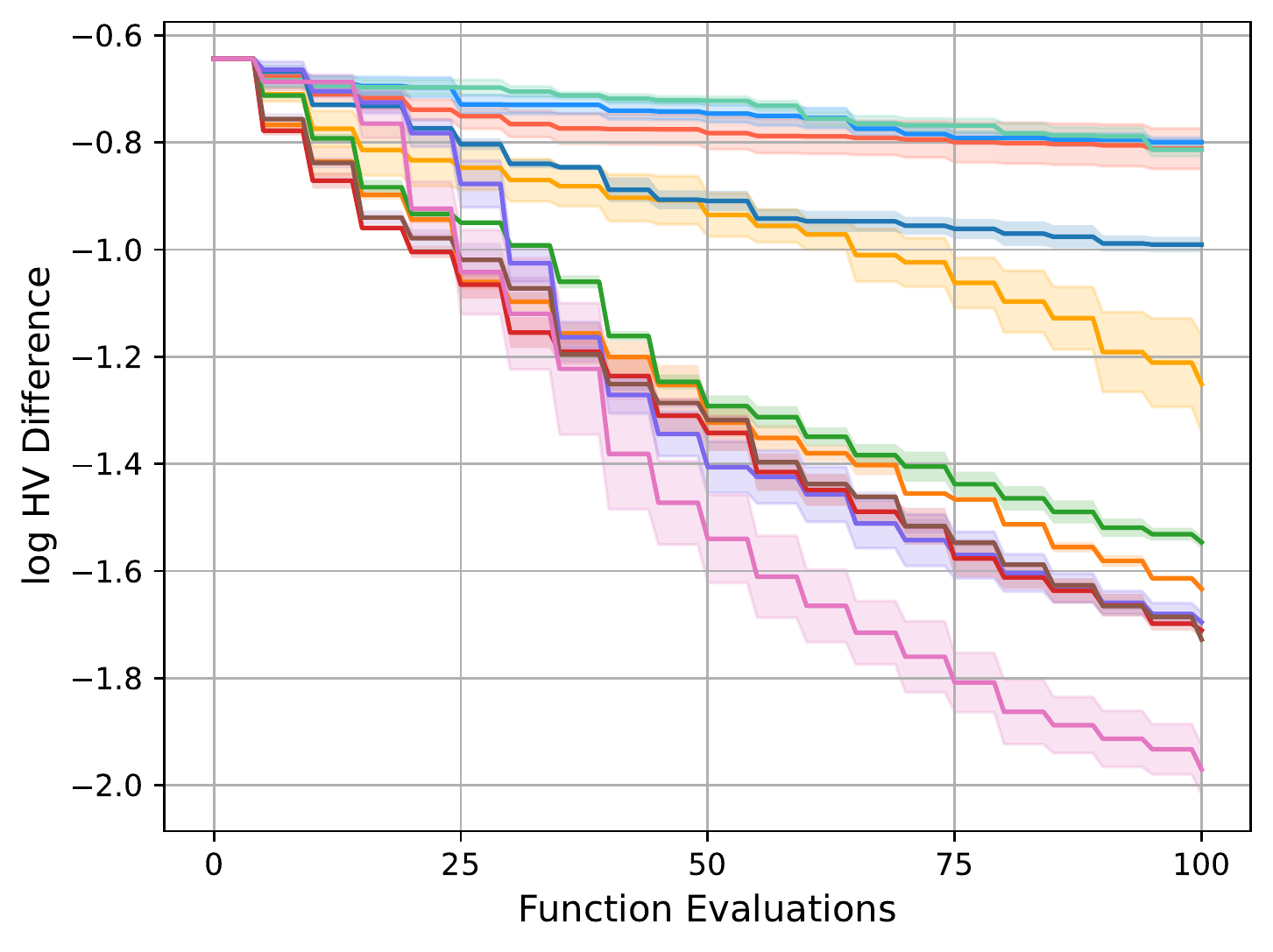}}\hfill
\subfloat[F3]{\includegraphics[width = 0.33\linewidth]{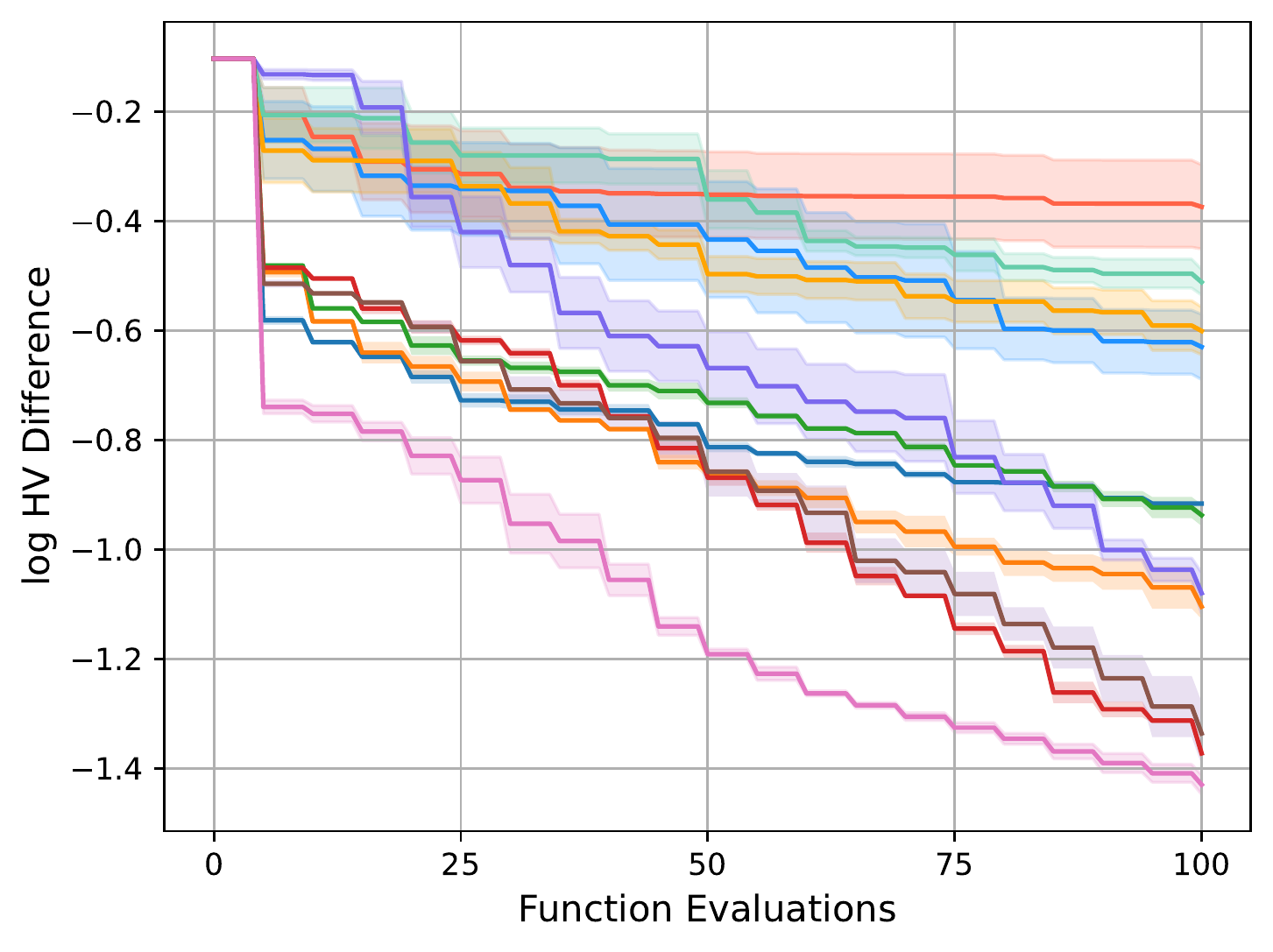}}\hfill \\
\vspace{-0.1in}
\subfloat[F4]{\includegraphics[width = 0.33\linewidth]{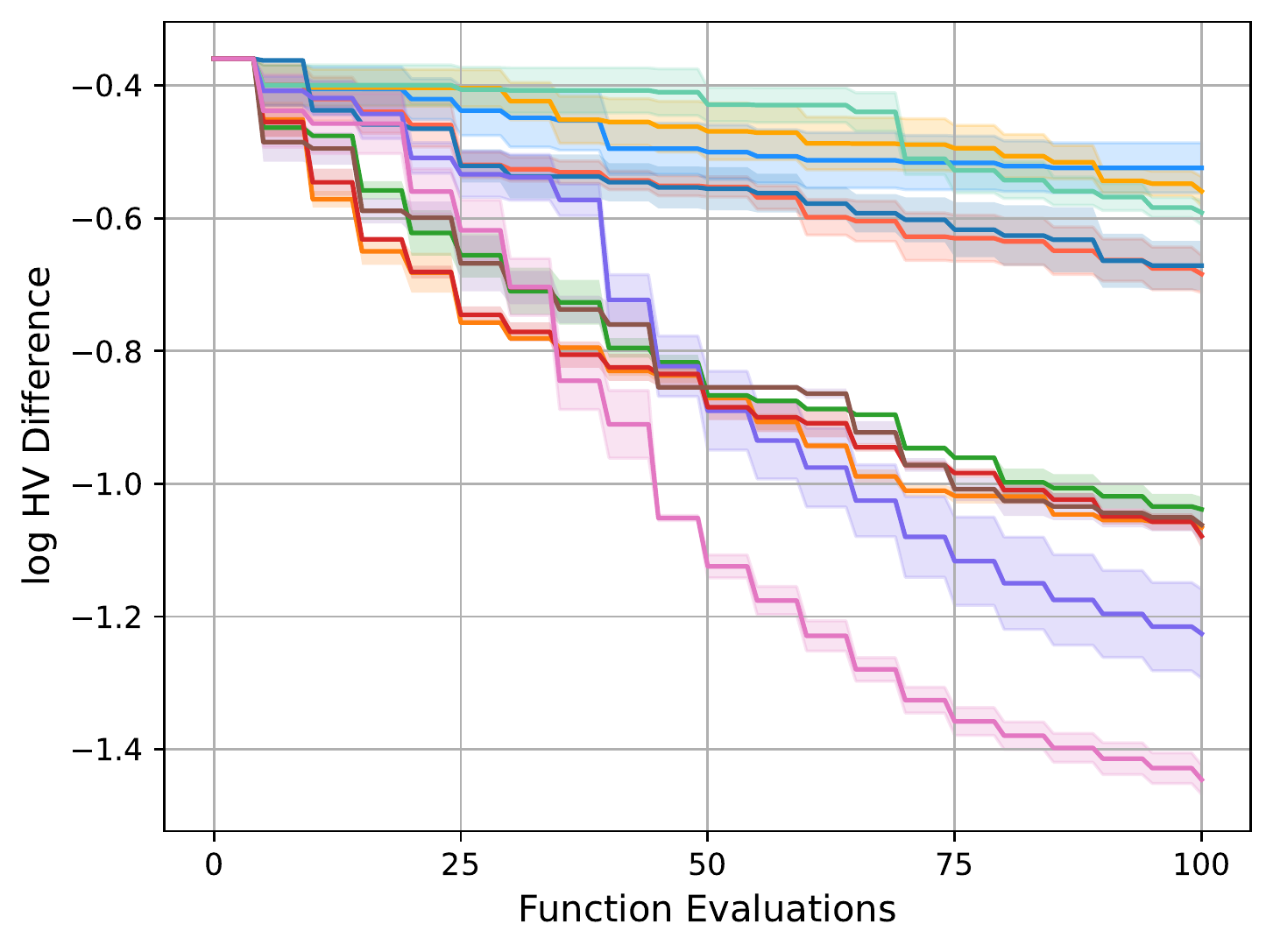}}\hfill
\subfloat[F5]{\includegraphics[width = 0.33\linewidth]{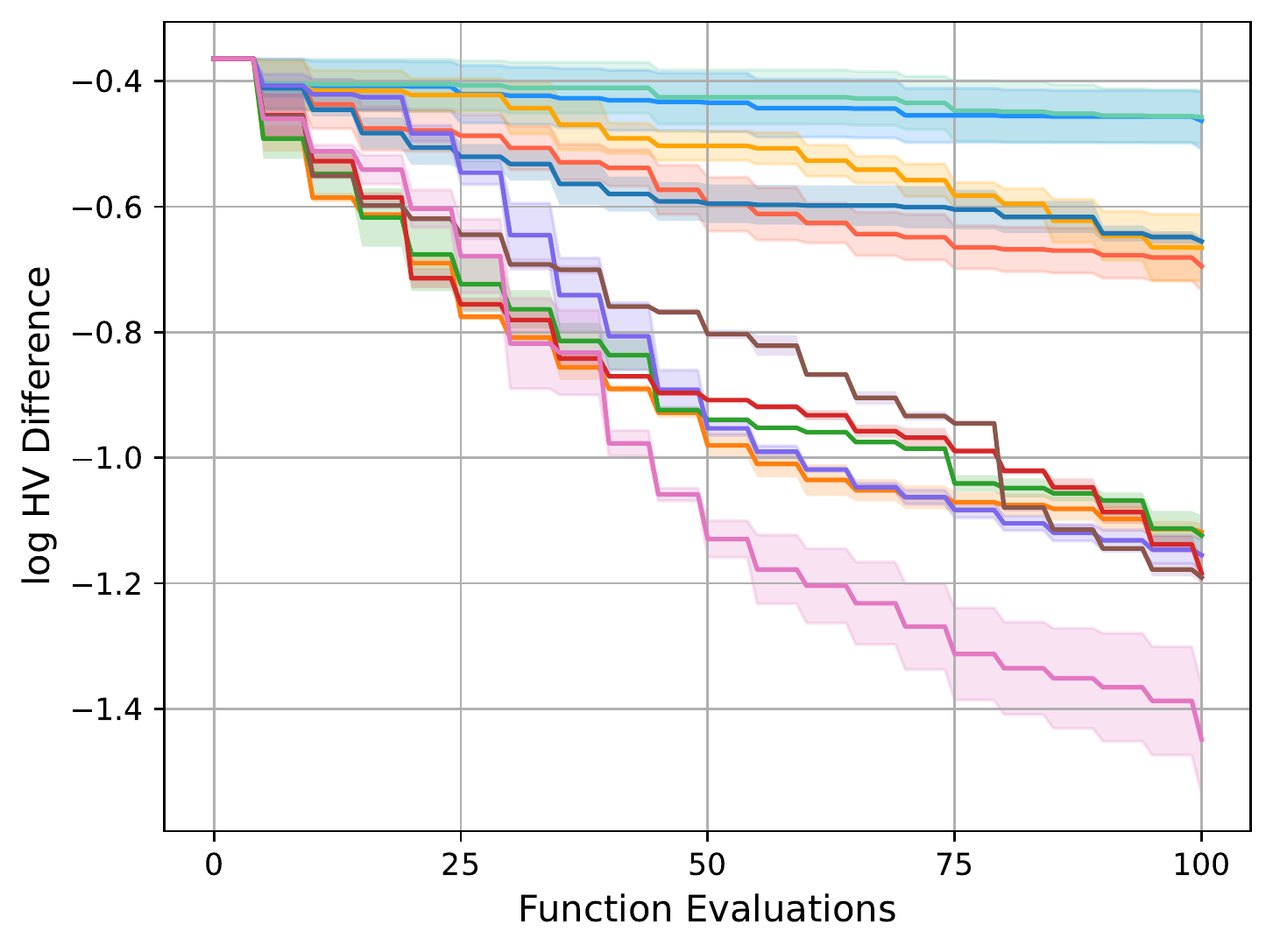}}\hfill
\subfloat[F6]{\includegraphics[width = 0.33\linewidth]{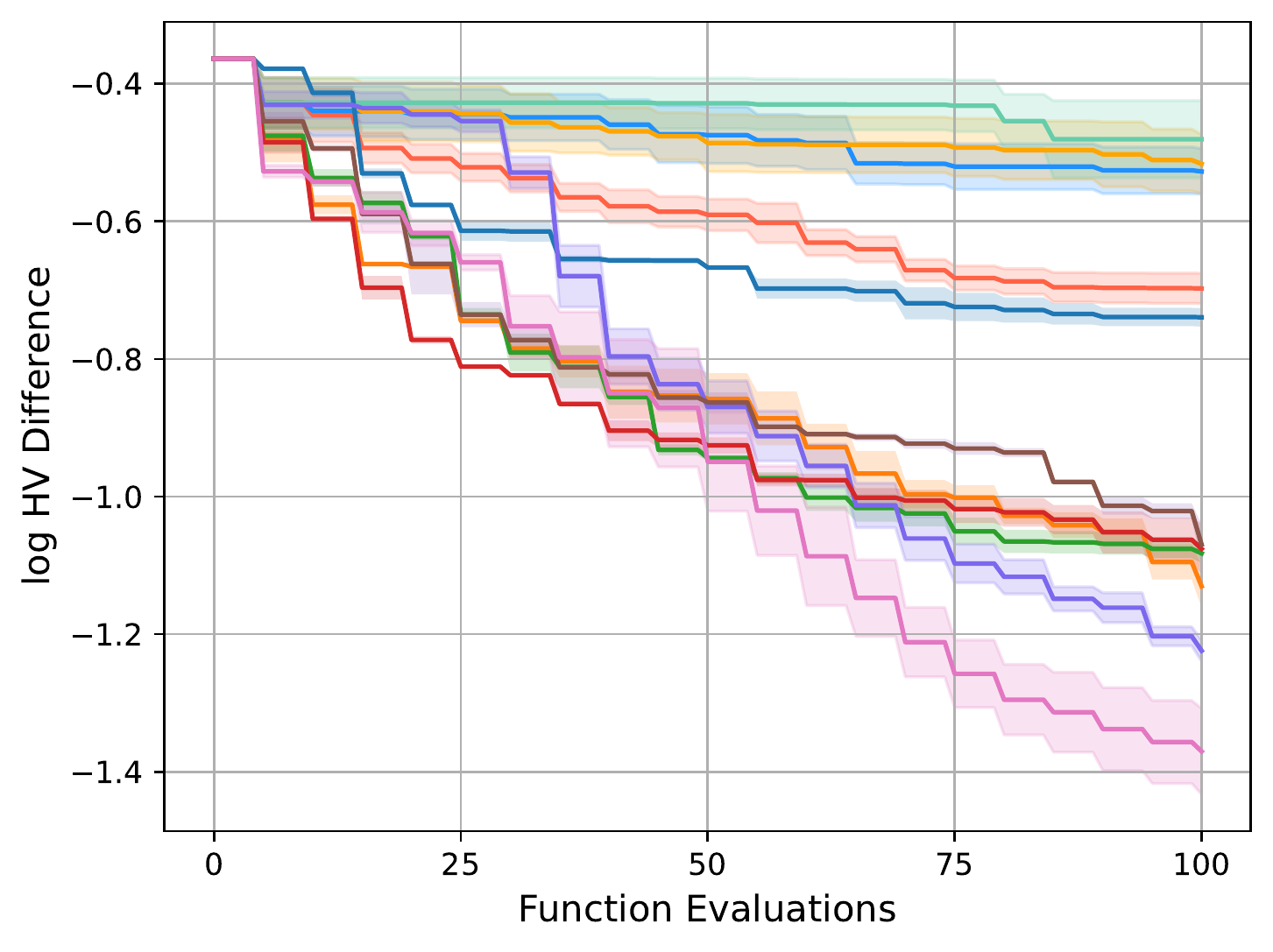}}\hfill \\
\vspace{-0.1in}
\subfloat[VLMOP1]{\includegraphics[width = 0.33\linewidth]{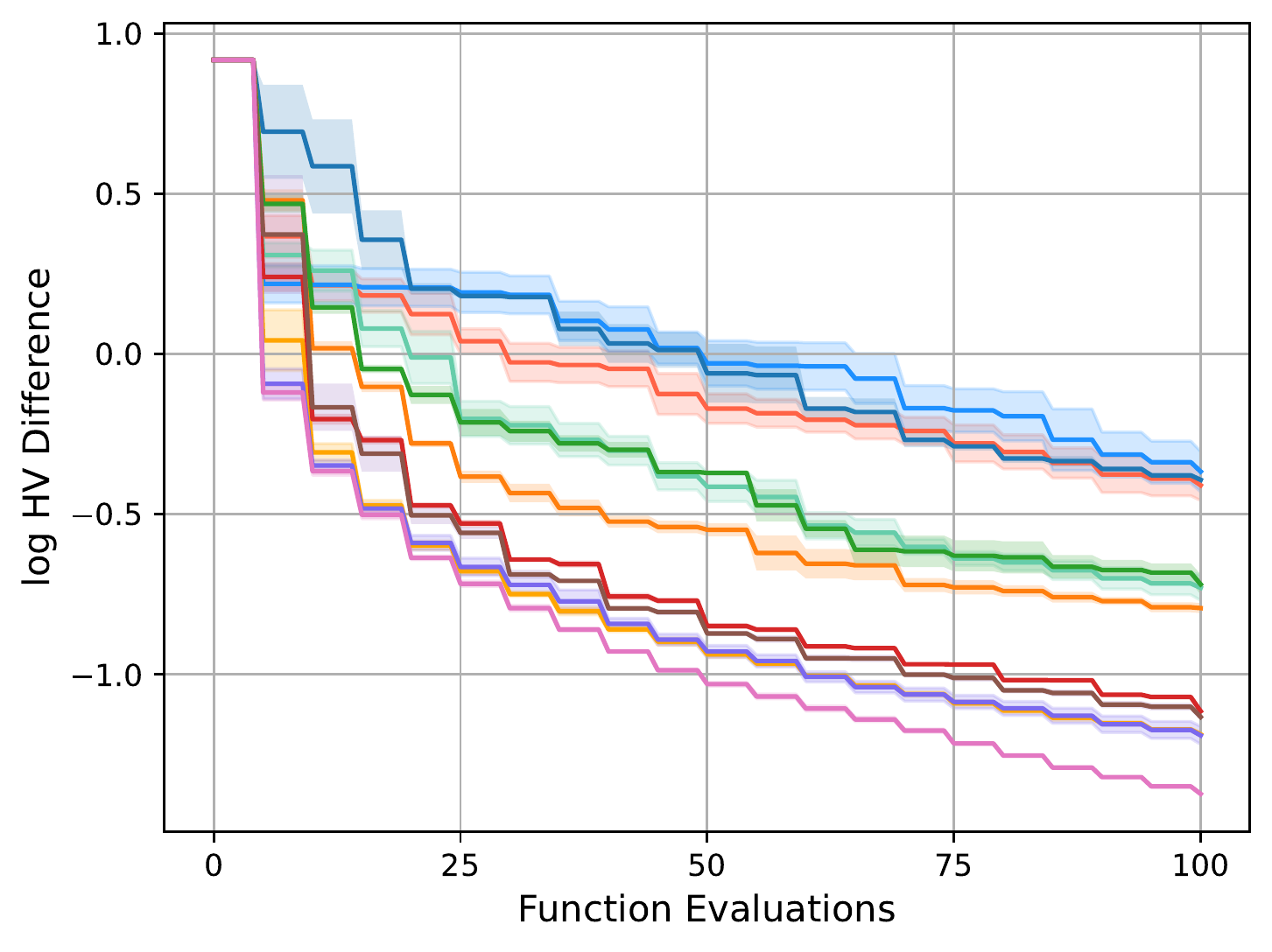}}
\subfloat[VLMOP2]{\includegraphics[width = 0.33\linewidth]{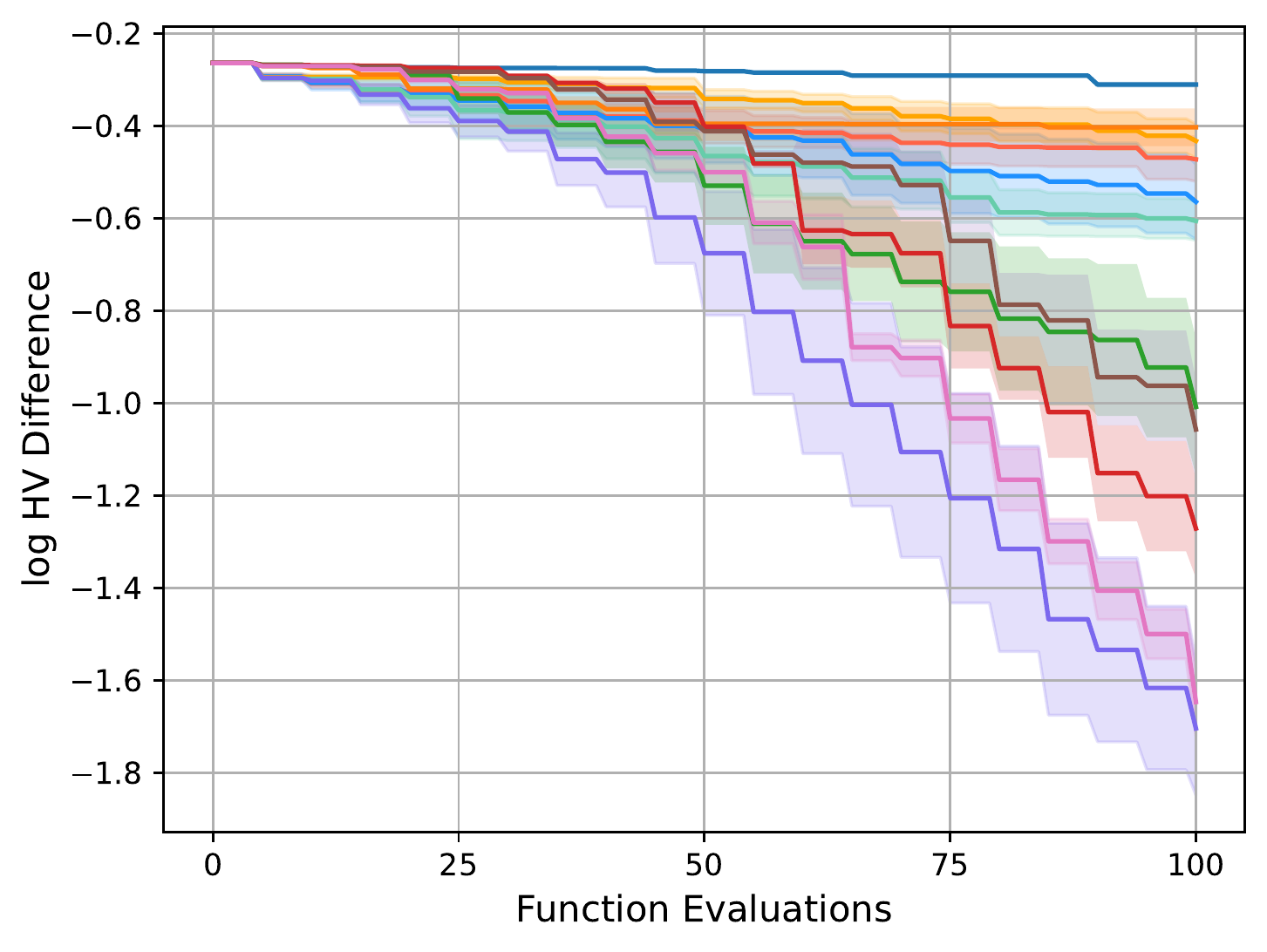}}\hfill
\subfloat[VLMOP3]{\includegraphics[width = 0.33\linewidth]{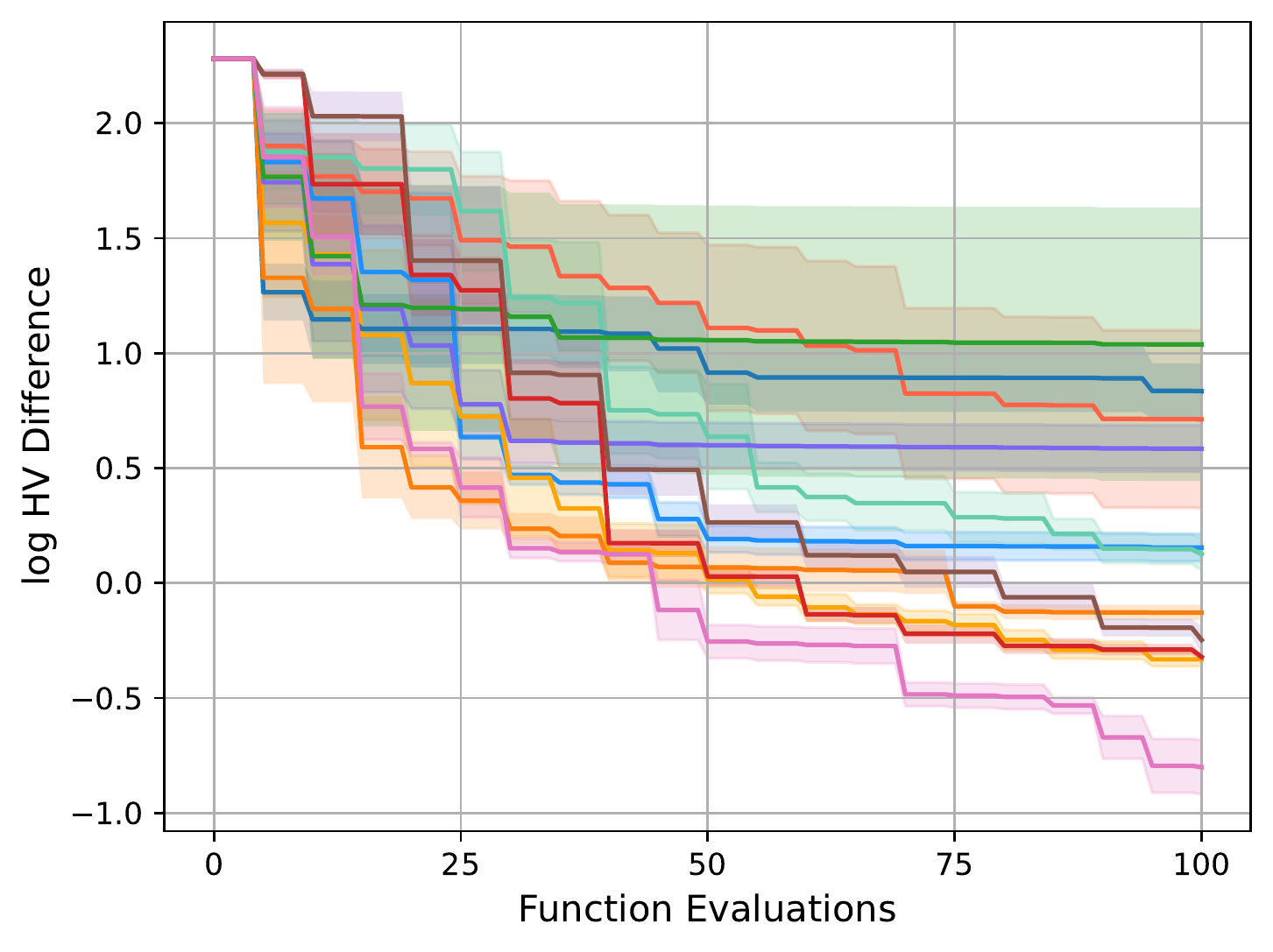}}\hfill \\
\vspace{-0.1in}
\subfloat[DTLZ2]{\includegraphics[width = 0.33\linewidth]{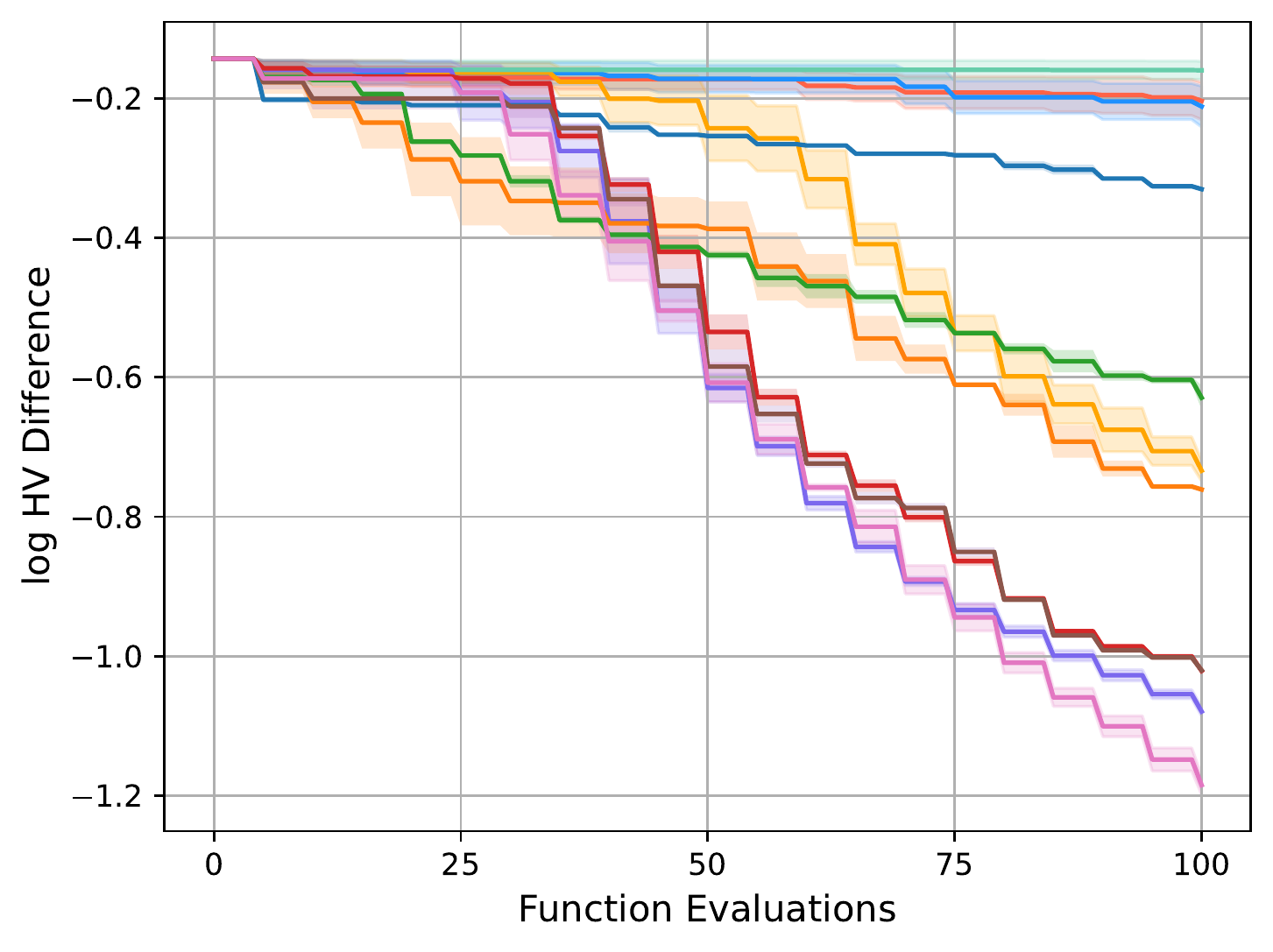}}\hfill
\subfloat[Four Bar Truss Design]{\includegraphics[width = 0.33\linewidth]{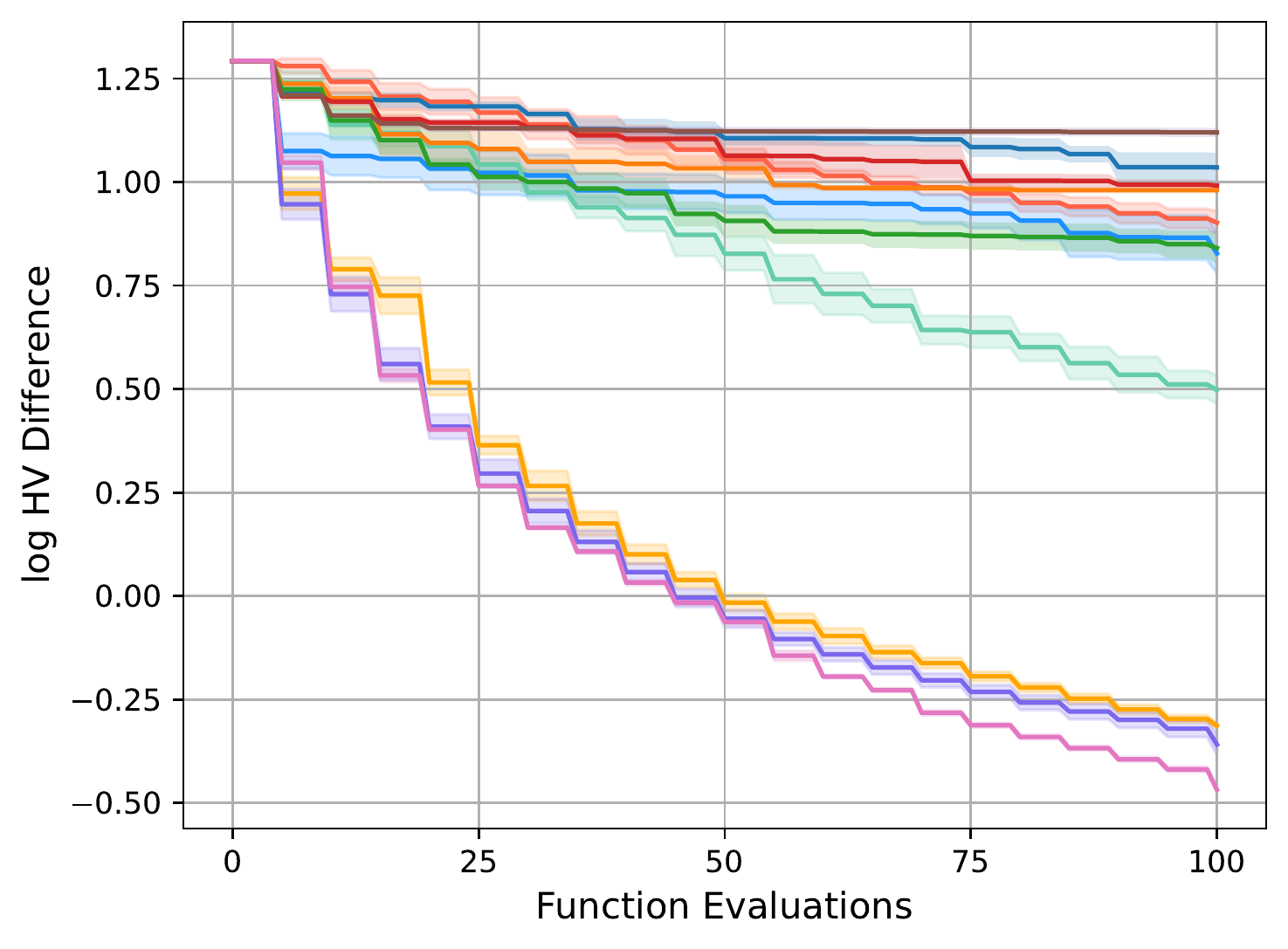}}
\subfloat[Pressure Vessel Design]{\includegraphics[width = 0.33\linewidth]{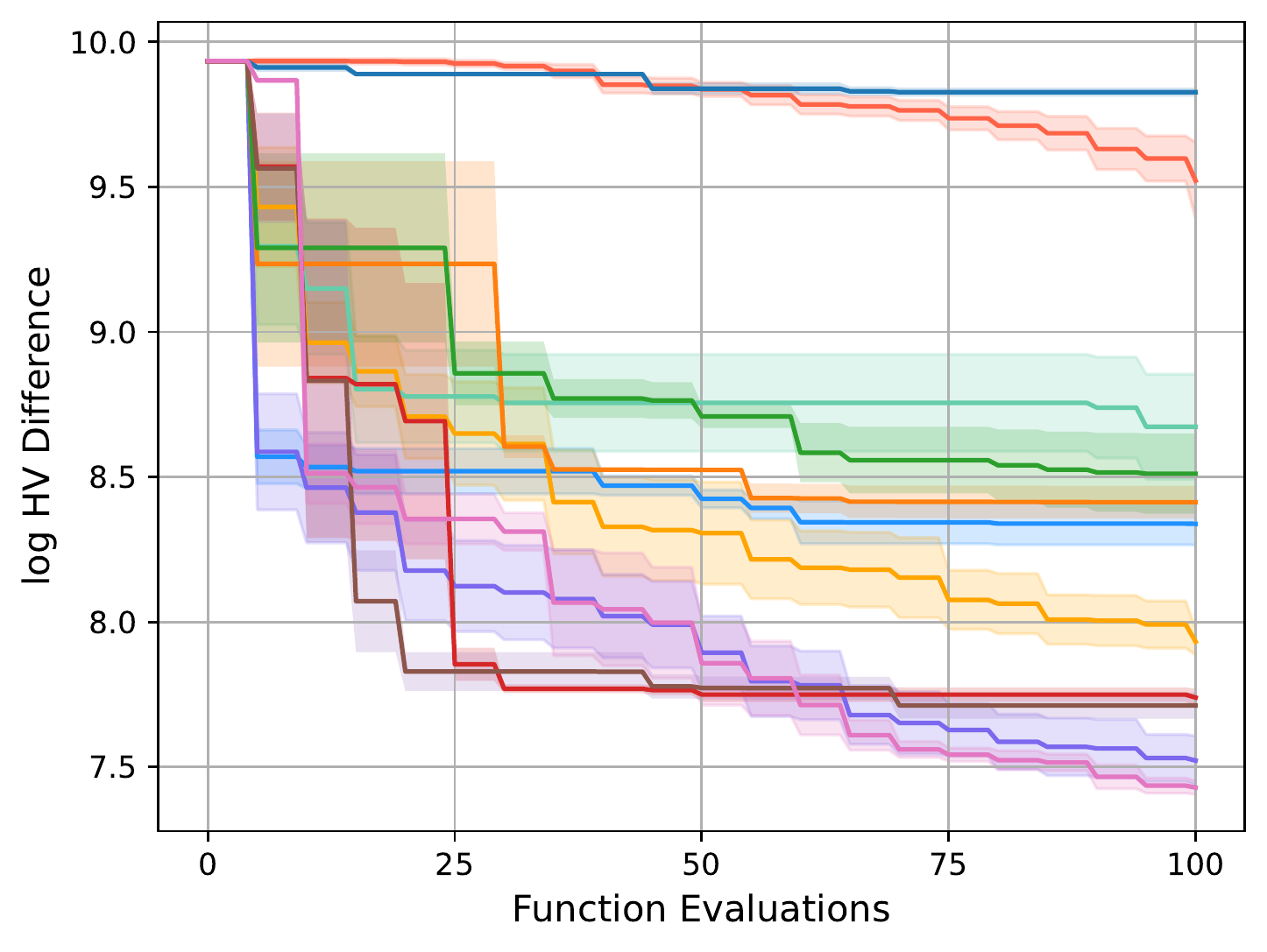}}\hfill \\
\vspace{-0.1in}
\subfloat[Disk Brake Design]{\includegraphics[width = 0.33\linewidth]{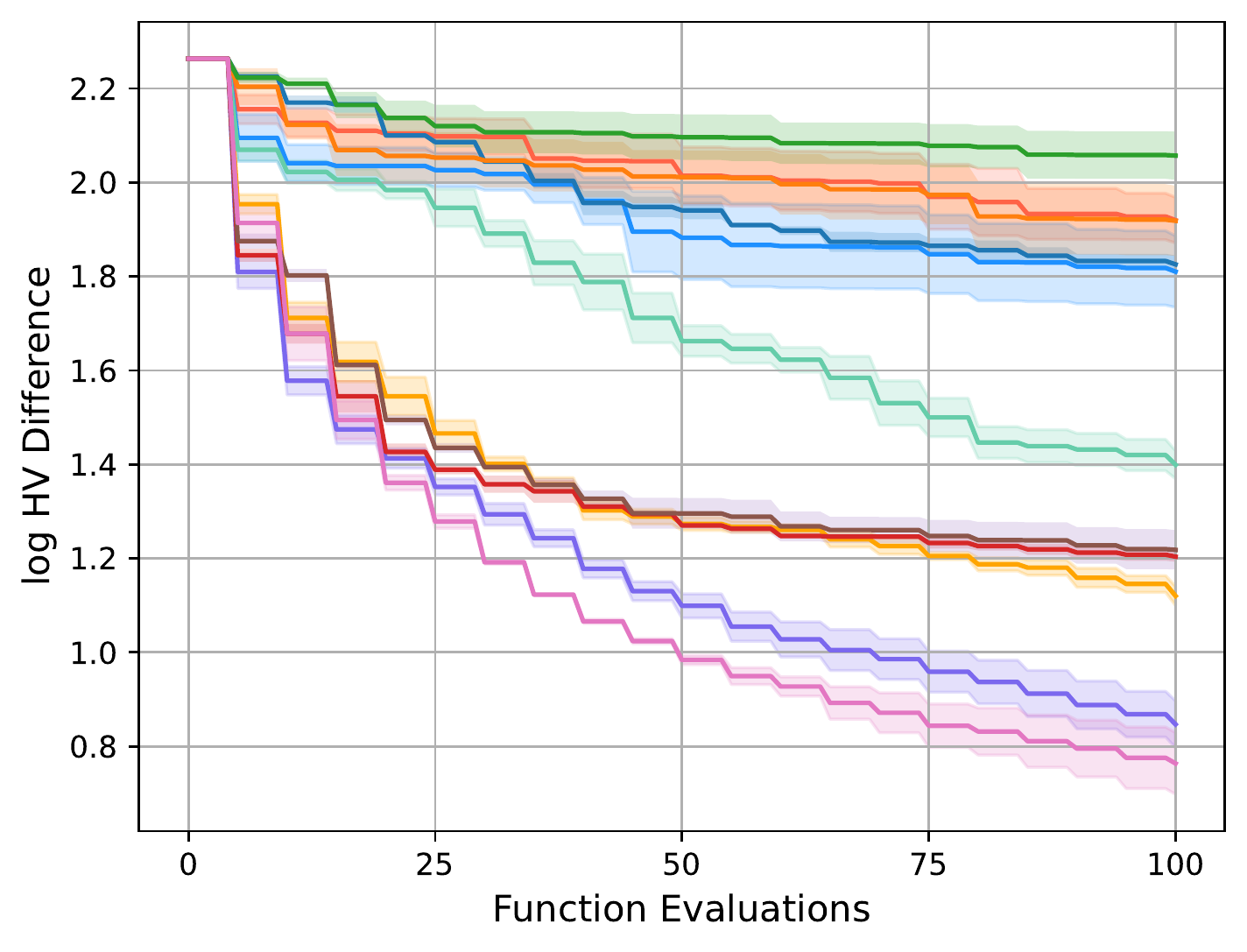}}\hfill
\subfloat[Gear Train Design]{\includegraphics[width = 0.33\linewidth]{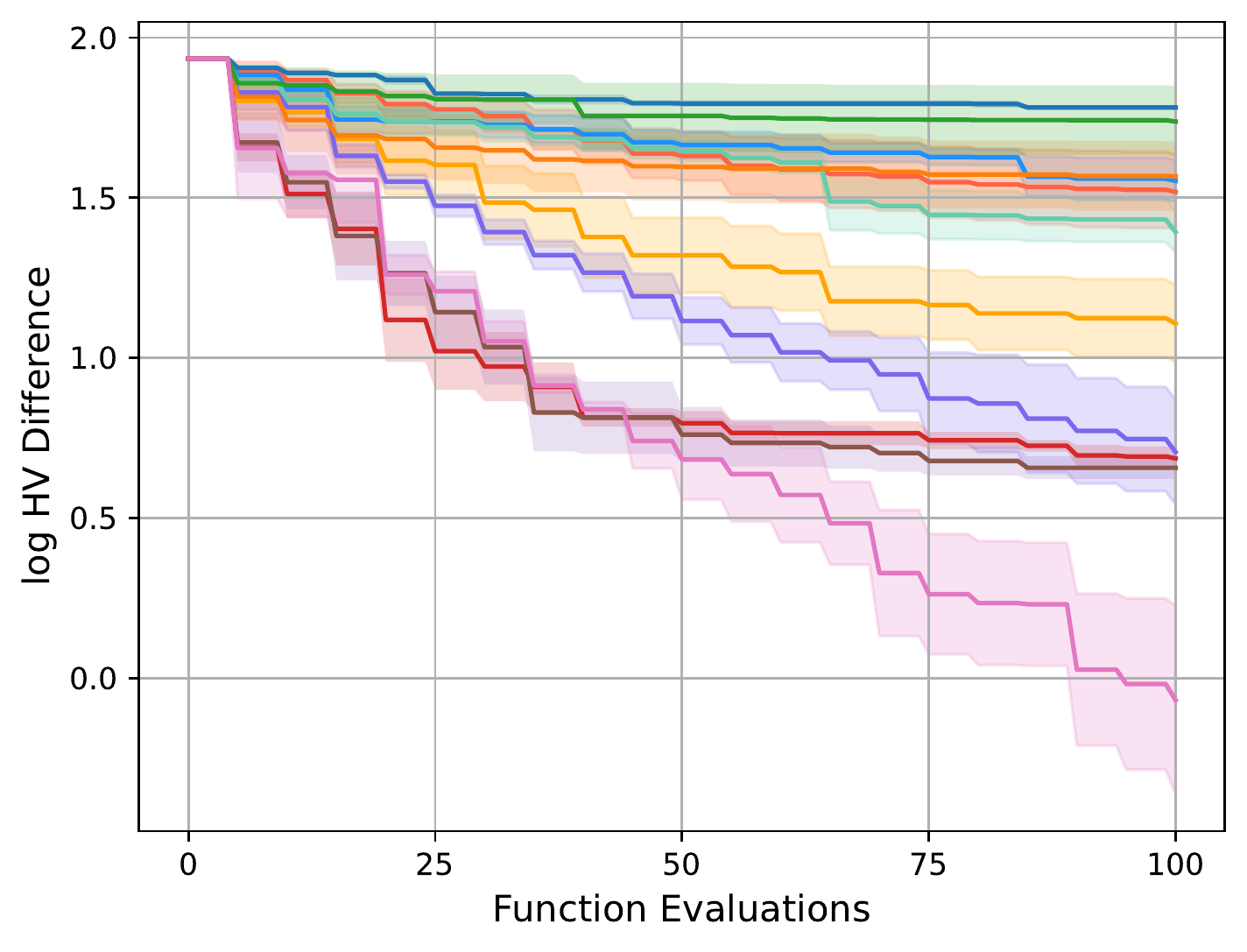}}\hfill
\subfloat[Rocket Injector Design]{\includegraphics[width = 0.33\linewidth]{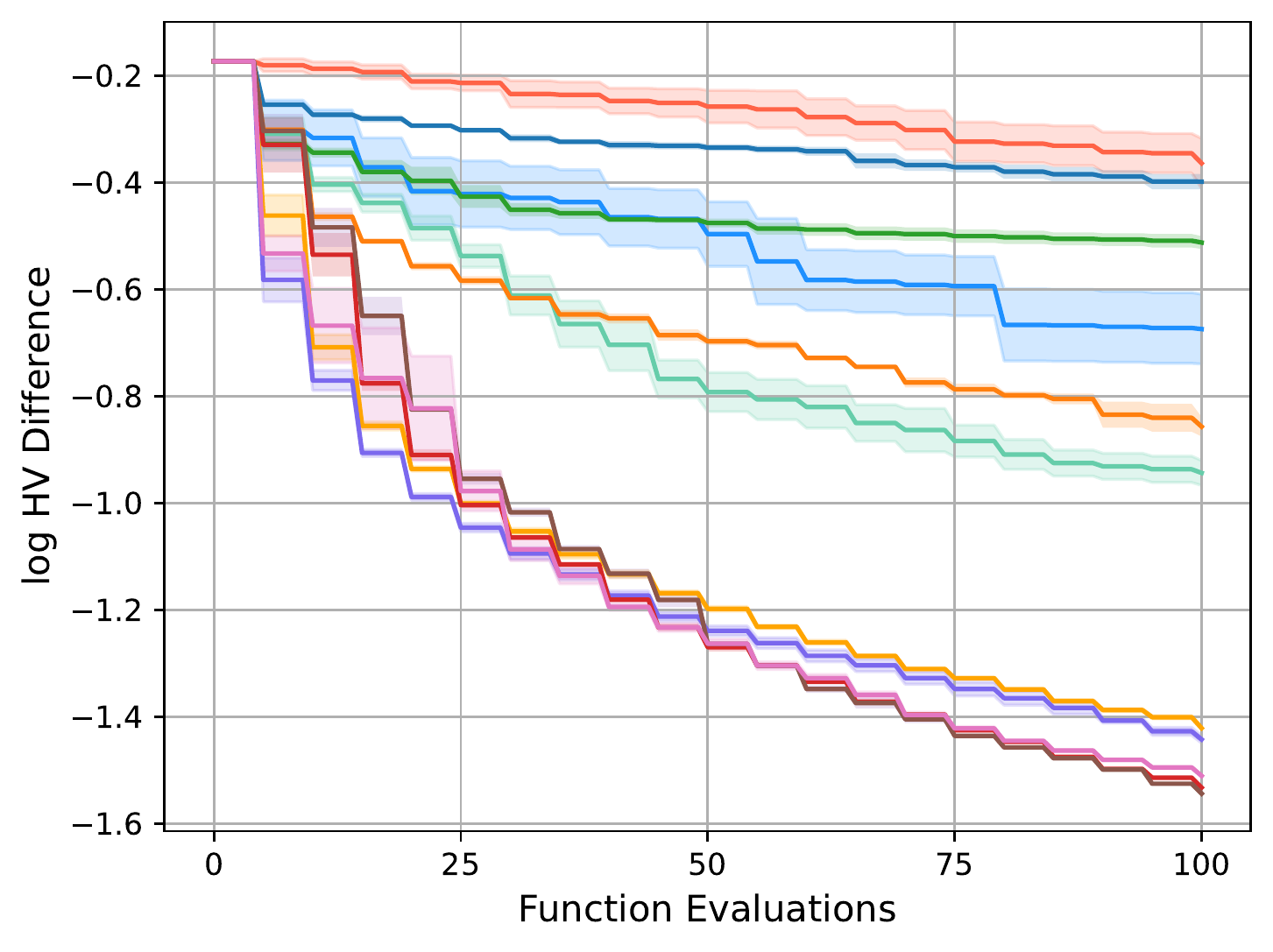}}
\caption{The log hypervolume difference w.r.t. the number of expensive evaluation of all algorithms for $15$ different problems. The solid line is the mean value averaged over $10$ independent runs for each algorithm, and the shaded region is the standard deviation around the mean value. \textbf{The labels of all algorithms can be found in Subfigure (a).}}
\label{fig_hv_trend}
\end{figure*}

\textbf{MOBO Performance.} We compare PSL with other MOBO methods on the performance of evaluated solutions. Figure~\ref{fig_hv_trend} shows the log hypervolume difference to the true/approximate Pareto front for the synthetic/real-world problems during the optimization process. The approximate Pareto fronts for the real-world design problems are from \citet{tanabe2020easy} with a large number of evaluations, which are also used in other MOBO works. In most experiments, our proposed PSL method has better or comparable performance with other MOBO algorithms. Especially, as a generalized scalarization-based method, PSL significantly outperforms the model-free counterparts such as qParEGO~\citep{knowles2006parego, daulton2021parallel}, MOEA/D-EGO~\citep{zhang2010expensive}, and TS-TCH~\citep{paria2020flexible}. These promising results validate the efficiency and usefulness of Pareto set learning for MOBO. More discussion of the proposed algorithm can be found in Appendix~\ref{subsec_supp_motivation} and Appendix~\ref{subsec_supp_hypervolume_improvement}.

As shown in Table~\ref{table_run_time}, PSL has a shorter or comparable total runtime (e.g., for modeling and batch selection) per iteration with other MOBO methods, which can be ignored in the expensive optimization problems (might take days). The algorithm runtimes for all problems can be found in Appendix~\ref{subsec_supp_runtime}. These results confirm that the Pareto set learning approach has a low computational overhead which is affordable for MOBO.  

\begin{figure}[t]
\captionsetup[subfigure]{font=scriptsize,labelfont=scriptsize}
\centering
\vspace{-0.15in}
\subfloat[F1 ($3.4 \times 10^{-3}$)]{\includegraphics[width = 0.22\linewidth]{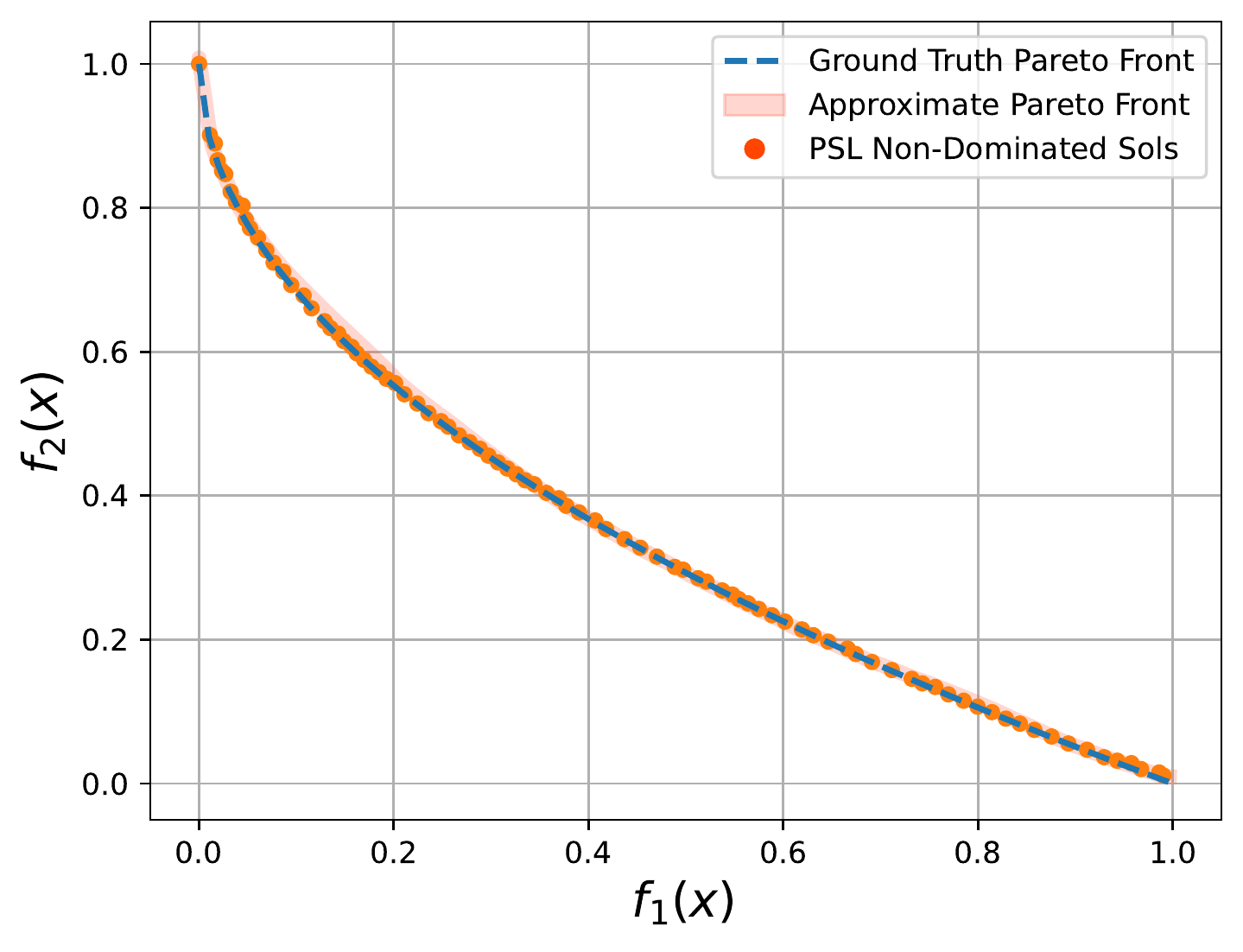}}
\hfill
\subfloat[F2 ($2.1 \times 10^{-3}$)]{\includegraphics[width = 0.22\linewidth]{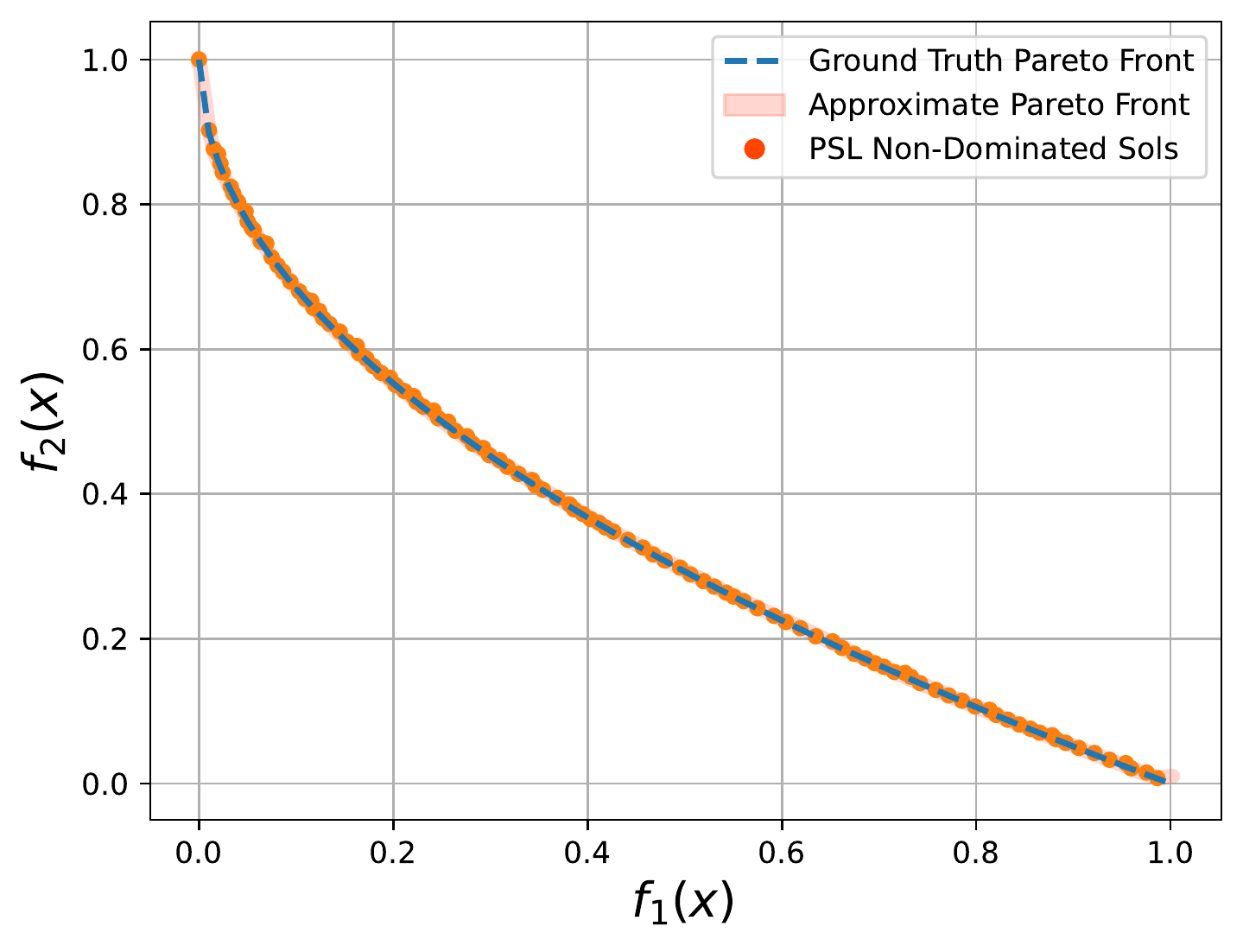}}
\hfill
\subfloat[VLMOP2 ($2.3 \times 10^{-5}$)]{\includegraphics[width = 0.22\linewidth]{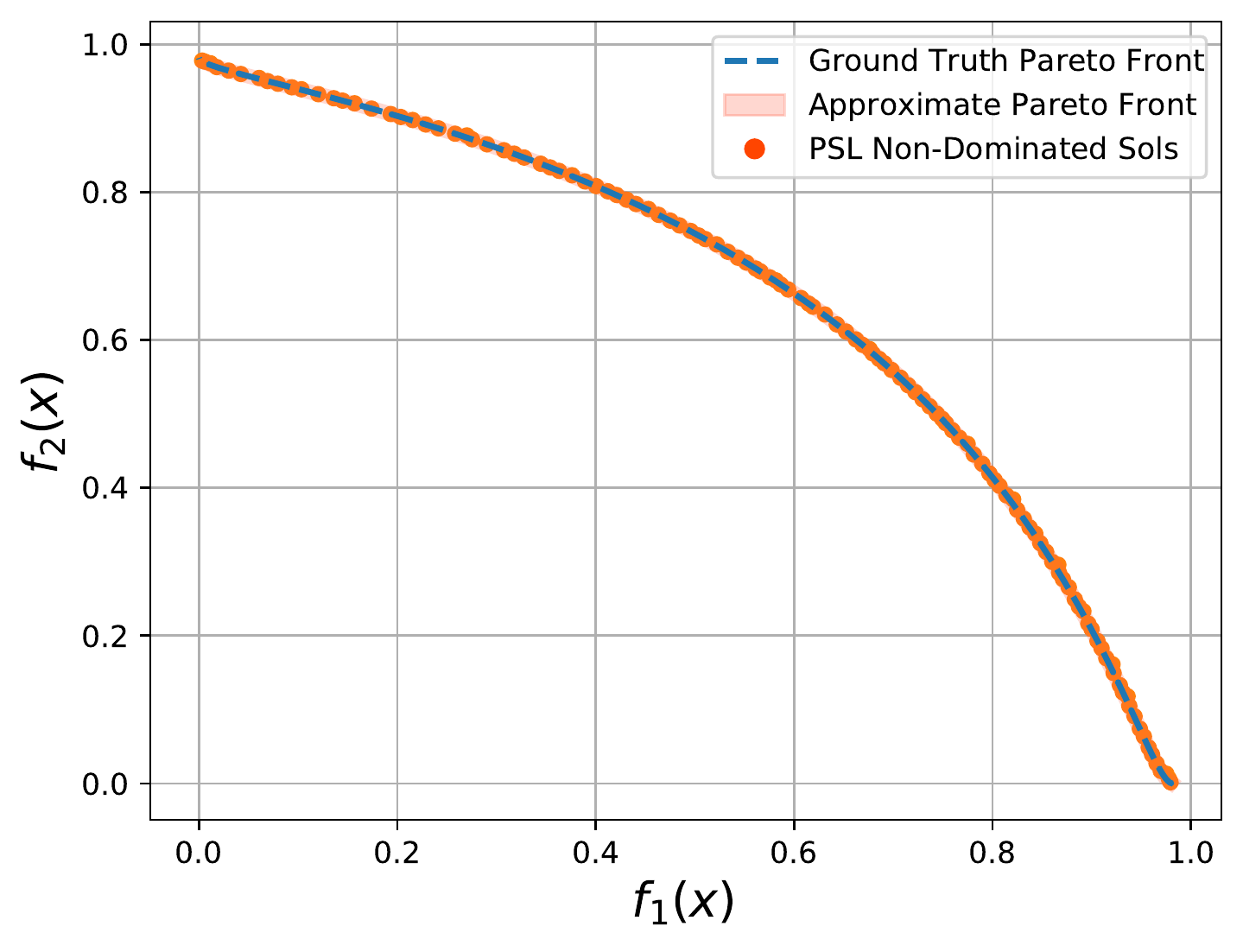}}
\hfill
\subfloat[DTLZ2 ($1.5 \times 10^{-5}$)]{\includegraphics[width = 0.22\linewidth]{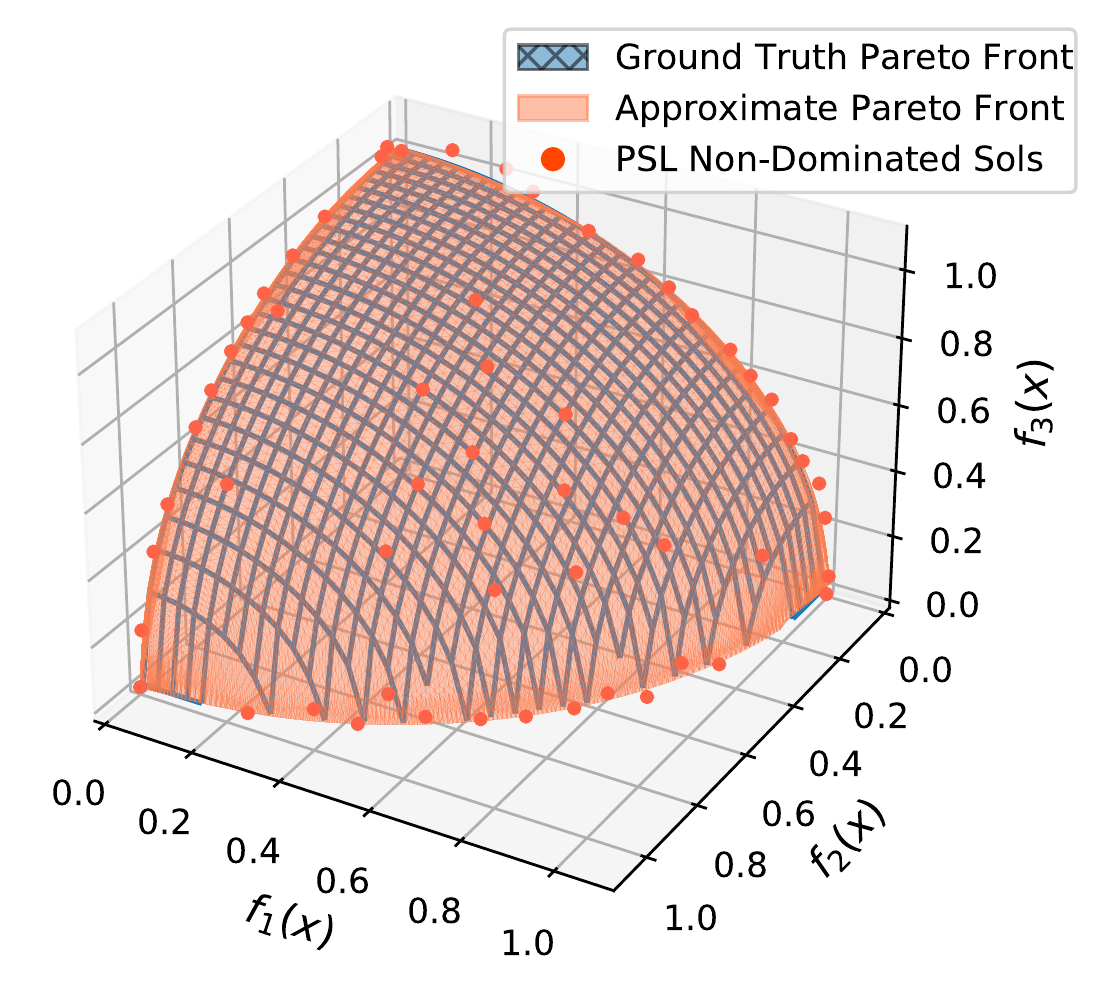}} \\
\vspace{-0.15in}
\subfloat[Four Bar Truss($2 \times 10^{-4}$)]{\includegraphics[width = 0.22\linewidth]{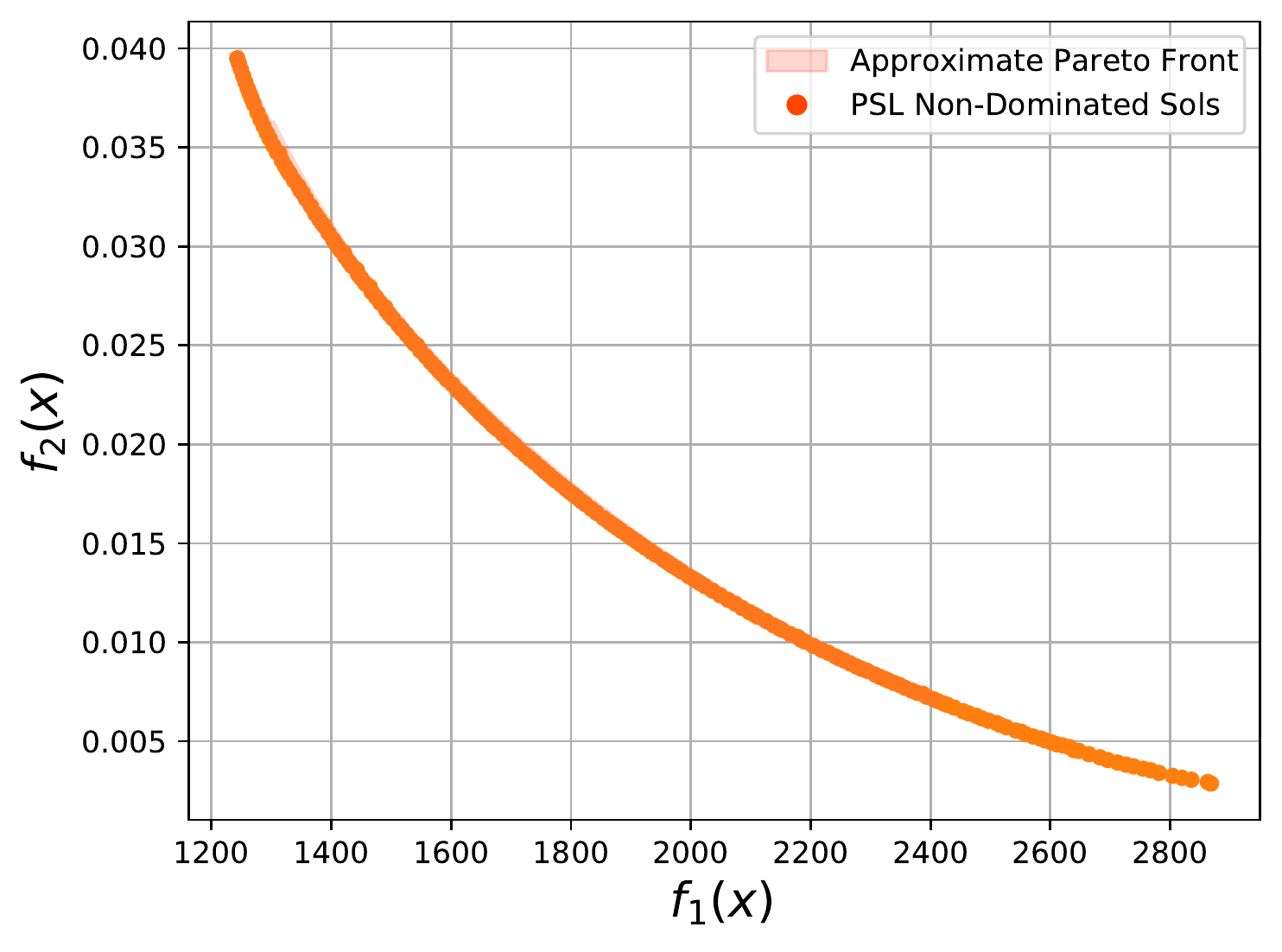}}
\hfill
\subfloat[Disk Brake ($7.7 \times 10^{-3}$)]{\includegraphics[width = 0.22\linewidth]{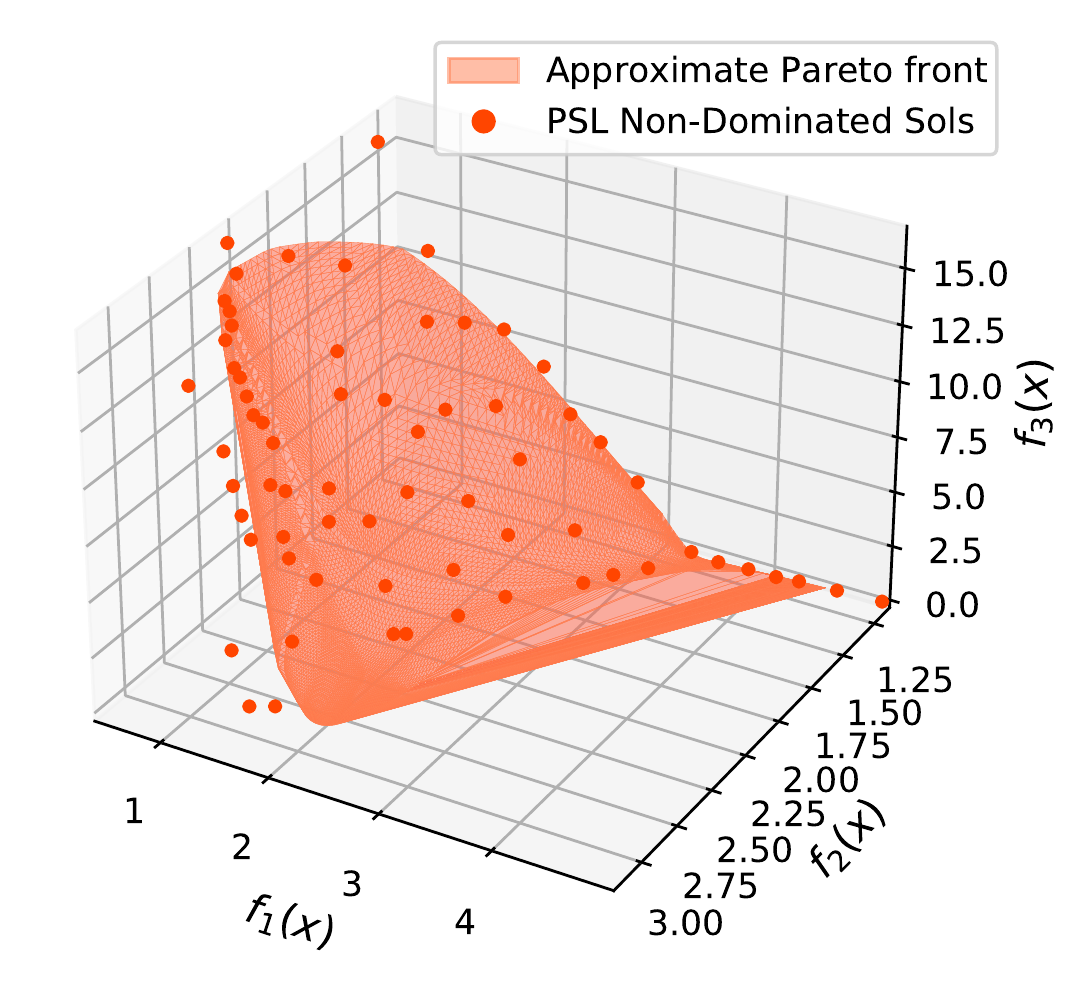}}
\hfill
\subfloat[Gear Train ($2.4 \times 10^{-4}$)]{\includegraphics[width = 0.22\linewidth]{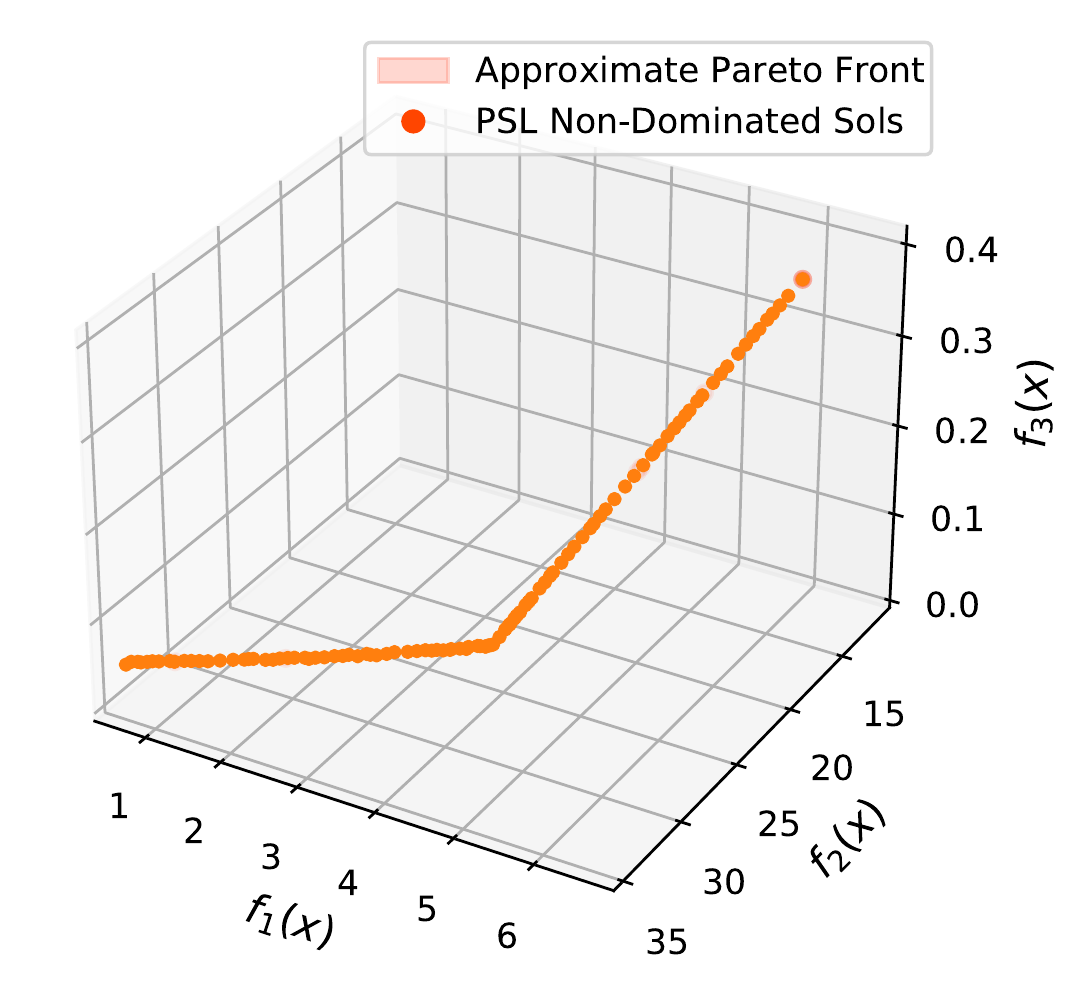}}
\hfill
\subfloat[Rocket Injector ($5 \times 10^{-4}$)]{\includegraphics[width = 0.22\linewidth]{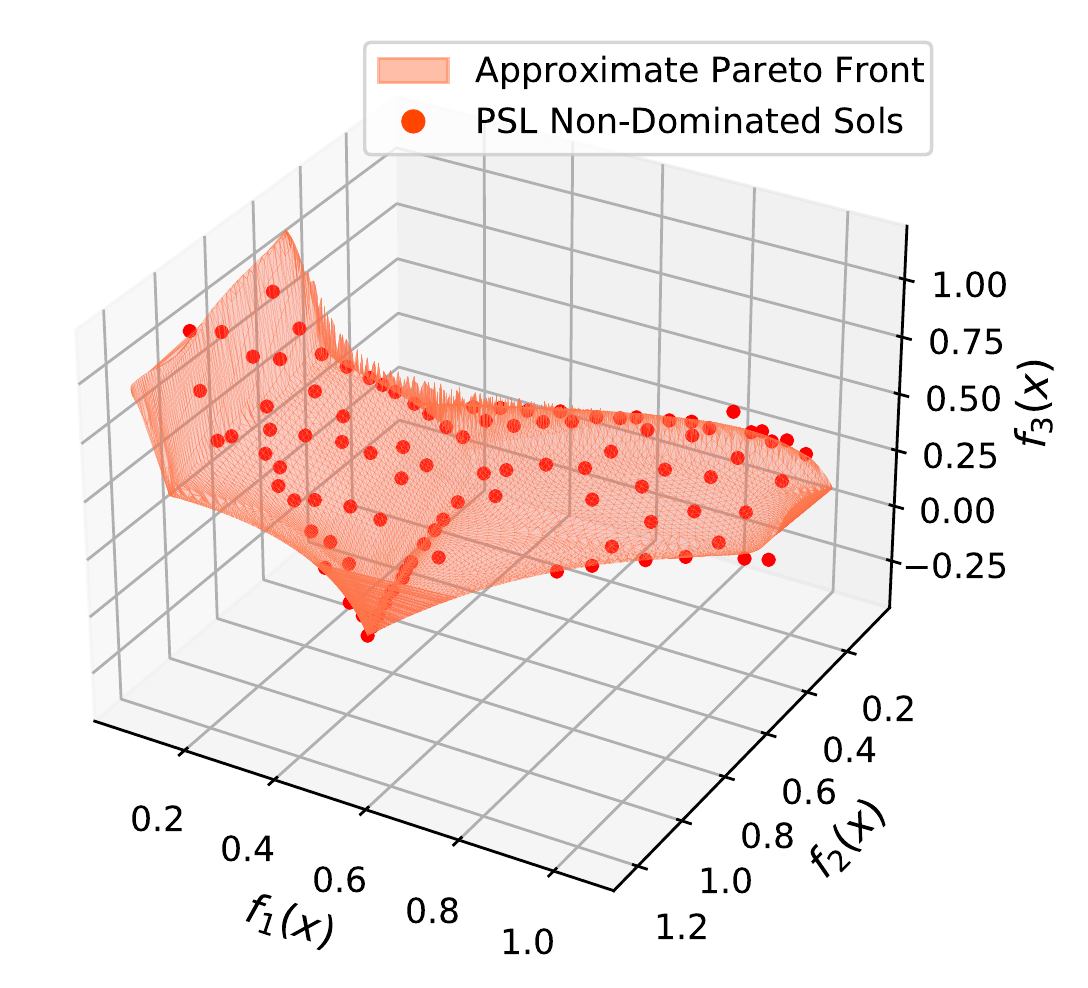}}
\caption{\textbf{The Learned Pareto Fronts (Relative Hypervolume Difference) by PSL:} Our learned Pareto fronts can match the ground truth Pareto fronts for the synthetic benchmarks, and have small relative hypervolume differences to the approximate Pareto fronts for real-world design problems. The learned Pareto front can well represent the optimal trade-offs among different objectives and provide valuable information to support flexible decision-making.} 
\label{fig_exp_pf_manifold}
\vspace{-0.2in}
\end{figure}

\textbf{The Learned Pareto Set.} We present the approximate Pareto set learned by PSL under the posterior mean after optimization in Figure~\ref{fig_exp_pf_manifold}, which is not supported by other MOBO methods. According to the results, PSL can successfully learn the Pareto sets for different benchmarks and real-world application problems with different shapes of Pareto fronts. For benchmark problems, PSL can match the ground truth Pareto front with a small evaluation budget. For real-world applications, the approximate Pareto fronts can capture the trade-off among objectives and provide valuable information to support decision-making. We further discuss the the practicality of the approximate Pareto set in Appendix~\ref{subsec_supp_practicality}.

\clearpage

\clearpage

\begin{wrapfigure}{R}{0.45\linewidth}
\vspace{-0.1in}
\centering
\subfloat[Different Preferences]{\includegraphics[width = 0.5\linewidth]{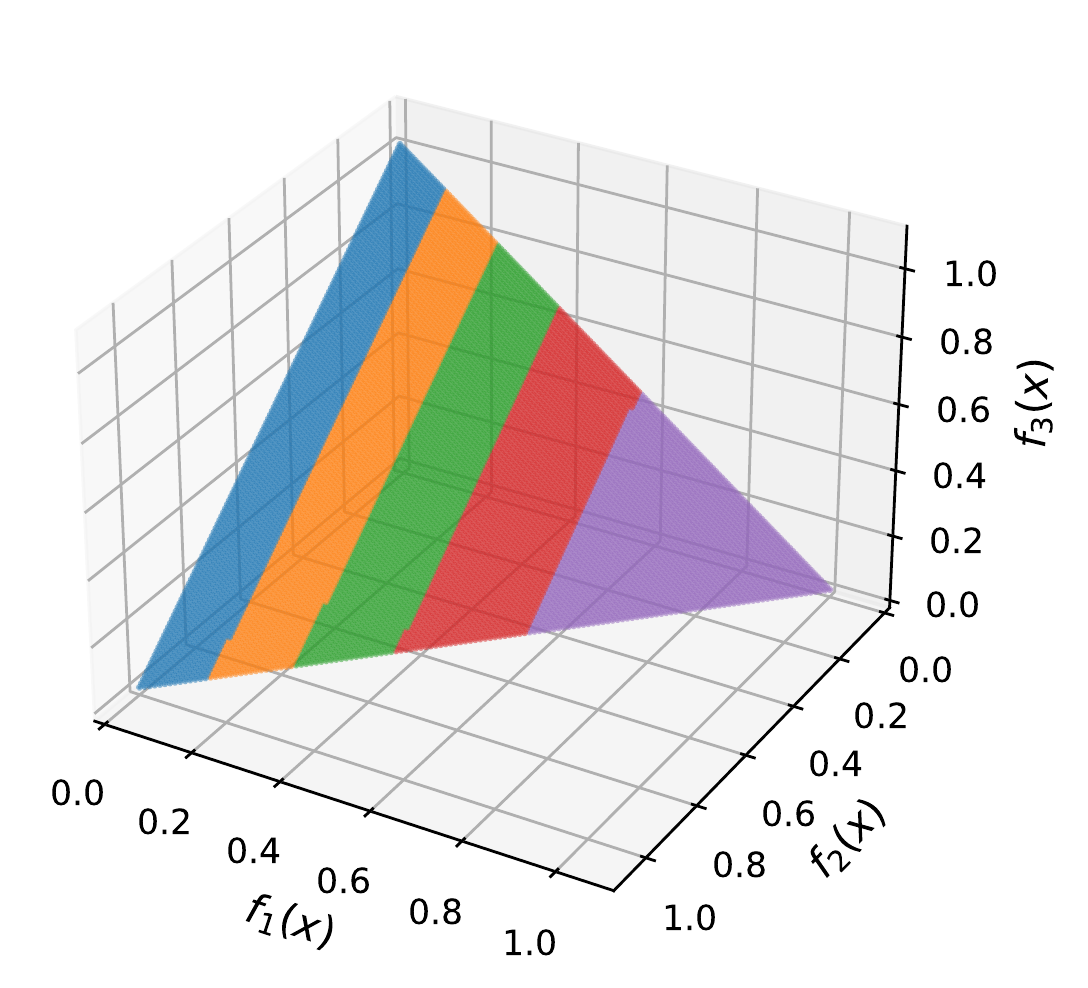}} \hfill
\subfloat[DTLZ2]{\includegraphics[width = 0.5\linewidth]{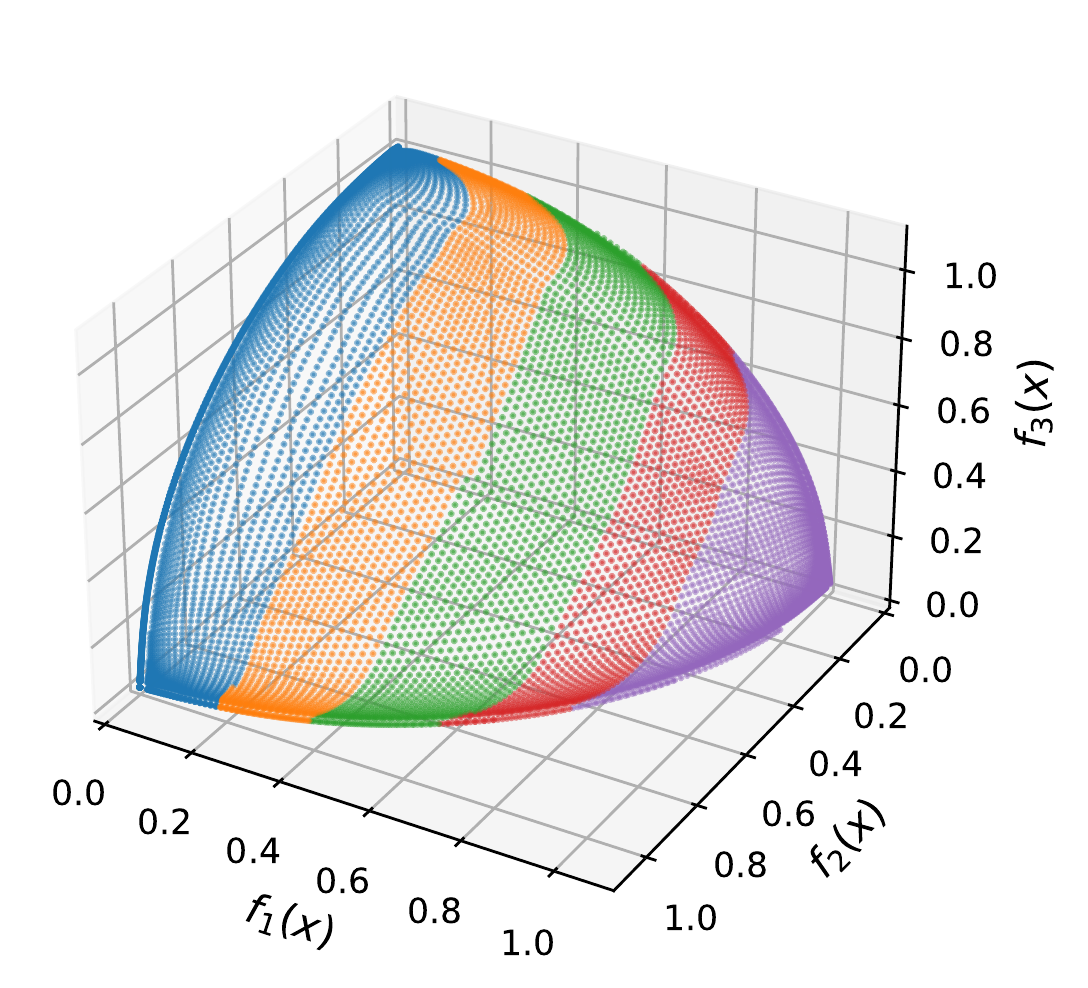}}\hfill
\caption{Different trade-off preferences and their corresponding solutions on the approximate Pareto set.}
\label{fig_psl_preference}
\vspace{-0.1in}
\end{wrapfigure}

\textbf{Flexible Trade-off Adjustment.} With our model, the decision-makers can easily explore the whole approximate Pareto set by themselves to select the most preferred trade-off solution(s) as shown in Figure~\ref{fig_psl_preference}. No time-consuming communication between the optimization modeler and the decision-maker is required. By directly exploring the approximate Pareto front in an interactive manner, the decision-makers can observe and understand the connection between the trade-off preferences and corresponding solutions in real-time. It is also beneficial for decision-makers to further adjust and assign their most accurate preferences. The ability to incorporate user's knowledge into decision making~\cite{garnett2022bayesian} could be crucial for many real-world applications. More experimental results and analyses can be found in Appendix~\ref{sec_supp_experiment}.

\section{Conclusion, limitation and future work}
\label{sec_conclusion}

\paragraph{Conclusion.} We have proposed a novel Pareto set learning method, which is a first attempt to approximate the whole Pareto set for expensive multi-objective optimization. The advantages of this approach are two-fold. First, by learning and utilizing the approximate Pareto set, it can serve as an efficient MOBO method that outperforms different existing approaches. Secondly, it allows decision-makers to readily explore the whole approximate Pareto set, which supports flexible and interactive decision-making. We believe the proposed Pareto set learning method could provide a novel way for solve expensive multi-objective optimization.  

\paragraph{Limitation and Future Work.} The quality of the approximate Pareto set mainly depends on the accuracy of the surrogate models and the performance of the Pareto set learning algorithm, which could be poor for problems with insufficient evaluation budget and/or large-scale search space. A more detailed discussion of limitations and potential future work can be found in Appendix~\ref{sec_supp_limitation}, and potential societal impact can be found in Appendix~\ref{sec_supp_impact}.

\subsubsection*{Acknowledgements}
This work was supported by the Hong Kong General Research Fund (11208121, CityU-9043148).



\clearpage

\bibliography{ref_multiobjective_optimization, ref_bayesian_optimization, ref_combinatorial_optimization, ref_multi_task_learning}
\bibliographystyle{abbrvnat}

\clearpage
\section*{Checklist}

\begin{enumerate}

\item For all authors...
\begin{enumerate}
  \item Do the main claims made in the abstract and introduction accurately reflect the paper's contributions and scope?
    \answerYes{}
  \item Did you describe the limitations of your work?
    \answerYes{See Appendix~\ref{sec_supp_limitation}.}
  \item Did you discuss any potential negative societal impacts of your work?
     \answerYes{See Appendix~\ref{sec_supp_impact}.}
  \item Have you read the ethics review guidelines and ensured that your paper conforms to them?
    \answerYes{}
\end{enumerate}

\item If you are including theoretical results...
\begin{enumerate}
  \item Did you state the full set of assumptions of all theoretical results?
    \answerNA{}
        \item Did you include complete proofs of all theoretical results?
    \answerNA{}
\end{enumerate}

\item If you ran experiments...
\begin{enumerate}
  \item Did you include the code, data, and instructions needed to reproduce the main experimental results (either in the supplemental material or as a URL)?
    \answerYes{See \url{https://github.com/Xi-L/PSL-MOBO}}
  \item Did you specify all the training details (e.g., data splits, hyperparameters, how they were chosen)?
    \answerYes{See Section~\ref{sec_experiment} and Appendix~\ref{sec_supp_problem}.}
        \item Did you report error bars (e.g., with respect to the random seed after running experiments multiple times)?
    \answerYes{}
        \item Did you include the total amount of compute and the type of resources used (e.g., type of GPUs, internal cluster, or cloud provider)?
    \answerYes{See Appendix~\ref{subsec_supp_runtime}.}
\end{enumerate}

\item If you are using existing assets (e.g., code, data, models) or curating/releasing new assets...
\begin{enumerate}
  \item If your work uses existing assets, did you cite the creators?
    \answerYes{See Section~\ref{sec_experiment}.}
  \item Did you mention the license of the assets?
    \answerYes{See Appendix~\ref{sec_supp_licenses}.}
  \item Did you include any new assets either in the supplemental material or as a URL?\answerNA{}
  \item Did you discuss whether and how consent was obtained from people whose data you're using/curating?
  \answerNA{}
  \item Did you discuss whether the data you are using/curating contains personally identifiable information or offensive content?
    \answerNA{}
\end{enumerate}

\item If you used crowdsourcing or conducted research with human subjects...
\begin{enumerate}
  \item Did you include the full text of instructions given to participants and screenshots, if applicable?
    \answerNA{}
  \item Did you describe any potential participant risks, with links to Institutional Review Board (IRB) approvals, if applicable?
    \answerNA{}
  \item Did you include the estimated hourly wage paid to participants and the total amount spent on participant compensation?
    \answerNA{}
\end{enumerate}

\end{enumerate}

\clearpage

\appendix


\appendix

We provide more discussion, details of the proposed algorithm and problem, and extra experimental results in this appendix:

\begin{itemize}
    \item More discussions of the proposed algorithm are provided in Section~\ref{sec_supp_discussion}.
    \item The limitations and potential future improvement for PSL are discussed in Section~\ref{sec_supp_limitation}.
    \item Potential societal impact of this work is discussed in Section~\ref{sec_supp_impact}.
    \item Details of the set model and batch selection algorithm are provided in Section~\ref{sec_supp_model}.
    \item Details of the benchmark and real-world application problems are given in Section~\ref{sec_supp_problem}. 
    \item More experimental results and analyses are presented in Section~\ref{sec_supp_experiment}. 
\end{itemize}

\section{More discussion}
\label{sec_supp_discussion}

\subsection{Motivation: Pareto set model over simple scalarization}
\label{subsec_supp_motivation}

Our proposed Pareto set learning method is closely related to the scalarization-based methods. But it can overcome a major disadvantage of the current scalarization methods for expensive optimization. 

The scalarization methods randomly select one (e.g., ParEGO~\citep{knowles2006parego}) or a batch of scalarized subproblems (e.g., MOEA/D-EGO~\citep{zhang2010expensive}, TS-TCH~\citep{paria2020flexible}, and qParEGO~\citep{daulton2021parallel}) at each iteration. By optimizing the acquisition function (e.g., EI, LCB, Thompson Sampling) on each scalarization, they generate a batch of solutions for expensive evaluation.  

A major limitation of all these scalarization methods is that they do not explicitly consider those already-evaluated solutions, neither for choosing the weighted scalarization, nor for selecting the next (batch of) solution(s) for expensive evaluation. Therefore, the selected weighted scalarization(s) might be close to those already-evaluated solutions. In other words, even the obtained solution(s) can maximize the EI/LCB for the selected scalarization, they could still be similar to those already-evaluated ones, and indeed not an optimal choice for multi-objective optimization. This limitation leads to inferior performance of these scalarization methods, as reported in \citep{daulton2020differentiable, lukovic2020diversity} and in our experimental results.

To overcome this limitation, our proposed method has a two-stage approach: 
\begin{itemize}
\item Stage 1: Using the Pareto set model, it first efficiently samples a dense set of candidate solutions to cover the whole approximate Pareto front (of posterior mean, EI, or LCB etc).
\item Stage 2: Then it selects a small batch of appropriate solutions from this dense set for expensive evaluation. For selection, we use the HVI criteria to take both the selected and already-evaluated solutions into consideration.
\end{itemize}
In this way, our proposed method can efficiently explore the whole approximate Pareto front, and choose the most appropriate solutions (on the approximate Pareto front, while far from the selected and already-evaluated ones) for expensive evaluation. The experimental results have validated the efficiency of the proposed method.

\subsection{Hypervolume improvement for batch selection}
\label{subsec_supp_hypervolume_improvement}

Efficiently finding a batch of solutions to optimize the EI/LCB of HVI could be very challenging. qEHVI~\citep{daulton2020differentiable} and qNEHVI~\citep{daulton2021parallel} are two promising approaches along this direction. From the viewpoint of optimizing HVI, our proposed method restricts the search procedure only on the approximate Pareto set, which is a low-dimensional (e.g., $(m-1)$-dimensional) manifold in the decision space~\citep{hillermeier2001generalized, zhang2008rm}. Therefore, we can use a simple two-stage sample-then-select approach to find the batch of solutions for evaluations. For the optimal situation, the set of solutions that optimize HVI should all be on the Pareto front. However, if an efficient method exists, directly optimizing the EI/LCB of HVI on the whole search space should be a principled approach for batch selection.  

On the other hand, the scalarization-based approach could have a close relationship with the hypervolume~\cite{zhang2020random}. It will be very interesting to study how to better leverage this relation for designing a more efficient algorithm, such as learning the whole Pareto set while inference the location of solutions (on the learned Pareto set) to optimize HVI at a single stage. These will be our future work. 

In summary, the advantages of PSL can be summarized from the following two viewpoints: 

\begin{itemize}
    \item From the viewpoint of scalarization based methods, PSL proposes a novel two-stage approach to select a small batch of appropriate solutions from the approximate Pareto front, which takes those already-evaluated solutions into consideration. As a result, it significantly outperforms other scalarization-based methods.
    
    \item From the viewpoint of hypervolume improvement based methods, in an ideal case, the set of solutions that optimize HVI should all be on the Pareto front. Restricting the search on the approximate Pareto set, PSL leads to an efficient HVI algorithm.
    
\end{itemize}

\subsection{Practicality of the approximate Pareto set}
\label{subsec_supp_practicality}

The learned approximate Pareto set could not always be accurate in real-world applications. Therefore, it is risky to only rely on the approximate Pareto set to make a decision (see more discussion in Section.~\ref{sec_supp_limitation}). With our proposed method, decision-makers can simultaneously have the evaluated solutions and the approximate Pareto set. The approximate Pareto set can provide extra useful information to better support their decision-making such as:  

\begin{itemize}
    \item It can help decision-makers to better understand the (approximate) trade-offs they will make for choosing any already evaluated solution. The approximate Pareto set and the corresponding surrogate objective values around the chosen solution can provide valuable information to understand what we can gain or lose by adjusting the chosen solution.
    
    \item It allows decision-makers to explore the whole approximate Pareto set easily. If none of the already evaluated solutions can satisfy the decision-maker's preferred trade-off, the decision-maker can rely on the approximate Pareto set/front to choose the preferred solution for further evaluation.
    
    \item When the optimization modeler and the final decision-maker are not the same person, the approximate Pareto set/front provides a much more efficient way for them to communicate and discuss the whole trade-offs among different objectives during or after the optimization process. This demand is common and essential for many applications~\citep{malkomes2021beyond}. 
    
    \item Finally, an important application for MOBO is to help domain experts efficiently conduct experiments. The approximate Pareto set might contain useful patterns and structures which can help the domain expert obtain more information from the experiment. It also provides a way for domain experts to incorporate their knowledge into the optimization process (e.g., choose the most concerned region, and eliminate some uninteresting locations). The proposed Pareto set learning method could be a novel approach to support the "bringing decision-makers back into the loop" approach for Bayesian optimization~\cite{garnett2022bayesian}.
\end{itemize}

Therefore, the learned approximate Pareto set could be a useful tool to support flexible decision-making for expensive multi-objective optimization. 

\section{Limitation and potential future work}
\label{sec_supp_limitation}

\subsection{The approximate Pareto set could be inaccurate}

It could be very risky, especially in those safety-critical applications, to only rely on an approximate Pareto set to make the final decision. The quality of the approximate Pareto set heavily depends on the quality of the surrogate models. We cannot obtain a good approximate Pareto set if the evaluation budget is insufficient to build a good set of surrogate models. This is also a general challenge for (multi-objective) Bayesian optimization~\citep{garnett2022bayesian}.

One possible method to address this challenge is to leverage the user preference information into the expensive optimization process~\citep{abdolshah2019multi, astudillo2020multi, paria2020flexible}. In this way, the general MOBO method can spend the limited evaluation budget mainly on the user-preferred region rather than the whole decision space. Similarly, our proposed method can only build a partial approximate Pareto set with the user-preferred trade-offs. However, the preference-based approach could be affected by the following limitation.

\subsection{Defining the user preference is challenging for black-box optimization}

Although some scalarization methods (e.g., Chebyshev scalarization) have good theoretical properties to connect the scalars to their corresponding Pareto solutions~\citep{choo1983proper}, it could still be hard to define user preferences in terms of scalars, especially in the black-box expensive optimization setting.

In many applications, the decision-makers might not even know their actual preferences before making their decision. Suppose the approximate Pareto set can be learned appropriately, instead of asking users to provide their preferences, our proposed approach can let them interactively explore the whole approximate Pareto set/front to select their most preferred solutions. In this interactive way, it could be much easier for the decision-makers to accurately express and assign their preferences. However, it could be difficult to precisely obtain user preference if a good approximate Pareto set is unavailable (e.g., in the early stage of optimization, or without enough budget).

The algorithm proposed in \citet{abdolshah2019multi} is an elegant and promising way to incorporate preference into the MOBO process via the preference-order constraints, which does not require any prior knowledge on the (approximate) Pareto front. The preference-order constraint also has a close relationship with the lexicographic approach for multi-objective optimization~\citep{romero2001extended}, which is a non-scalarizing method with good theoretical property (e.g., connection to weakly Pareto optimal solution). Incorporating this information into our proposed Pareto set learning model for more flexible preference incorporation could be interesting for future work.

\subsection{Scalability}

The scalability of the search space dimension is a major challenge for general MOBO algorithms. It is also very difficult for our proposed PSL method to learn a good enough Pareto set for such large-scale problems. The PSL's performance depends on a set of good surrogate models, which requires a large number of evaluated solutions for the problem with a high-dimensional search space.

Recently, some efficient search region decomposition/management methods have been proposed to better allocate the limited computational budget for problems with high-dimensional search space~\citep{eriksson2019scalable, wang2020learning}. These ideas can be naturally generalized to solve expensive multi-objective optimization problems, such as MORBO~\citep{daulton2022multi} and LaMOO~\citep{zhao2022multi}. Since the search region management methods are algorithm-agnostic (e.g., a meta-algorithm), they can be combined with different (multi-objective) optimization algorithms. Studying how to efficiently combine the MORBO/LaMOO approach with our proposed PSL method will be an important future work to tackle this limitation.

\section{Potential societal impact}
\label{sec_supp_impact}

Our proposed Pareto set learning method mainly has two strengths: (1) it is good for better multi-objective Bayesian optimization in terms of speed and sample efficiency, and (2) it can help decision-makers to navigate the estimated Pareto set for better decision-making. These strengths could lead to many positive potential societal impacts. For example, as an efficient MOBO algorithm, it can reduce the cost of obtaining a set of diverse Pareto solutions in many applications, such as materials science, engineering design, and recommender systems. The learned approximate Pareto set provides a novel way for decision-makers to easily explore different trade-offs, which could be an important component for a more user-friendly multi-criteria decision system.   

On the negative side, as discussed in the limitation section, there is no guarantee that the learned approximate Pareto set could be accurate. Solely relying on the approximate Pareto set to make a decision could be risky. The decision-makers should leverage all the information they have to make the final decision. In addition, the leakage of the learned Pareto set model might unintentionally reveal the problem information and user preference, which should be avoided in real-world applications.

\newpage

\section{Model and algorithm details}
\label{sec_supp_model}

\subsection{Pareto set model}

\textbf{Model Structure.} In this work, we use a multi-layered perceptron (MLP) as the Pareto set model $\vx(\vlambda) = h_{\vtheta}(\vlambda)$. For all experiments, the Pareto set model has $2$ hidden layers each with $256$ hidden units, and the activation is ReLU. The input and output dimension of the set model is the number of objectives $m$ and the dimension of decision variables $n$, respectively. Therefore, the set model $h_{\vtheta}(\vlambda)$ has the following structure:
\begin{align}
h_{\vtheta}(\vlambda): &\text{Input } \vlambda \rightarrow \text{Linear}(m, 256) \rightarrow \text{ReLU} \rightarrow \text{Linear}(256, 256) \nonumber \\
&\rightarrow \text{ReLU} \rightarrow \text{Linear}(256, 256) \rightarrow \text{ReLU}  \\
&\rightarrow  \text{Linear}(256, n) \rightarrow \text{Output } \vx(\vlambda), \nonumber
\label{supp_eq_pml_model}
\end{align}
where the input is an $m$-dimensional preference $\vlambda \in \vLambda = \{\vlambda \in \bbR^m_{+}|\sum \lambda_i = 1\}$, the output is a $n$-dimensional decision variables $\vx \in \bbR^n$, and $\vtheta \in \bbR^d$ represents all learnable parameters for the Pareto set model. We build independent Gaussian process models with Mat\'{e}rn 5/2 kernel for each objective as the surrogate models. The objective to optimize is the augmented Tchebycheff scalarization on the surrogate value with respect to all valid preferences. 

\textbf{Model Training.} In the proposed Pareto set learning method, we have the following model structure:
\begin{align}
\vlambda  &\rightarrow \textbf{Pareto Set Model } \rightarrow \vx(\vlambda) = h_{\vtheta}(\vlambda) \\ &\rightarrow \textbf{GP Model} \rightarrow \hat \vf(\vx) \rightarrow  \textbf{TCH Scalarization} \rightarrow \hat g_{\text{tch\_aug}}(\vx(\vlambda)),
\label{supp_eq_psl_framework}
\end{align}
where the preference $\vlambda$ goes through the Pareto set model $h_{\vtheta}(\vlambda)$ and Gaussian process model, then gets the augmented Tchebycheff scalarized value $\hat g_{\text{tch\_aug}}(\vx(\vlambda))$ at the end (a detailed version can be found in Figure~\ref{fig_model_PSL} in the main paper). To train the Pareto set model, we use a gradient-based method to optimize the model parameter $\vtheta$ with respect to the scalarized value $\hat g_{\text{tch\_aug}}(\vx(\vlambda))$. With the chain rule, we have:
\begin{eqnarray}
 \nabla_{\theta} \hat g_{\text{tch\_aug}}(\vx = h_{\vtheta}(\vlambda)|\vlambda) = \frac{\partial \hat g_{\text{tch\_aug}}}{\partial \vtheta} = \frac{\partial \hat g_{\text{tch\_aug}}}{\partial \hat \vf} \frac{\partial \hat \vf}{\partial \vx} \frac{\partial \vx}{\partial \vtheta},
 \label{supp_eq_chain_rule}
\end{eqnarray}
where we can calculate each term respectively. 

\begin{itemize}
    \item For the first term $\frac{\partial \hat g_{\text{tch\_aug}}}{\partial \hat \vf}$, the augmented Tchebycheff scalarization $\hat g_{\text{tch\_aug}}(\vx|\vlambda) = \max_{1 \leq i \leq m} \{ \lambda_i (\hat f_i(\vx) - (z^*_i - \varepsilon)) \}  + \rho \sum_{i=1}^{m} \lambda_i \hat f_i(\vx)$ has a max operator which is technically only subdifferentiable, but it is known to have good subgradients~\citep{wilson2017reparameterization} for surrogate model based optimization. We simply take the subgradients for this term.
    \item The second term $\frac{\partial \hat \vf}{\partial \vx}$ is the gradient of surrogate values to the input $\vx$. In this work, we build independent Gaussian process models for each objective with Mat\'{e}rn 5/2 kernel. The gradients of the kernel $\frac{\partial \vk_i}{\partial \vx}$, predictive mean $\frac{\partial \hat \mu_i}{\partial \vx}$ and standard deviation $\frac{\partial \hat \sigma_i}{\partial \vx}$ for each objective can be easily calculated. With these terms, the calculation for the surrogate value (e.g., predictive mean, UCB and EI) for each objective $\frac{\partial \hat f_i}{\partial \vx}$ is also straightforward.
    \item The third term $\frac{\partial \vx}{\partial \vtheta} = \frac{\partial h_{\vtheta}(\vx)}{\partial \vtheta}$ is the gradient of the Pareto set model to the parameter $\vtheta$. In this work, we use a simple MLP as the Pareto set model, and its gradient be easily obtained by auto-differentiation such as in PyTorch~\cite{paszke2019pytorch}.
\end{itemize}

For \textbf{Algorithm~\ref{alg_pml_gp}} in the main paper, we randomly sample $K = 10$ different preferences $\{\vlambda_k\}_{k=1}^{K = 10} \sim \vLambda$ in batch at each step. Without any prior information (e.g., user's preference), we uniformly sample preference $\boldsymbol{\lambda}$ from $[0,1]^{m}$ and then normalize it such that $\sum_{i=1}^{m} \boldsymbol{\lambda}_i = 1$. The total update step is $T = 1000$ for training the Pareto set model at each iteration. The model optimizer is Adam with learning rate $\eta = 1\text{e}-3$ and no weight decay. The learning process typically only requires less than $10$ seconds to finish, which is enough to obtain a good Pareto set approximation. Detailed experimental results on the runtime can be found in Table~\ref{table_alg_runtime}. 

We use the same model and training procedure for all experiments, and the only problem-dependent setting is the input/output dimension $m$ and $n$. The proposed Pareto set model and its learning approach have robust and promising performances for different expensive multi-objective optimization benchmarks and real-world problems.

\subsection{Batch selection with Pareto set learning}
\label{subsec_supp_batch}

\setcounter{algorithm}{3}
\begin{algorithm}[ht]
	\caption{Sampling-based Greedy Batch Selection}
	\label{supp_alg_greedy_selection}
 	\begin{algorithmic}[1]
    	\STATE \textbf{Input:} Evaluated Solutions $\{\vX_{t-1}, \vy_{t-1}\}$, Candidate Solutions $\vX_C = \vX$, Surrogate Models
    	\STATE Initialize the selected solution set $\vX_B = \varnothing$ with predicted values $\hat \vy_B = \varnothing$
		\FOR{$t = 1$ to $B$}
		    \STATE Calculate predicted HVIs for each individual solution in $\vX_C$ with respective to $\vy_{t-1} \cup \hat \vy_B$
		    \STATE Choose the single solution $\vx_s$ with the best predicted HVI
		    \STATE $\vX_B \leftarrow \vX_B \cup \{\vx_s\}$, $\hat \vy_B \leftarrow \hat \vy_B \cup \{\hat \vf(\vx_s)\}$, $\vX_C \leftarrow \vX_C \setminus \{\vx_s\}$
		\ENDFOR	
		\STATE \textbf{Output:} The Selected Solutions $\vX_B$.
 	\end{algorithmic}
\end{algorithm}

The MOBO with Pareto Set Learning framework (e.g., \textbf{Algorithm~\ref{alg_mobo_pml}}) is similar to other MOBO methods, and the main difference is on the batch selection with the learned Pareto set (\textbf{Algorithm~\ref{alg_pml_batch_selection}}). At each iteration, we first randomly sample $P = 1000$ preferences $\{\vlambda^{(p)}\}_{p=1}^{P = 1000} \sim \Lambda$ and directly obtain their corresponding solutions $\vX = \{\vx(\vlambda^{(p)})\}_{p=1}^{P = 1000}$ on the current learned Pareto set $\mathcal{M}_{\text{psl}}$. We then use the predicted hypervolume improvement (HVI) with respect to the already evaluated solutions $\{\vX_{t-1}, \vy_{t-1}\}$ to select a subset $\vX_B \subset \vX$ for expensive evaluation. The surrogate values for the candidate solutions are directly used for the predicted HVI calculation if they are the predicted mean $\hat \vmu$ or the lower confidence bound $\hat \vmu - \beta \vsigma$. When the surrogate values are expected improvement, although the approximate Pareto set is learned with EI for each scalarization, we let $\hat \vf(\vx_s) = \hat \vmu - \vsigma$ for the predicted HVI calculation. The main reasons for this choice are: 1) we want to avoid the (repeatedly) time-consuming Monte Carlo integration for calculating the expected hypervolume improvement; 2) the LCB (or the posterior mean only) is on the same scale as the value of those already-evaluated solutions, which make the calculation of HVI meaningful.  

Since it is computationally intensive to find $B$ solutions to exactly optimize HVI, we select the solution batch $\vX_B$ in a sequentially greedy manner from $\vX$. Similar approaches have also been used in the current MOBO methods~\citep{lukovic2020diversity,daulton2022multi}. The greedy selection approach is presented in \textbf{Algorithm~\ref{supp_alg_greedy_selection}}. We sequentially sample solutions from the candidate set $\vX_C$, and add the single solution $\vx_s$ with the best HVI value into the selected set $\vX_B$. For all problems, the acquisition search procedure (solution sampling + individual search + batch selection) with batch size $B = 5$ typically costs less than $3$ seconds. The detailed results can be found in Table~\ref{table_alg_runtime}. 

\clearpage

\section{Problem details}
\label{sec_supp_problem}

\begin{table*}[ht]
\centering
\caption{Problem information and reference point for both synthetic benchmarks and real-world engineering design problems.}
\label{supp_table_problem_refpoint}
\tabcolsep=0.12cm
\begin{tabular}{lccc}
\toprule
Problem                & n & m & Reference Point($\vr$)    \\ \midrule
F1                     & 6 & 2 & (1.1,1.1)                 \\
F2                     & 6 & 2 & (1.1,1.1)                 \\
F3                     & 6 & 2 & (1.1,1.1)                 \\
F4                     & 6 & 2 & (1.1,1.1)                 \\
F5                     & 6 & 2 & (1.1,1.1)                 \\
F6                     & 6 & 2 & (1.1,1.1)                 \\
VLMOP1                 & 1 & 2 & (4.4,4.4)                 \\
VLMOP2                 & 6 & 2 & (1.1,1.1)                 \\
VLMOP3                 & 2 & 3 & (11,66,1.1)               \\
DTLZ2                  & 6 & 3 & (1.1,1.1,1.1)             \\
Four Bar Truss Design  & 4 & 2 & (3175.0065, 0.0400)       \\
Pressure Vessel Design & 4 & 2 & (6437.2649, 1417536.7586) \\
Disk Brake Design      & 4 & 3 & (5.8374, 3.4412, 27.5)    \\
Gear Train Design      & 4 & 3 & (6.5241, 61.6, 0.3913)    \\
Rocket Injector Design & 4 & 3 & (1.0884, 1.0522, 1.0863)  \\  \bottomrule
\end{tabular}
\end{table*}

\subsection{Synthetic benchmark problems}
\label{subsec_supp_benchmark}

To better evaluate our proposed PSL method, we propose 6 new synthetic test problems with different shapes of Pareto sets which can be found in next page. We also test our proposed PSL algorithm on different widely-used synthetic multi-objective optimization benchmark problems, namely VLMOP1-3~\citep{van1999multiobjective} and DTLZ2~\citep{deb2002scalable}. The input and output dimensions of these problems are shown in Table~\ref{supp_table_problem_refpoint}. These synthetic problems have known Pareto sets $\mathcal{M}_{\text{ps}}$ and Pareto fronts $\vf(\mathcal{M}_{\text{ps}})$ with the nadir point: 
\begin{align}
&\vz_{\text{nadir}} = (\max f_1(\vx_1),\ldots, \max f_m(\vx_m)), \\
&\forall \vx_1, \vx_2, \ldots, \vx_m \in \mathcal{M}_{\text{ps}}, \nonumber
\label{supp_eq_nadir_point}
\end{align}
where $\vx_1, \ldots, \vx_m$ are solutions in the Pareto set. We set the reference point $\vr = 1.1 \times \vz_{\text{nadir}}$ for each problem.

\newpage

\begin{tabular}{|l|l|} 
\hline
F1 & 
\parbox{10cm}{\begin{align}
&f_1(\vx) = (1 + \frac{s_1}{|J_1|})\vx_1, \quad f_2(\vx) = (1 + \frac{s_2}{|J_2|})\left(1 - \sqrt{\frac{\vx_1}{1 + \frac{s_2}{|J_2|}}} \right) \nonumber \\
&\text{where } s_1 = \sum_{j \in J_1} (\vx_j - (2\vx_1 - 1)^2)^2 \text{ and } s_2 = \sum_{j \in J_2} (\vx_j - (2\vx_1 - 1)^2)^2, \nonumber \\
&J_1 = \{j|j \text{ is odd and } 2 \leq j \leq n\} \text{ and } J_2 = \{j|j \text{ is even and } 2 \leq j \leq n\}. \nonumber
\end{align}} \\ 
\hline
F2 &
\parbox{10cm}{\begin{align}
&f_1(\vx) = (1 + \frac{s_1}{|J_1|})\vx_1, \quad f_2(\vx) = (1 + \frac{s_2}{|J_2|})\left(1 - \sqrt{\frac{\vx_1}{1 + \frac{s_2}{|J_2|}}} \right) \nonumber  \\
&\text{where } s_1 = \sum_{j \in J_1} (\vx_j - \vx_1^{0.5(1.0 +\frac{3(j-2)}{n-2})})^2 \text{ and } s_2 = \sum_{j \in J_2} (\vx_j - \vx_1^{0.5(1.0 +\frac{3(j-2)}{n-2})})^2, \nonumber \\
&J_1 = \{j|j \text{ is odd and } 2 \leq j \leq n\} \text{ and } J_2 = \{j|j \text{ is even and } 2 \leq j \leq n\}. \nonumber
\end{align}} \\
\hline
F3 &
\parbox{10cm}{\begin{align}
&f_1(\vx) = (1 + \frac{s_1}{|J_1|})\vx_1, \quad f_2(\vx) = (1 + \frac{s_2}{|J_2|})\left(1 - \sqrt{\frac{\vx_1}{1 + \frac{s_2}{|J_2|}}} \right) \nonumber\\
&\text{where } s_1 = \sum_{j \in J_1} (\vx_j - \sin(4\pi\vx_1 + \frac{j\pi}{n}))^2 \text{ and } s_2 = \sum_{j \in J_2} (\vx_j - \sin(4\pi\vx_1 + \frac{j\pi}{n}))^2, \nonumber \\
&J_1 = \{j|j \text{ is odd and } 2 \leq j \leq n\} \text{ and } J_2 = \{j|j \text{ is even and } 2 \leq j \leq n\}. \nonumber
\end{align}} \\
\hline
F4 & 
\parbox{10cm}{\begin{align}
&f_1(\vx) = (1 + \frac{s_1}{|J_1|})\vx_1, \quad f_2(\vx) = (1 + \frac{s_2}{|J_2|})\left(1 - \sqrt{\frac{\vx_1}{1 + \frac{s_2}{|J_2|}}} \right) \nonumber\\
&\text{where } s_1 = \sum_{j \in J_1} (\vx_j - 0.8\vx_1 \cos(4\pi\vx_1 + \frac{j\pi}{n}))^2 \text{ and } s_2 = \sum_{j \in J_2} (\vx_j - 0.8\vx_1 \sin(4\pi\vx_1 + \frac{j\pi}{n}))^2, \nonumber \\
&J_1 = \{j|j \text{ is odd and } 2 \leq j \leq n\} \text{ and } J_2 = \{j|j \text{ is even and } 2 \leq j \leq n\}. \nonumber
\end{align}}\\
\hline
F5 &
\parbox{10cm}{$$\begin{aligned}
&f_1(\vx) = (1 + \frac{s_1}{|J_1|})\vx_1, \quad f_2(\vx) = (1 + \frac{s_2}{|J_2|})\left(1 - \sqrt{\frac{\vx_1}{1 + \frac{s_2}{|J_2|}}} \right), \\
&\text{where } s_1 = \sum_{j \in J_1} (\vx_j - 0.8\vx_1 \cos(\frac{4\pi\vx_1 + \frac{j\pi}{n}}{3}))^2 \text{ and } s_2 = \sum_{j \in J_2} (\vx_j - 0.8\vx_1 \sin(4\pi\vx_1 + \frac{j\pi}{n}))^2, \\
&J_1 = \{j|j \text{ is odd and } 2 \leq j \leq n\} \text{ and } J_2 = \{j|j \text{ is even and } 2 \leq j \leq n\}.
\end{aligned}$$} \\
\hline
F6 &
\parbox{10cm}{\begin{align}
&f_1(\vx) = (1 + \frac{s_1}{|J_1|})\vx_1, \quad f_2(\vx) = (1 + \frac{s_2}{|J_2|})\left(1 - \sqrt{\frac{\vx_1}{1 + \frac{s_2}{|J_2|}}} \right) \nonumber \\
&\text{where } s_1 = \sum_{j \in J_1} \{\vx_j - [0.3\vx_1^2 \cos(12\pi\vx_1 + \frac{4j\pi}{n}) )+ 0.6 \vx_1]\cos(6\pi\vx_1 + \frac{j\pi}{n})\} \nonumber \\
&\text{and } s_2 = \sum_{j \in J_2} \{\vx_j - [0.3\vx_1^2 \cos(12\pi\vx_1 + \frac{4j\pi}{n}) )+ 0.6 \vx_1]\sin(6\pi\vx_1 + \frac{j\pi}{n})\}, \nonumber \\
&J_1 = \{j|j \text{ is odd and } 2 \leq j \leq n\} \text{ and } J_2 = \{j|j \text{ is even and } 2 \leq j \leq n\}. \nonumber
\end{align}}\\
\hline
\end{tabular}

\newpage

\subsection{Real-world application problems}
\label{subsec_supp_real_world_problem}
We also conduct experiments on $5$ real-world multi-objective engineering design problems~\citep{tanabe2020easy} that are initially proposed in different communities for different applications:

\textbf{Four Bar Truss Design.} This problem is to design a four-bar truss. The two objectives to optimize are its structural volume and joint displacement. The decision variables are the length of the four bars. The details of this problem can be found in \citet{cheng1999generalized}. 

\textbf{Pressure Vessel Design.} This problem is to design a cylindrical pressure vessel. The two objectives to minimize are the total cost (material, forming, and welding) and the violations of three different design constraints. This problem has four decision variables, which are the shell thicknesses, the pressure vessel head, the inner radius, and the length of the cylindrical section. The details of this problem can be found in \citet{kramer1994augmented}.

\textbf{Disk Brake Design.} This problem is to design a disc brake. The three objectives to minimize are the mass, the minimum stopping time, and the violations of four design constraints. This problem has four decision variables: the inner radius, the outer radius, the engaging force, and the number of friction surfaces. The details of this problem can be found in \citet{ray2002swarm}.  

\textbf{Gear Train Design.} This problem is to design a gear train with four gears. The three objectives to minimize are the difference between the realized gear ration and the required specification, the maximum size of four gears, and the design constraint violations. The decision variables are the numbers of teeth in each of the four gears. The details of this problem can be found in \citet{deb2006innovization}.   

\textbf{Rocket Injector Design.} This problem is to design a rocket injector that needs to minimize the maximum temperature of the injector face, the distance from the inlet, and the temperature on the post tip. It has four decision variables, namely, the hydrogen flow angle, the hydrogen area, the oxygen area, and the oxidizer post tip thickness. The details of this problem can be found in \citet{vaidyanathan2003cfd}.  

These real-world multi-objective design problems do not have known exact Pareto fronts. We use the approximate Pareto fronts provided by \citet{tanabe2020easy} with a large number of evaluations as our refereed Pareto fronts, which have also been used in other MOBO works. We set the reference point $\vr = 1.1 \times \hat \vz_{\text{nadir}}$ where $\hat \vz_{\text{nadir}}$ is the nadir point of the approximate Pareto front. The problem information and reference points can be found in Table~\ref{supp_table_problem_refpoint}.

\subsection{Experiment settings}
For all experiments, we first randomly sample and evaluate $10$ valid solutions with Latin hypercube sampling~\citep{mckay2000comparison}, and set them as the initial solutions $\{\vX_0, \vy_0\}$. Then we further run the MOBO algorithm to optimize the given problem with a limited evaluation budget ($100$ for all problems). In the main paper, we report the results with a batch size $5$, and there are only $100 / 5 = 20$ batched iterations for each algorithm. In section~\ref{sec_supp_experiment} in this appendix, we also report experimental results with different batch sizes.

\newpage

\section{Additional experiments}
\label{sec_supp_experiment}

\subsection{Run time}
\label{subsec_supp_runtime}

\begin{table*}[ht]
\small
\centering
\caption{Algorithm run time per iteration (in seconds).}
\label{table_alg_runtime}
\tabcolsep=0.12cm
\begin{tabular}{lcccccc}
\toprule
Problem         & MOEA/D-EGO & TSEMO & \multicolumn{1}{l}{USeMO-EI} & \multicolumn{1}{l}{DGEMO} & qEHVI & \multicolumn{1}{l}{PSL(Ours): Model + Selection} \\ \midrule
F1              & 55.21      & 4.82  & 6.12                         & 61.48                     & 36.71 & 5.26 + 1.33 = 6.59                               \\
F2              & 60.12      & 5.18  & 7.06                         & 62.21                     & 42.80 & 6.01 + 1.21 = 7.22                               \\
F3              & 58.66      & 5.63  & 6.84                         & 59.88                     & 38.28 & 5.82 + 1.43 = 7.25                               \\
F4              & 57.81      & 5.03  & 6.51                         & 61.73                     & 40.16 & 5.72 + 1.15 = 6.87                               \\
F5              & 63.08      & 5.41  & 7.24                         & 57.29                     & 38.83 & 5.53 + 1.38 = 6.91                               \\
F6              & 58.57      & 5.26  & 5.97                         & 69.71                     & 42.77 & 5.89 + 1.19 = 7.08                               \\
VLMOP1          & 57.13      & 4.19  & 5.23                         & 60.33                     & 28.37 & 3.88 + 1.06 = 4.94                               \\
VLMOP2          & 61.05      & 4.96  & 6.88                         & 68.20                     & 39.51 & 4.62 + 1.29 = 5.91                               \\
VLMOP3          & 68.72      & 8.20  & 9.03                         & 79.82                     & 71.25 & 5.61 + 2.49 = 8.10                               \\
DTLZ2           & 71.83      & 7.28  & 8.76                         & 83.57                     & 75.92 & 7.02 + 1.59 = 8.61                               \\
Four Bar Truss  & 63.29      & 3.83  & 5.92                         & 62.46                     & 42.73 & 6.01 + 0.78 = 6.79                               \\
Pressure Vessel & 62.61      & 4.61  & 6.59                         & 69.73                     & 45.32 & 4.98 + 1.33 = 6.31                               \\
Disk Brake      & 72.54      & 7.03  & 8.62                         & 75.20                     & 88.79 & 5.50 + 2.28 = 7.78                               \\
Gear Train      & 68.48      & 6.92  & 9.31                         & 84.52                     & 79.55 & 4.24 + 2.13 = 6.37                               \\
Rocket Injector & 69.43      & 8.72  & 10.03                        & 79.31                     & 85.30 & 5.09 + 2.26 = 7.35                               \\ \bottomrule
\end{tabular}
\end{table*}

We report the run time for each algorithm with batch size $10$ per iteration in Table~\ref{table_alg_runtime}. For our proposed PSL algorithm, we further report the detailed run time for learning the set model and for searching the batch solutions. According to the results, PSL can efficiently learn the set model and select a batch of solutions for evaluation at each iteration with a total time budget of less than $10$ seconds. The short PSL batch selection run time is expected, since all solutions are directly sampled from the learned Pareto set with a few further local search steps, rather than optimizing the acquisition function(s) from scratch as in other MOBO methods.  

We want to emphasize that the run time for different MOBO algorithms strongly depends on the implementation, which is also highlighted in the current works~\citep{daulton2020differentiable, lukovic2020diversity}. The objective evaluations in real-world applications are usually very expensive and involve costly real-world experiments or simulations that take days to run. All the MOBO algorithms are efficient and have negligible run-time overheads in these real-world application scenarios. 

\newpage

\subsection{Quality of the learned Pareto set}
\label{subsec_supp_quality}

\begin{figure}[ht]
\centering
\vspace{-0.15in}
\subfloat[Four Bar Truss]{\includegraphics[width = 0.23\linewidth]{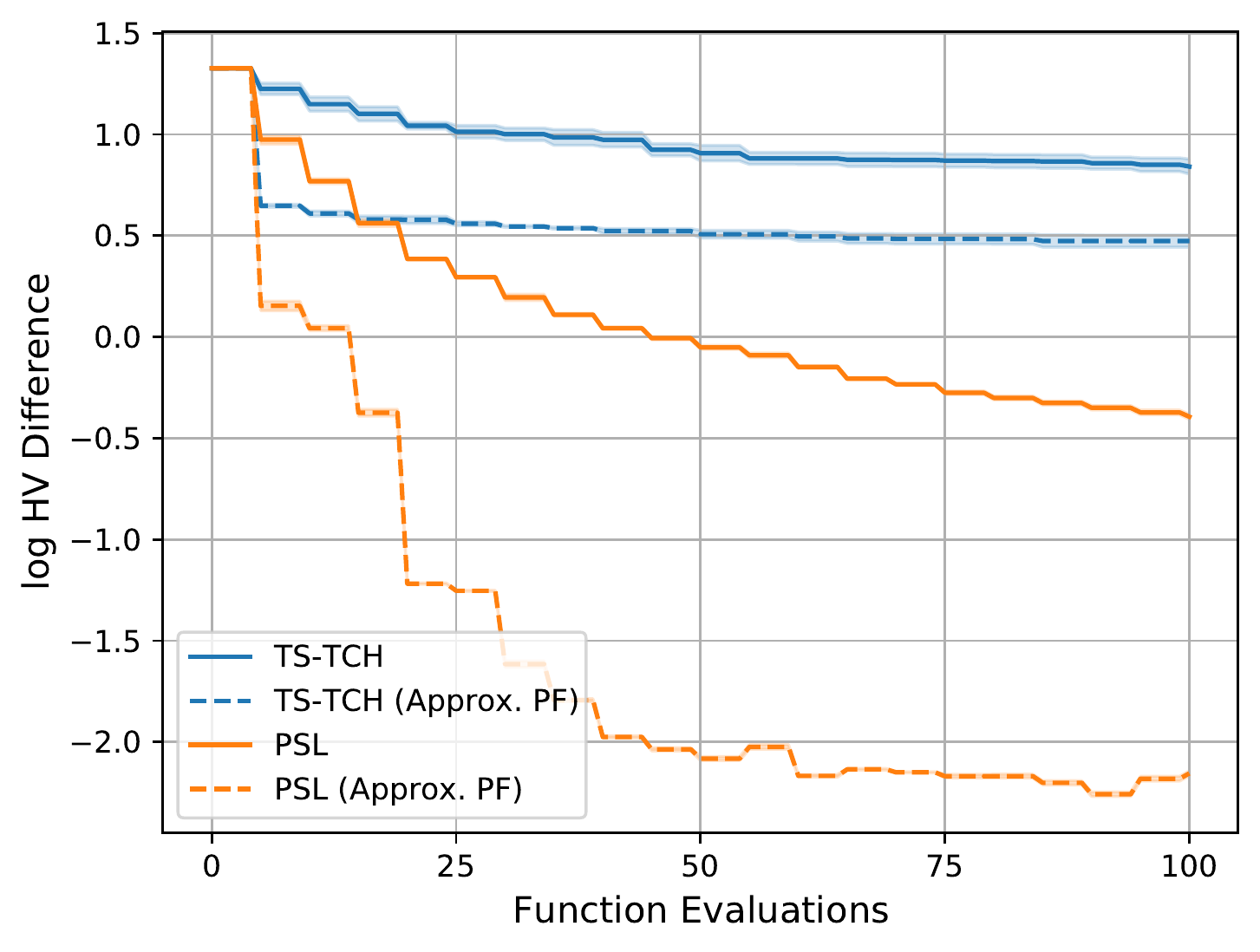}} \hfill
\subfloat[Disk Brake]{\includegraphics[width = 0.23\linewidth]{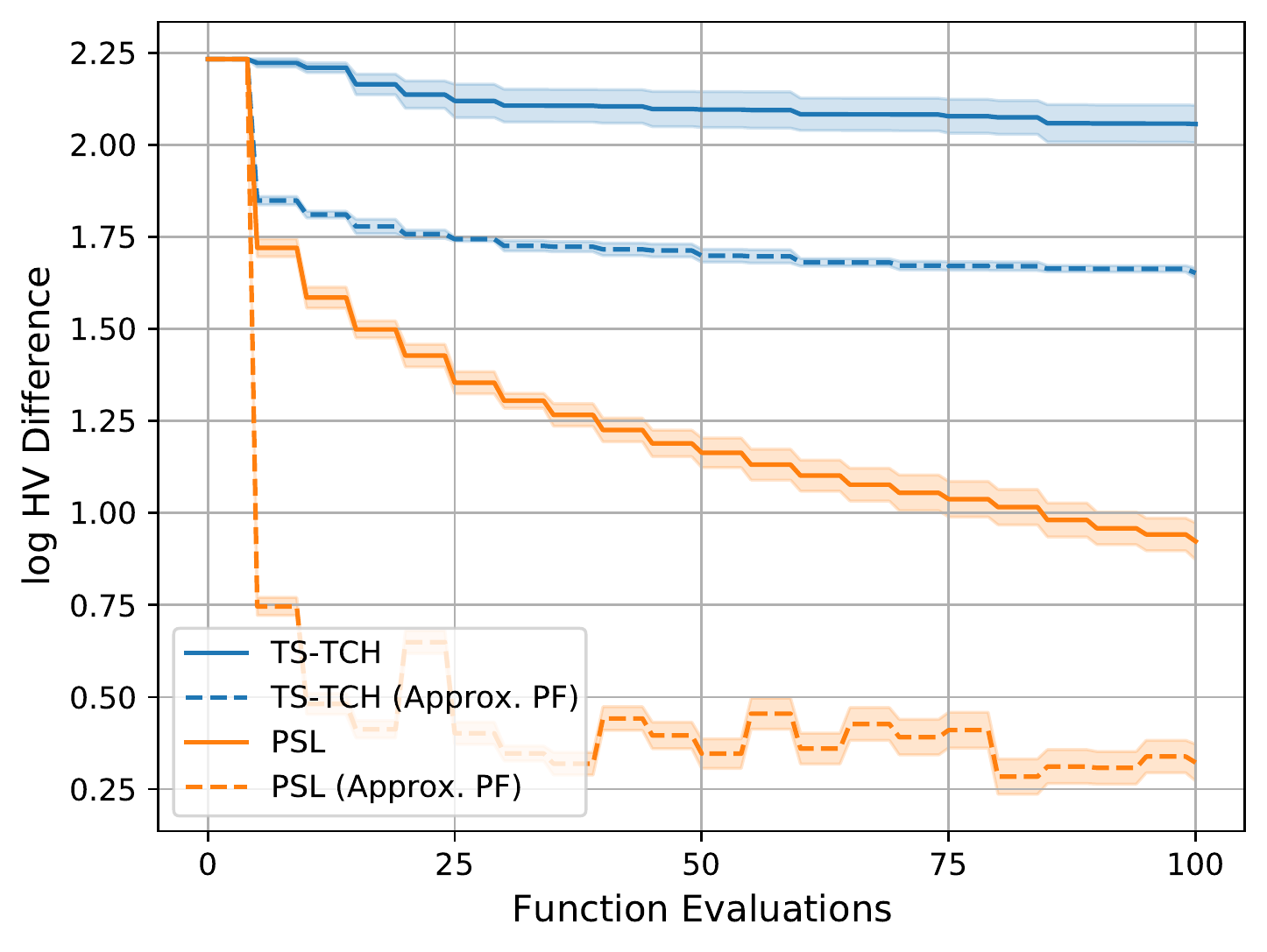}}\hfill
\subfloat[Gear Train]{\includegraphics[width = 0.23\linewidth]{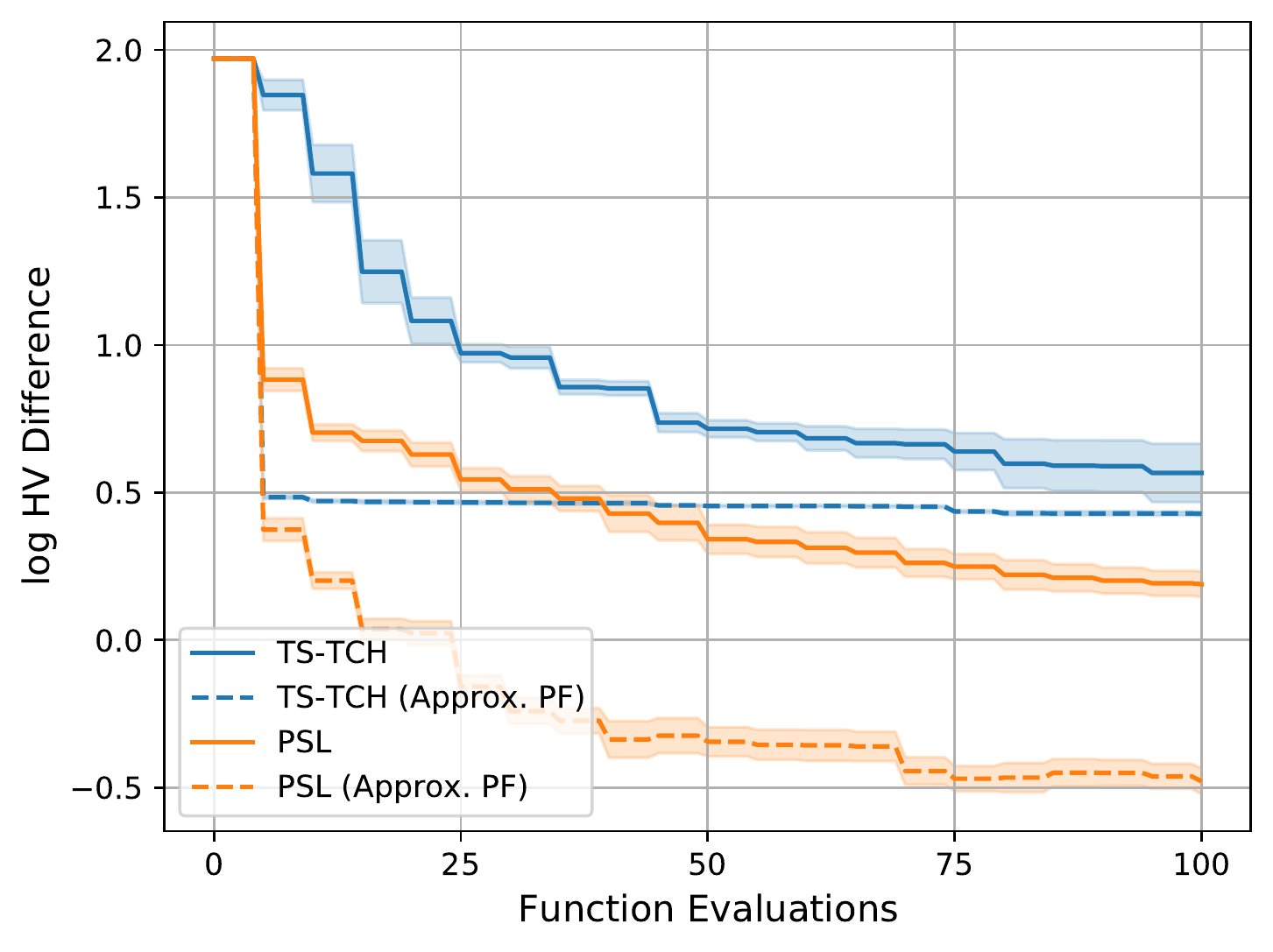}}\hfill
\subfloat[Rocket Injector]{\includegraphics[width = 0.23\linewidth]{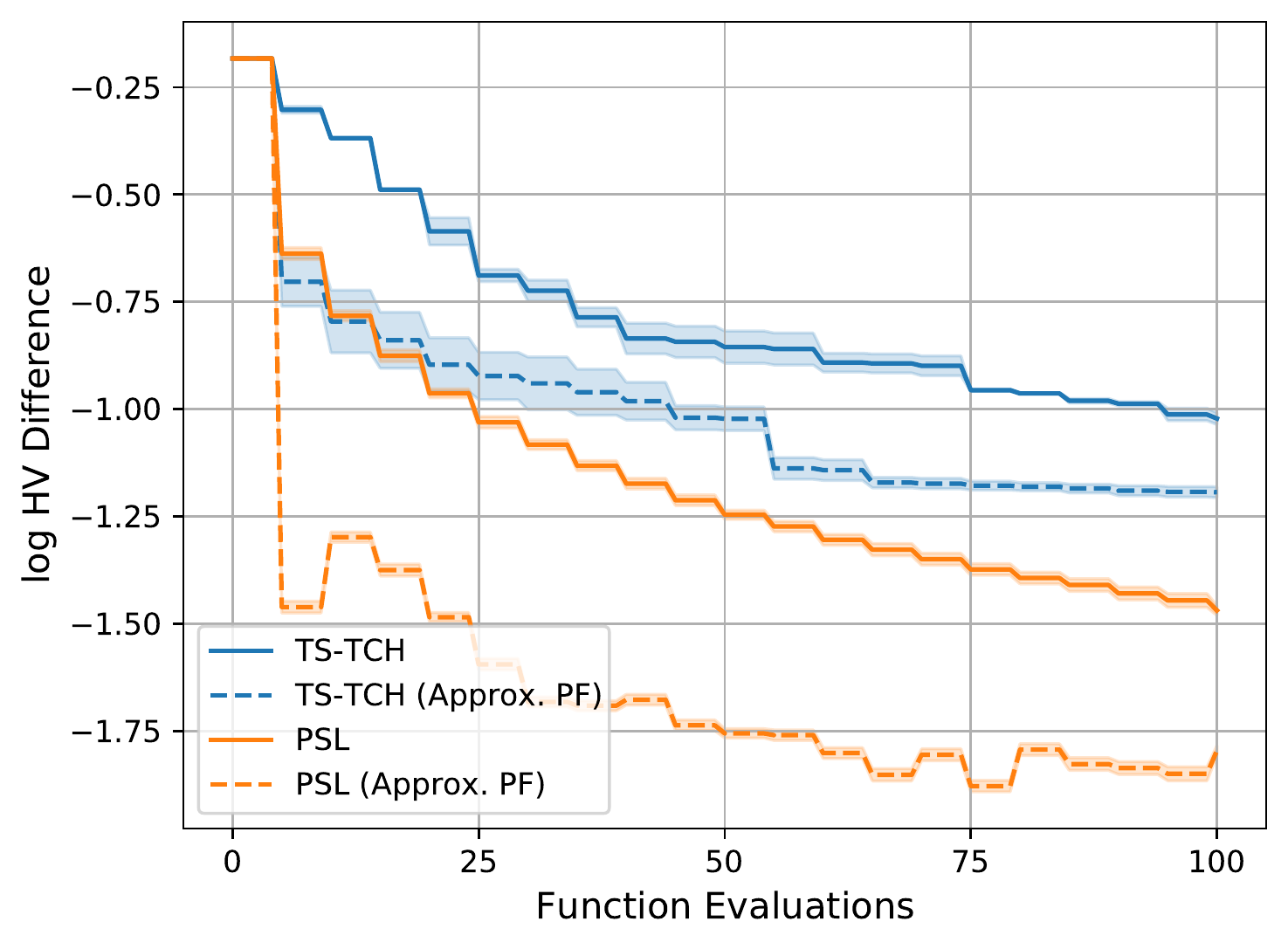}}
\caption{The log hypervolume difference w.r.t. the number of expensive evaluations for PSL and TS-TCH. We report both results for the evaluated solutions (solid lines) and the approximate Pareto front (dashed line).}
\label{supp_fig_pf_quality}
\end{figure}

In the main paper, we have demonstrated that the proposed PSL method can successfully approximate the ground truth Pareto front with a limited evaluation budget as in Figure~\ref{fig_exp_pf_manifold}. In this subsection, we further investigate the performance of the learned Pareto fronts during the optimization process. As shown in Figure~\ref{supp_fig_pf_quality}, the learned Pareto fronts have good quality (with 1,000 random samples) during optimization and lead to promising evaluated solutions performance.

In addition, we also compare our Pareto set model with another modeling approach based on evaluated solutions. Specifically, we run the TS-TCH algorithm and record the evaluated solutions with their corresponding preferences during the optimization process. Then we build a model to approximate the Pareto front based on the evaluated and non-dominated solutions at each iteration step. The model we built has an identical structure (2-layer MLP) to the PSL model, but is now trained in a supervised manner. According to the results in Figure~\ref{supp_fig_pf_quality}, our proposed PSL model has better quality than the model trained in a supervised manner with evaluated solutions.

\subsection{Impact of the Pareto set model for scalarization-based method}
\label{subsec_supp_aggregation}

\begin{figure*}[ht]
\captionsetup[subfigure]{font=scriptsize,labelfont=scriptsize}
\centering
\vspace{-0.15in}
\subfloat[F4]{\includegraphics[width = 0.33\linewidth]{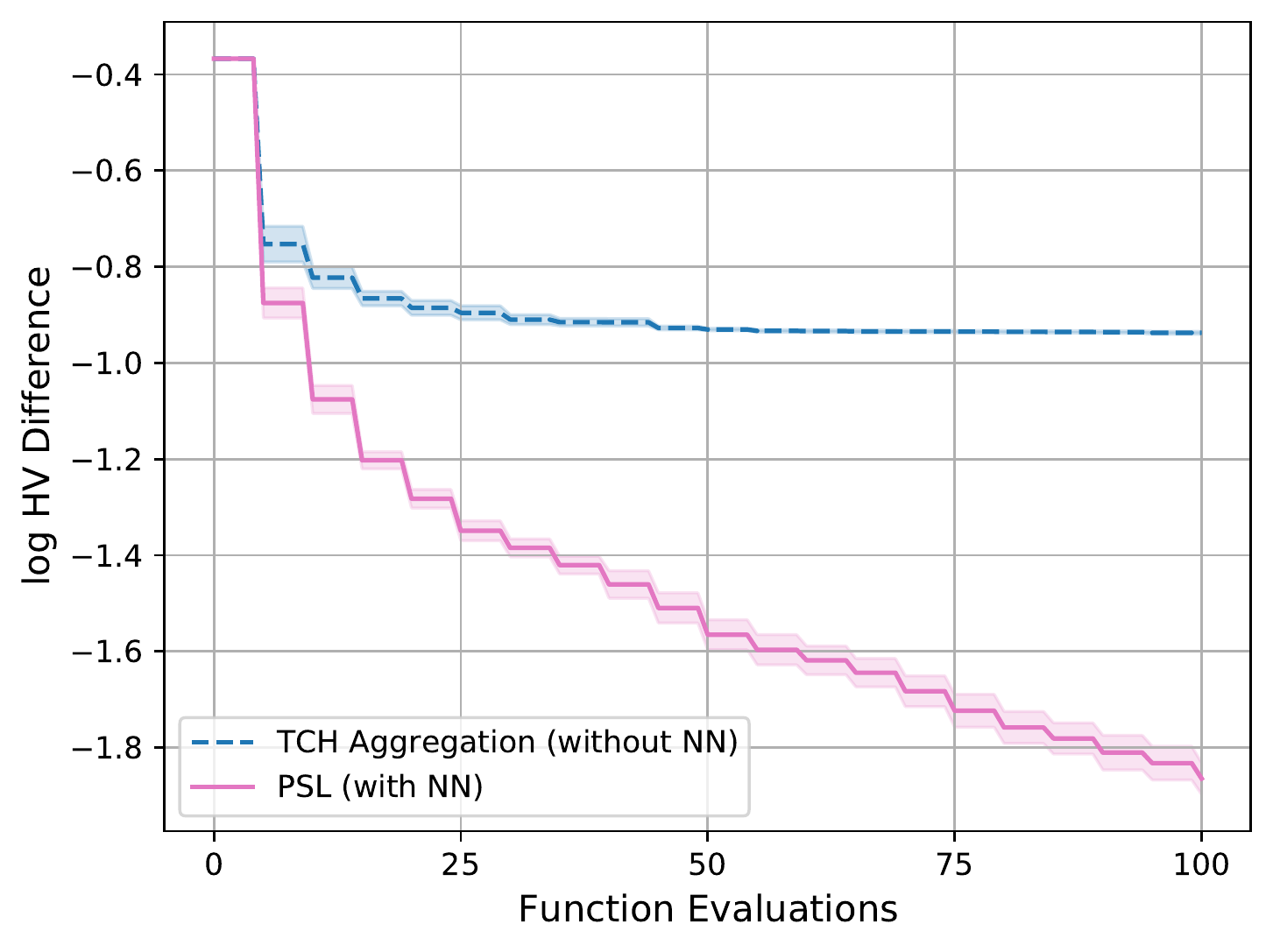}}\hfill
\subfloat[F5]{\includegraphics[width = 0.33\linewidth]{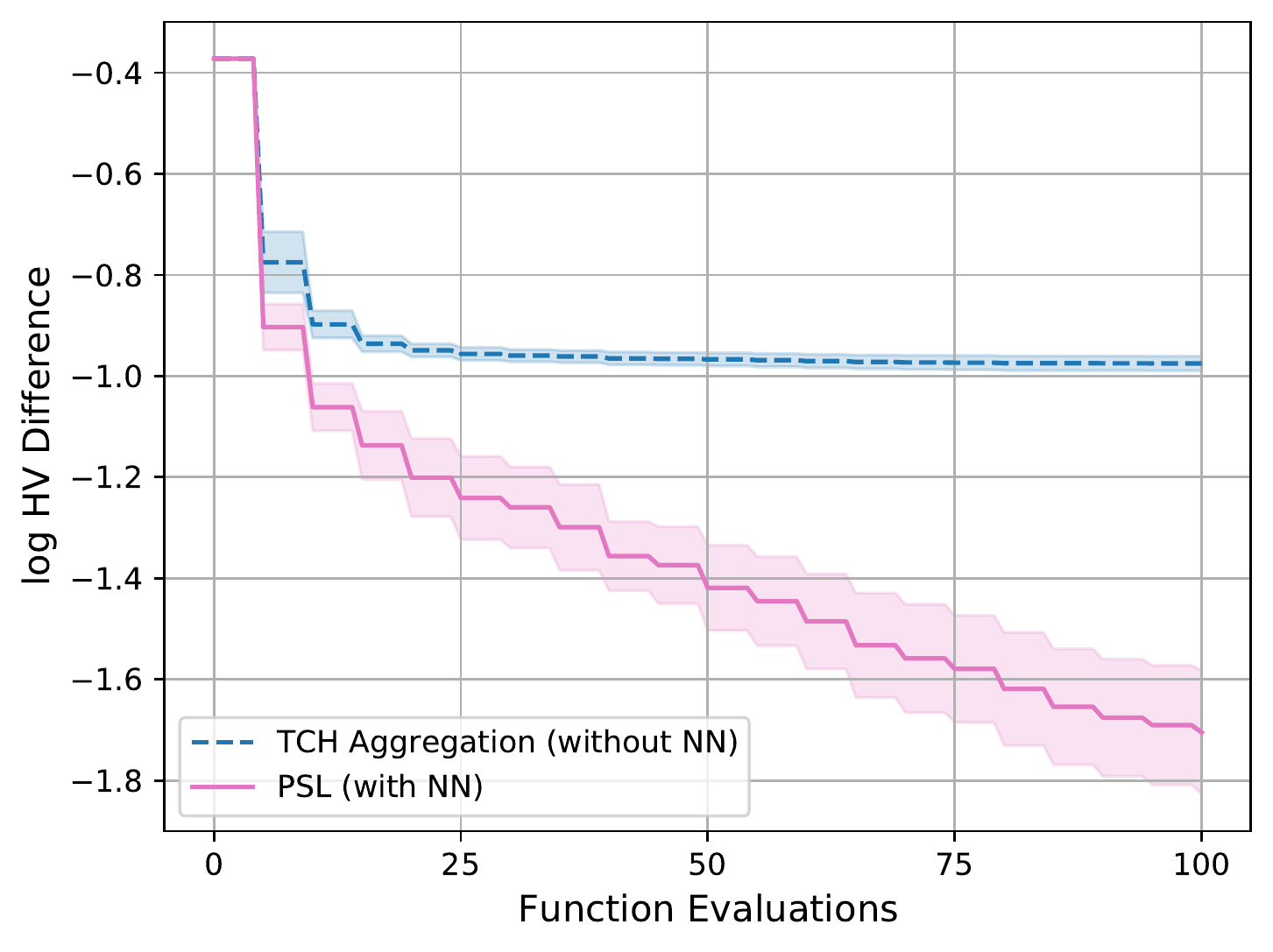}}\hfill
\subfloat[F6]{\includegraphics[width = 0.33\linewidth]{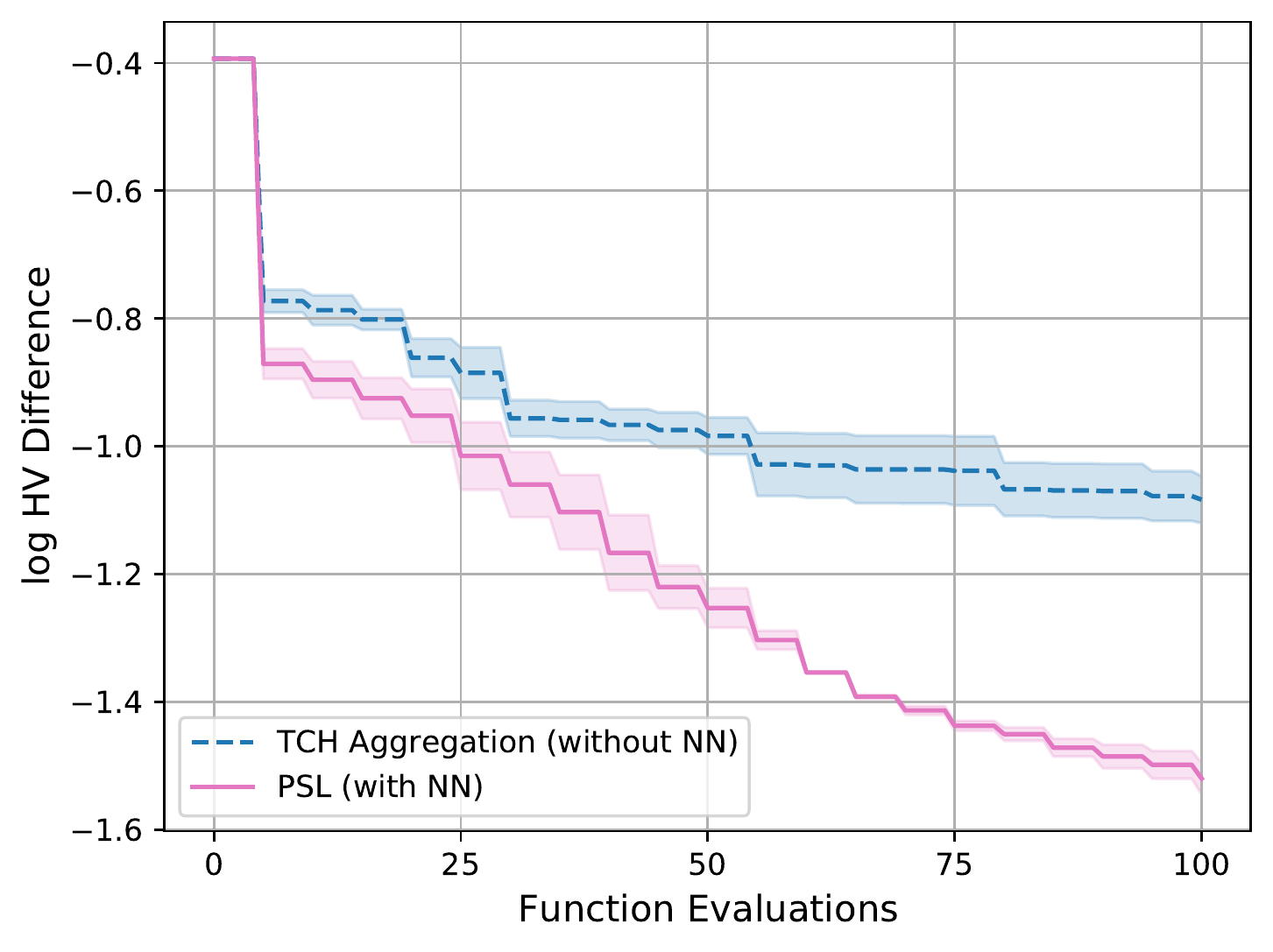}}
\caption{The log hypervolume difference w.r.t. the number of expensive evaluation for different algorithms for F4, F5 and F6.}
\label{fig_hv_without_pareto_set_model}
\end{figure*}

In this work, we choose the weighted Chebyshev scalarization with an ideal point and epsilon mainly due to its good theoretical property. In this subsection, we conduct an ablation study on our proposed PSL method with and without the Pareto set model. The results in Figure~\ref{fig_hv_without_pareto_set_model} show that simply optimizing the acquisition function for scalarization could lead to a significant performance drop. It confirms that the proposed Pareto set model is important for the overall promising performance.

\clearpage

\subsection{Gaussian perturbation for generating candidate set}
\label{subsec_supp_guassian_pertubation}

\begin{figure*}[h]
\captionsetup[subfigure]{font=scriptsize,labelfont=scriptsize}
\centering
\vspace{-0.15in}
\subfloat[ZDT1]{\includegraphics[width = 0.33\linewidth]{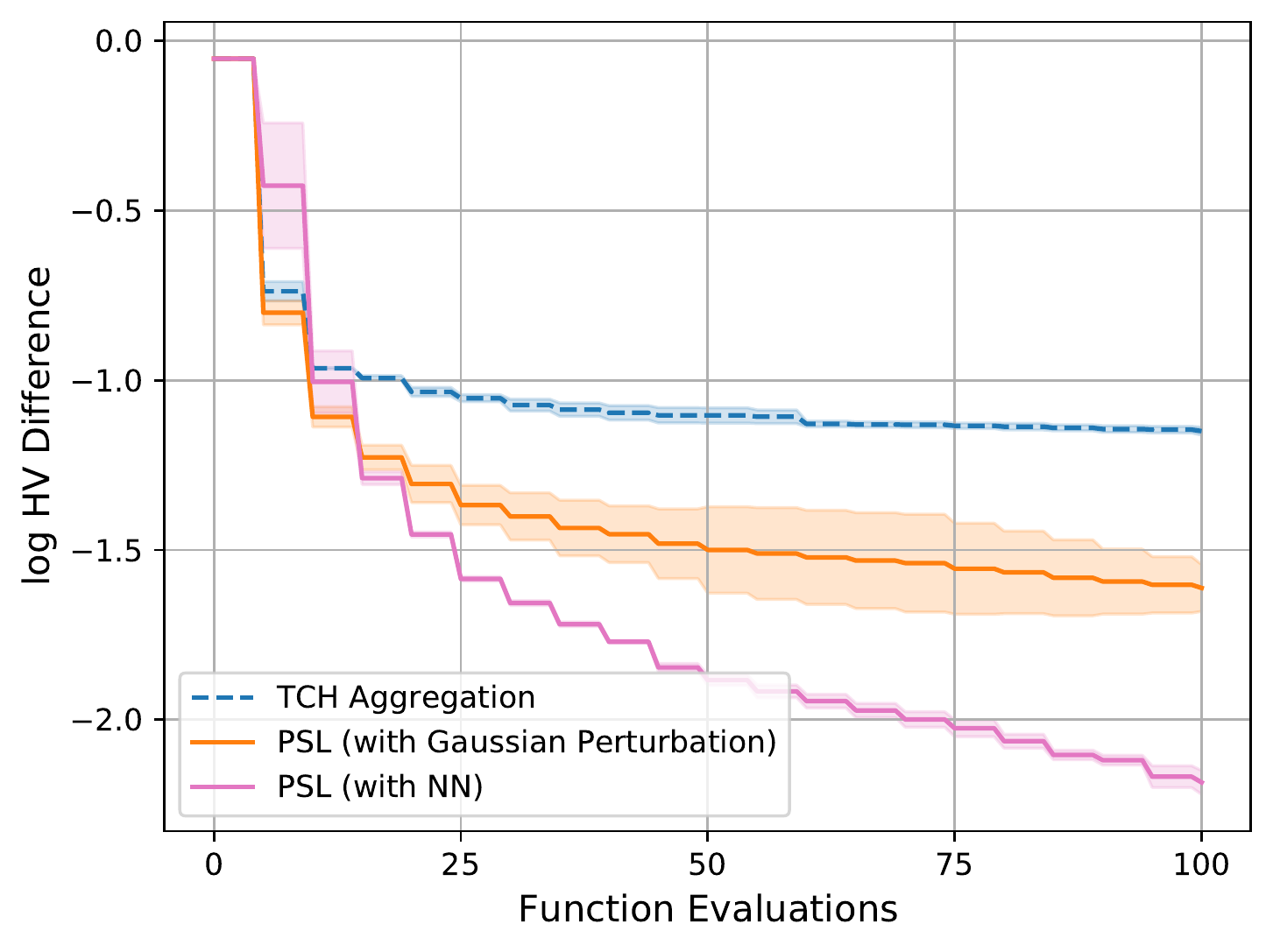}}\hfill
\subfloat[F5]{\includegraphics[width = 0.33\linewidth]{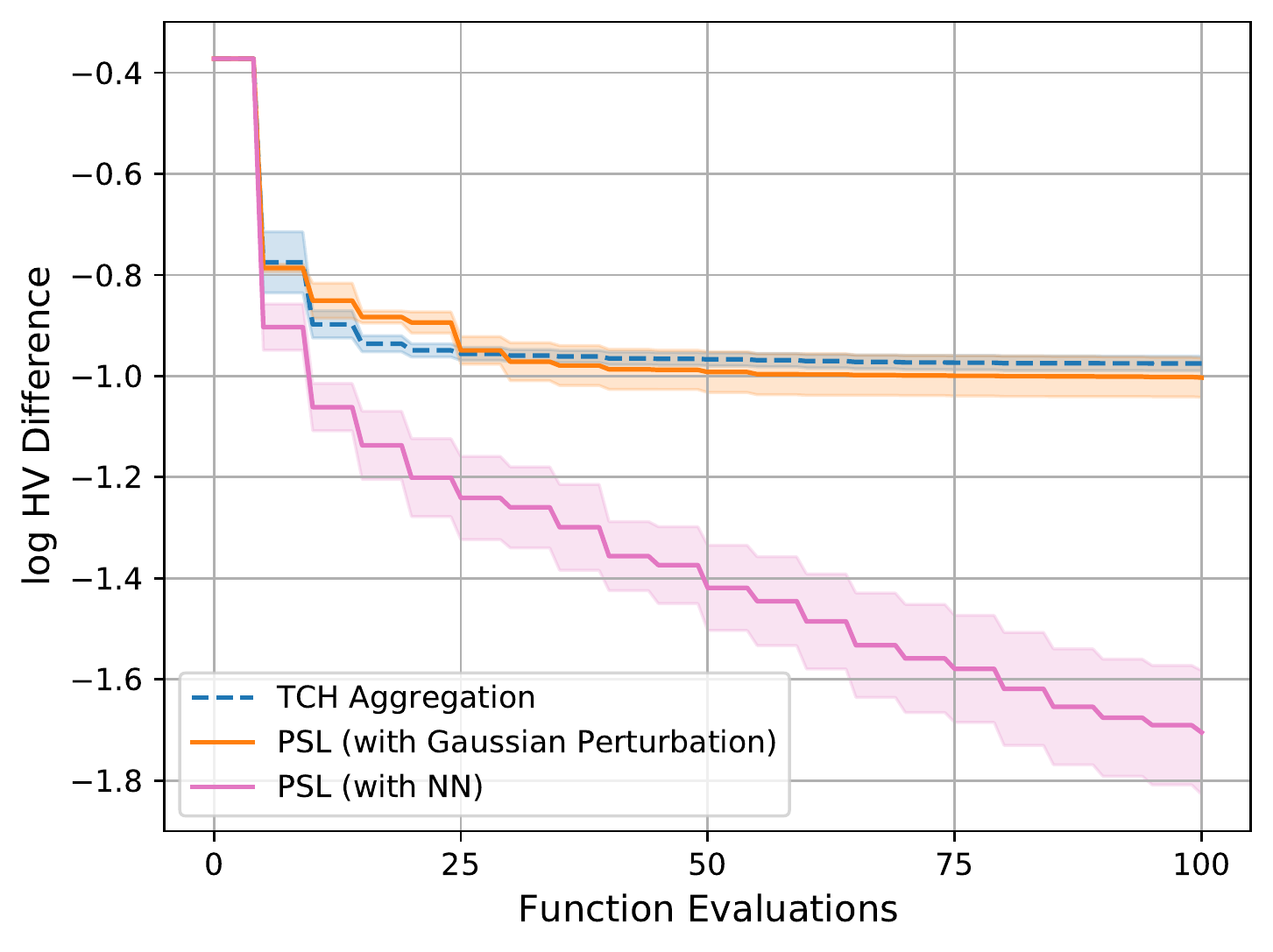}}\hfill
\subfloat[F6]{\includegraphics[width = 0.33\linewidth]{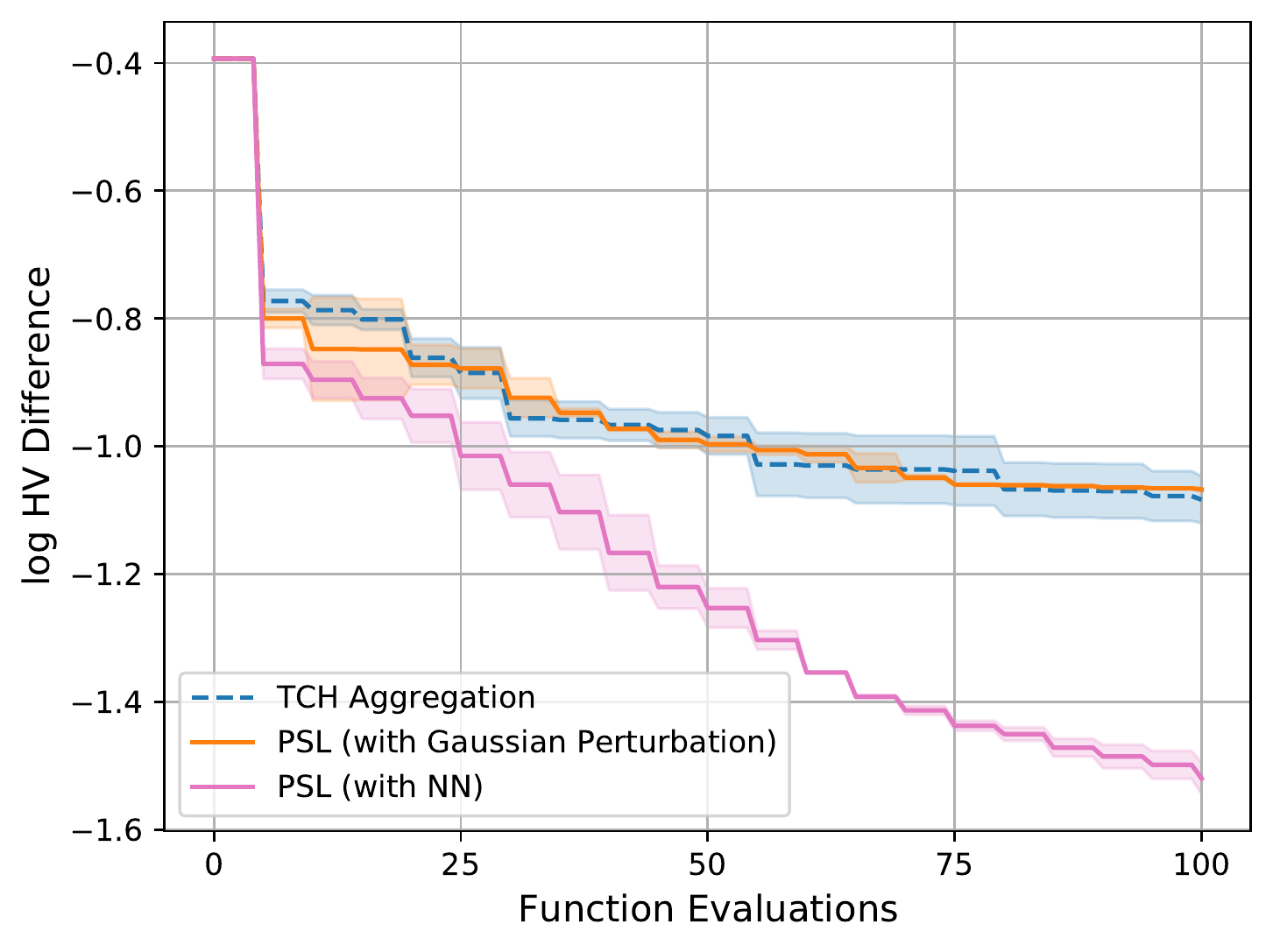}}
\caption{The HVI-LCB acquisition values obtained by different search methods along the PSL's optimization process on ZDT1, F5, and F6.}
\label{fig_guassian_pertubation}
\end{figure*}

In this subsection, we conduct an ablation on small perturbing the best points v.s. Pareto set model for generating the candidate set. Based on the results shown in Figure~\ref{fig_guassian_pertubation}, the Gaussian perturbing method can improve the performance of simple scalarization on problems with simple Pareto set (e.g., ZDT1~\cite{zitzler2000comparison}) but not the problems with complicated Pareto set (e.g., F5 and F6). On all problems, our proposed PSL method still achieves the best performance.

DGEMO~\cite{lukovic2020diversity} also has a local search approach to expand the candidate set around the best points (in terms of surrogate value) with the first and second derivatives of the GP surrogate model. In our experiments, PSL can outperform DGEMO on most problems, which confirms the importance of the Pareto set model for generating the candidate set.

On the other hand, the local search methods might provide complementary candidate solutions to PSL, especially at the early stage of optimization when the approximate Pareto set is inaccurate. We will investigate how to efficiently combine PSL with the local search approaches in future work.

\subsection{Comparisons with HVI-LCB}
\label{subsec_supp_hvi_lcb}

\begin{figure*}[h]
\captionsetup[subfigure]{font=scriptsize,labelfont=scriptsize}
\centering
\vspace{-0.15in}
\subfloat[F4]{\includegraphics[width = 0.33\linewidth]{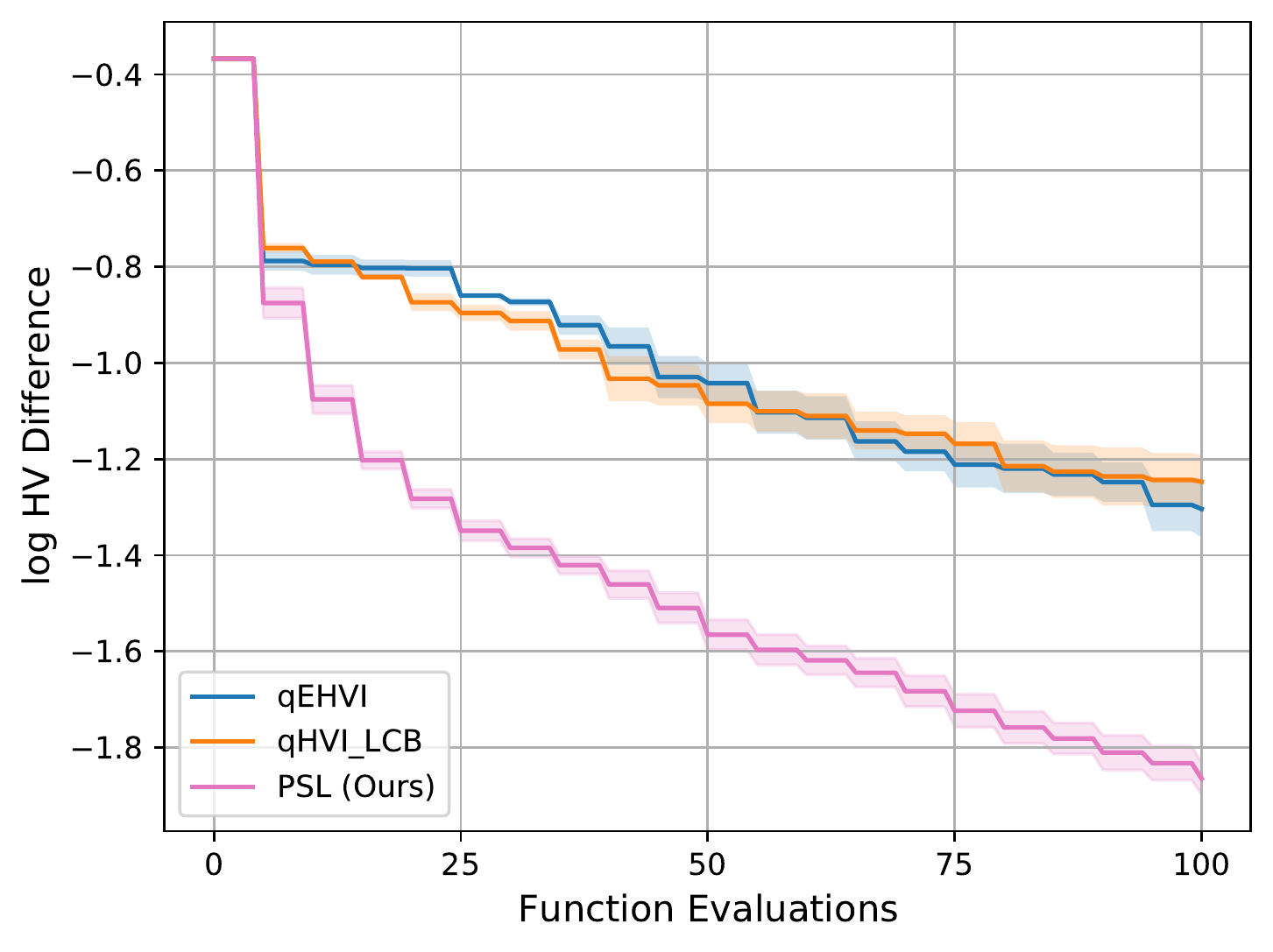}}\hfill
\subfloat[F5]{\includegraphics[width = 0.33\linewidth]{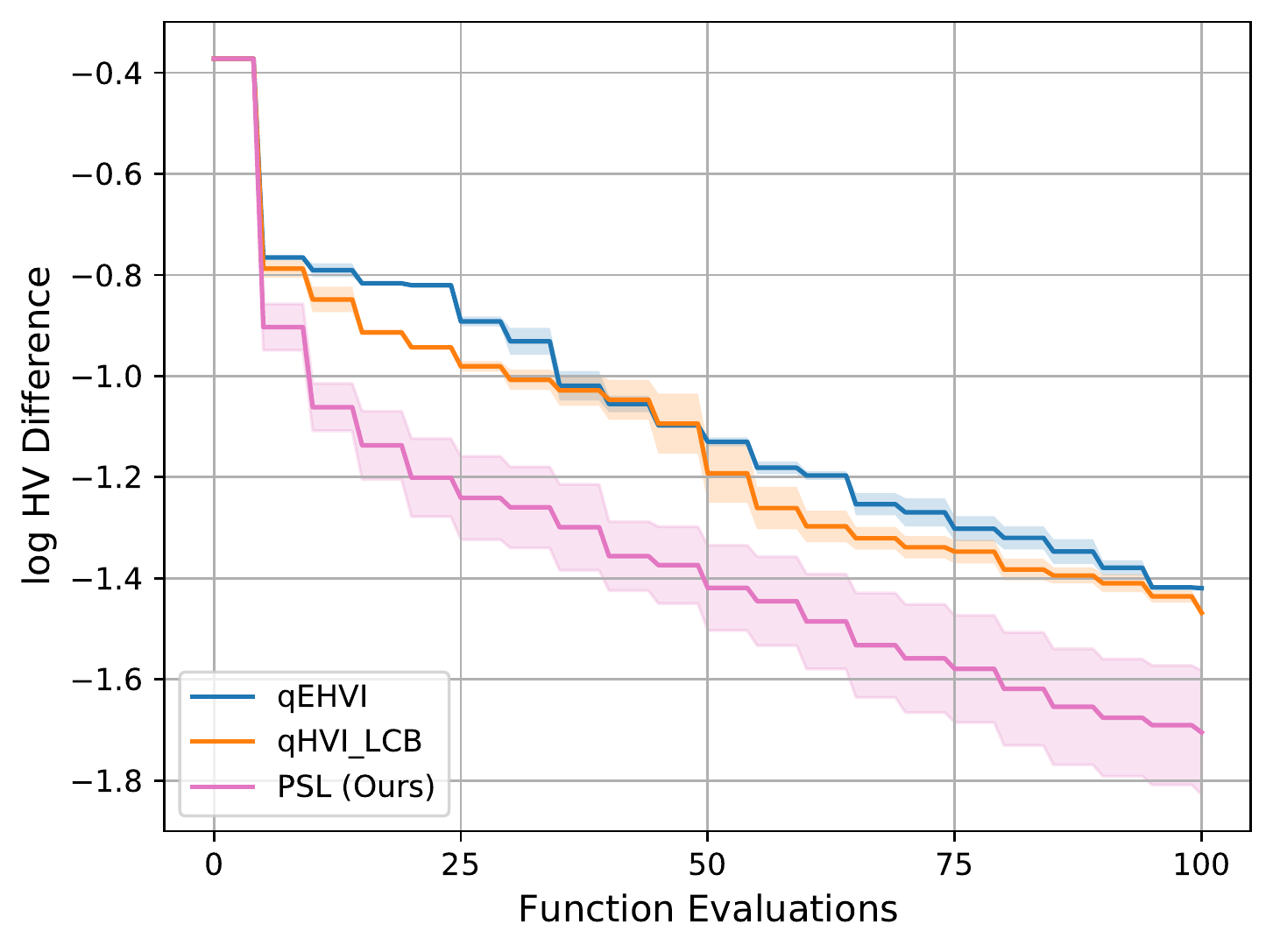}}\hfill
\subfloat[F6]{\includegraphics[width = 0.33\linewidth]{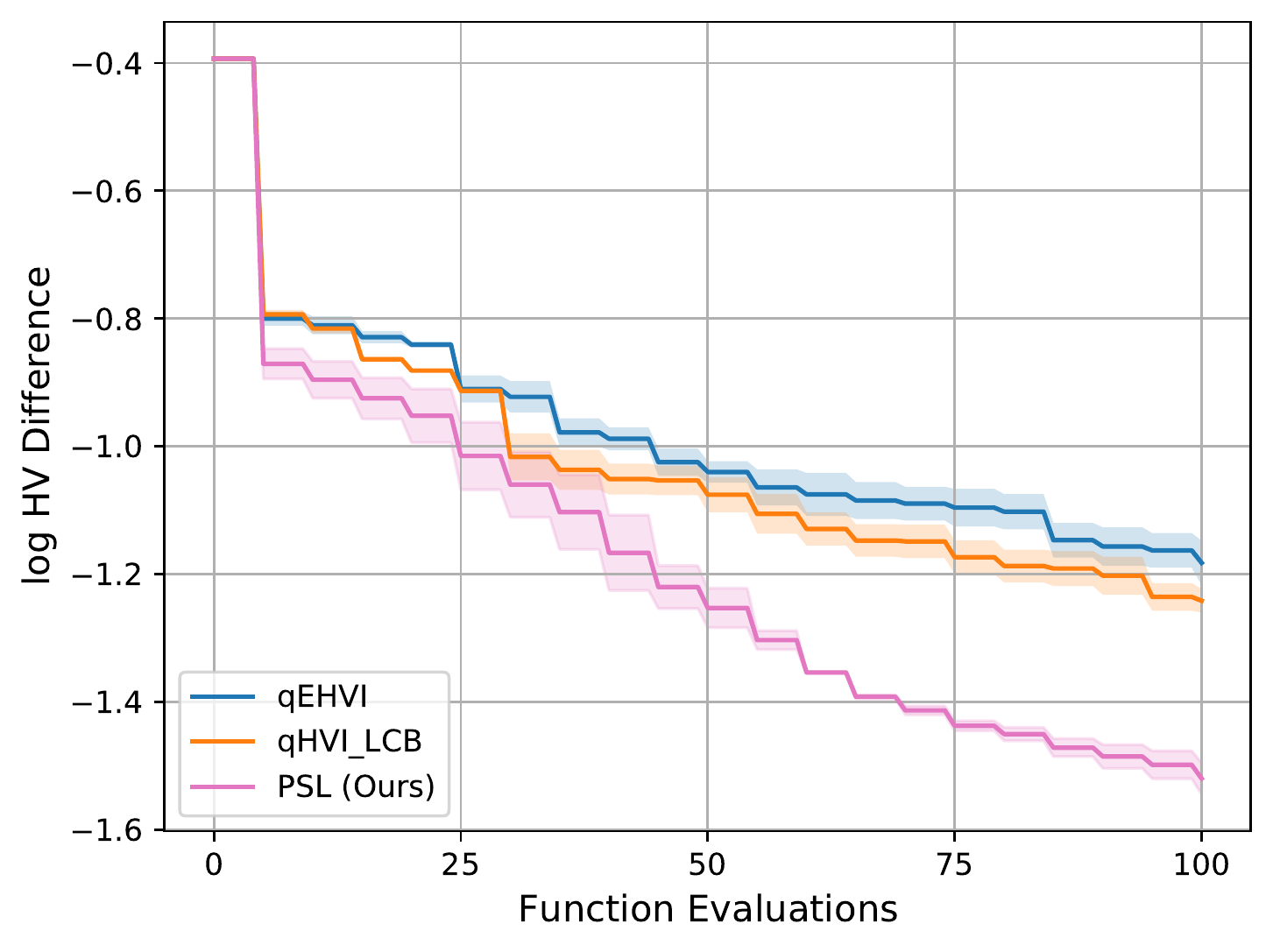}}
\caption{The log hypervolume difference w.r.t. the number of expensive evaluation for qEHVI, qHVI-LCB, and our proposed PSL on F4, F5, and F6.}
\label{fig_hv_trend_hvi_lcb}
\end{figure*}

In this subsection, we conduct an ablation study on maximizing the hypervolume improvement with LCB on the approximate Pareto set vs. searching across the whole design space. The experimental results on three newly proposed problems (i.e., F4, F5, and F6) with complicated Pareto sets are shown in Figure~\ref{fig_hv_trend_hvi_lcb}. We also provide the qEHVI results as reference.  

Based on these results, it is clear that our proposed PSL method can outperform directly searching the entire search space to maximize HVI with LCB. The qHVI-LCB method performs similarly to qEHVI, which indicates that different acquisitions (for HVI) do not have a significant impact on the optimization performance. In other words, with the learned Pareto set, our proposed PSL method can efficiently conduct the acquisition optimization mainly on the promising low-dimensional (e.g., (m-1)-dimensional) manifold, and then lead to significantly better Bayesian optimization performance.

One possible reason for this performance gap could be the difficulty of directly searching the whole design search for maximizing HVI. As shown in recent works~\cite{zhao2022multi,daulton2022multi}, the performance of qEHVI can be improved using meta search region management algorithms such as LaMOO~\cite{zhao2022multi} and trust region methods~\cite{daulton2022multi}. It implies that directly searching the whole design space is not efficient. These meta-algorithms adaptively decompose the whole search region into different subregions along with the optimization process, while our PSL method directly learns the whole approximate Pareto set at each step. Studying how to efficiently combine PSL with these algorithms could be an important future work as discussed in Appendix~\ref{sec_supp_limitation}. 

\subsection{The HVI acquisition values during optimization}
\label{subsec_supp_hvi_acq_value}

\begin{figure*}[h]
\captionsetup[subfigure]{font=scriptsize,labelfont=scriptsize}
\centering
\vspace{-0.15in}
\subfloat[F4]{\includegraphics[width = 0.33\linewidth]{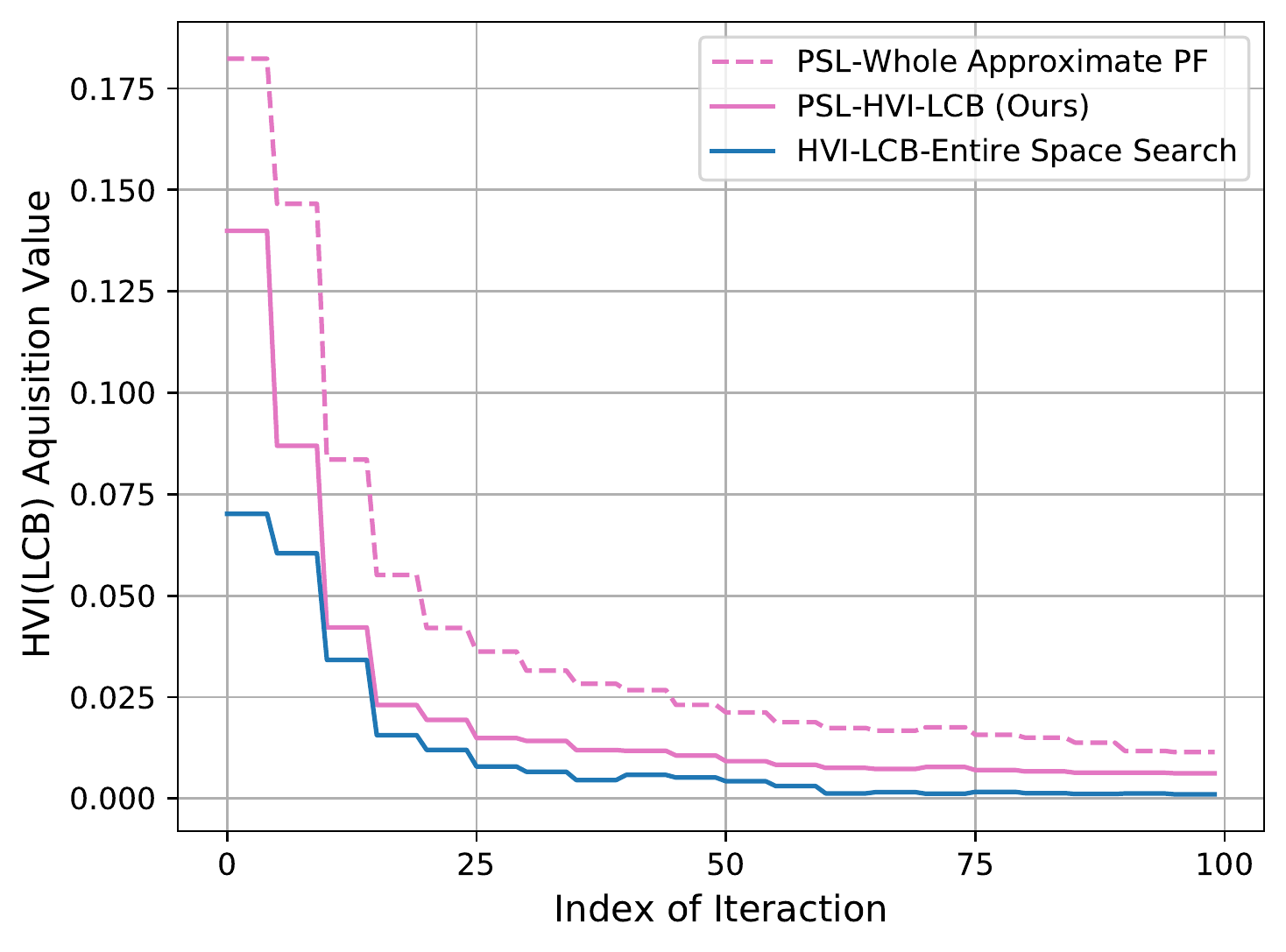}}\hfill
\subfloat[F5]{\includegraphics[width = 0.33\linewidth]{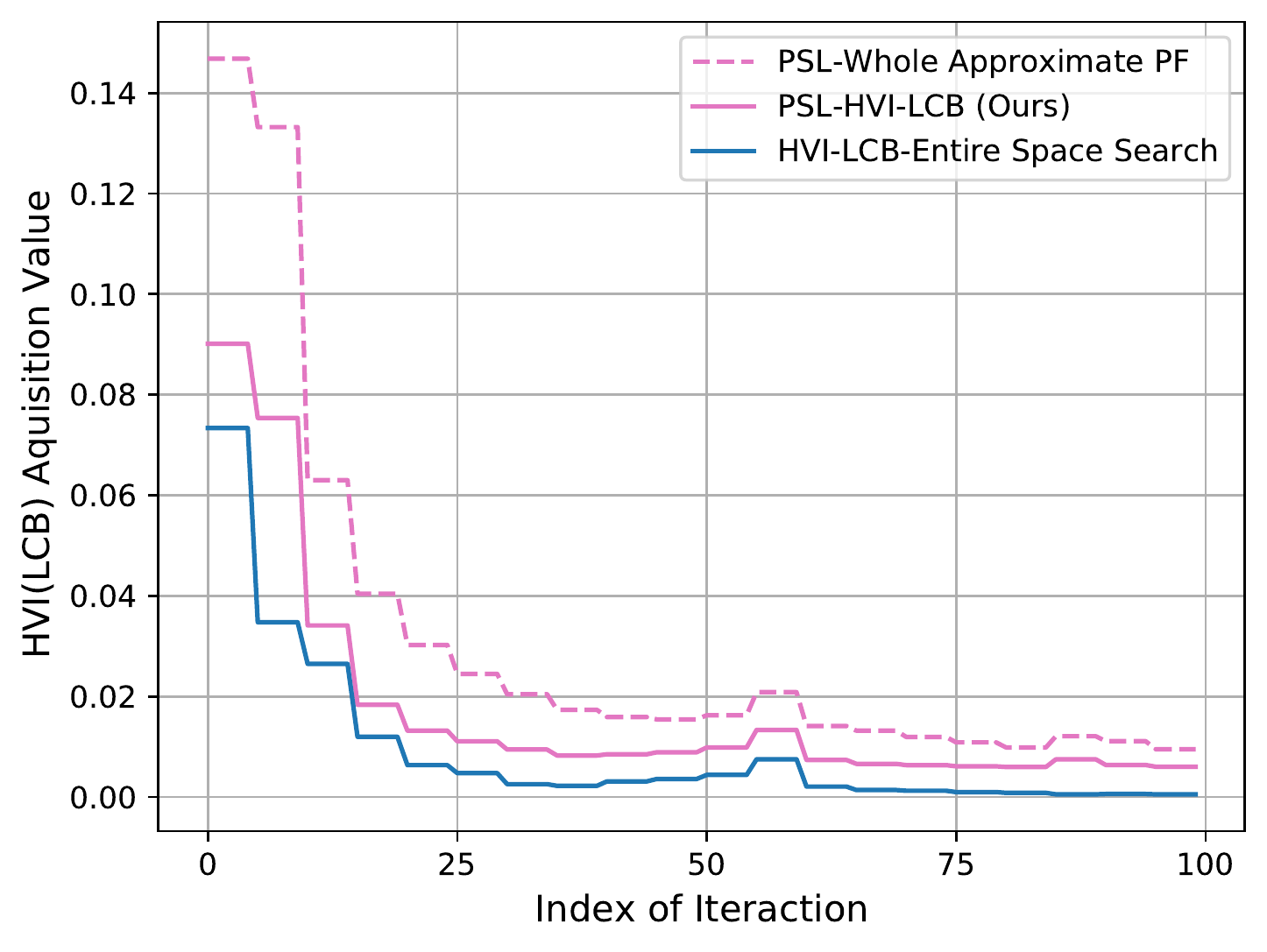}}\hfill
\subfloat[F6]{\includegraphics[width = 0.33\linewidth]{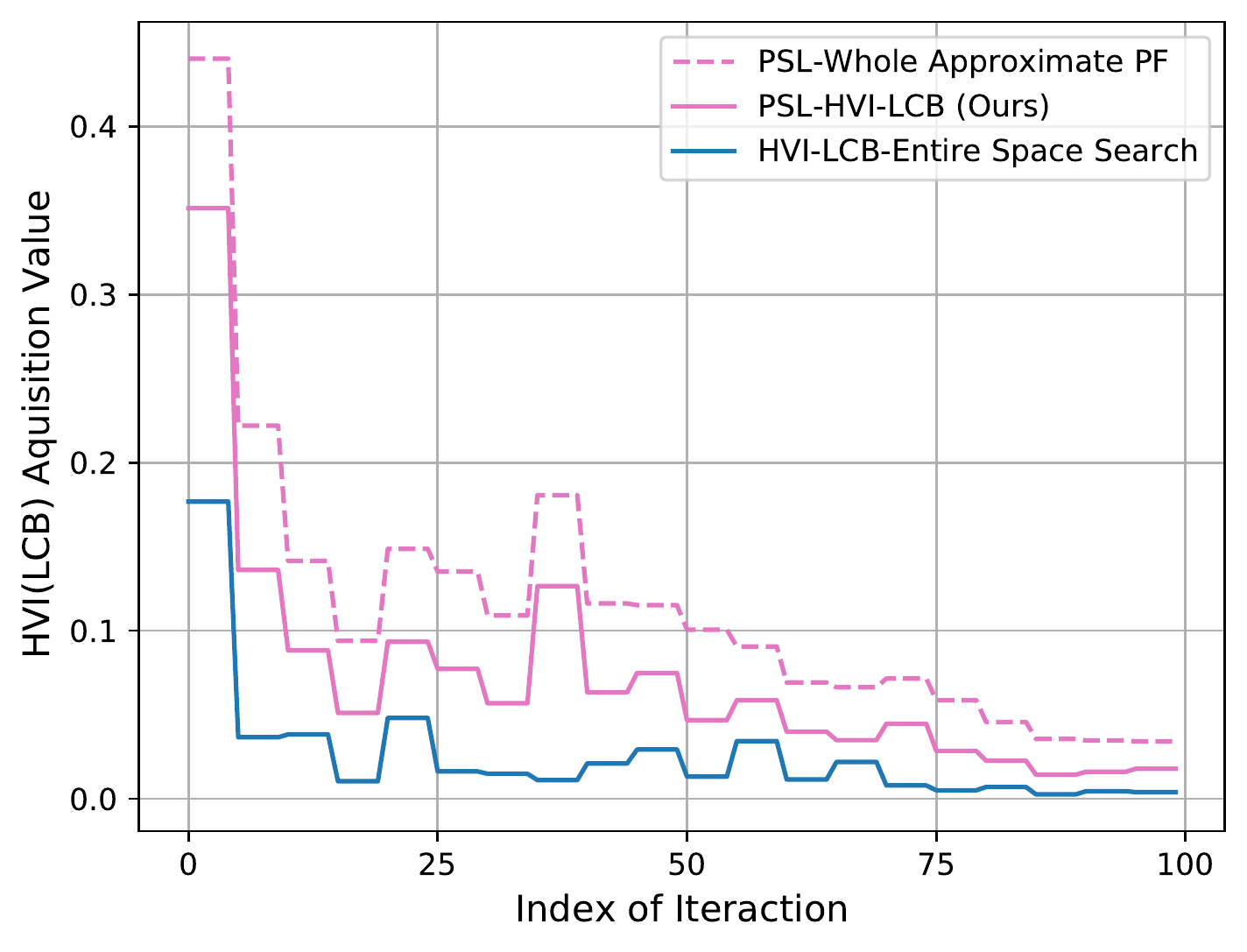}}
\caption{The HVI-LCB acquisition values obtained by different search methods along the PSL's optimization process on F4, F5, and F6.}
\label{fig_hv_trend_hvi_acq_value}
\end{figure*}

In this subsection, we compare the HVI-LCB acquisition values obtained by different search methods. We run our proposed PSL as the optimization algorithm, and conduct different search methods at each iteration. The compared search methods are: 1) searching across the entire design space, 2) our PLS search on the approximate Pareto front, and 3) the HVI-LCB acquisition values for the whole approximate Pareto front with 1,000 sampled solutions (which can be treated as the upper bound for our method). Based on the results, it is clear that our proposed PSL search can produce better HVI-LCB acquisition values than directly searching the entire design space. 

It should be noticed that the reported results are obtained by running PSL, and the search methods have the same surrogate functions and baseline HV (from the same evaluated solutions chosen by PSL) at each iteration. By running qHVI-LCB alone, the less efficient acquisition search at each iteration will lead to the accumulated worse overall performance as shown in Figure~\ref{fig_hv_trend_hvi_lcb}.

\clearpage

\subsection{PSL with different preferences}
\label{subsec_supp_preference}

\begin{figure}[H]
\centering
\vspace{-0.15in}
\subfloat[Two Obj. Prefs.]{\includegraphics[width = 0.23\linewidth]{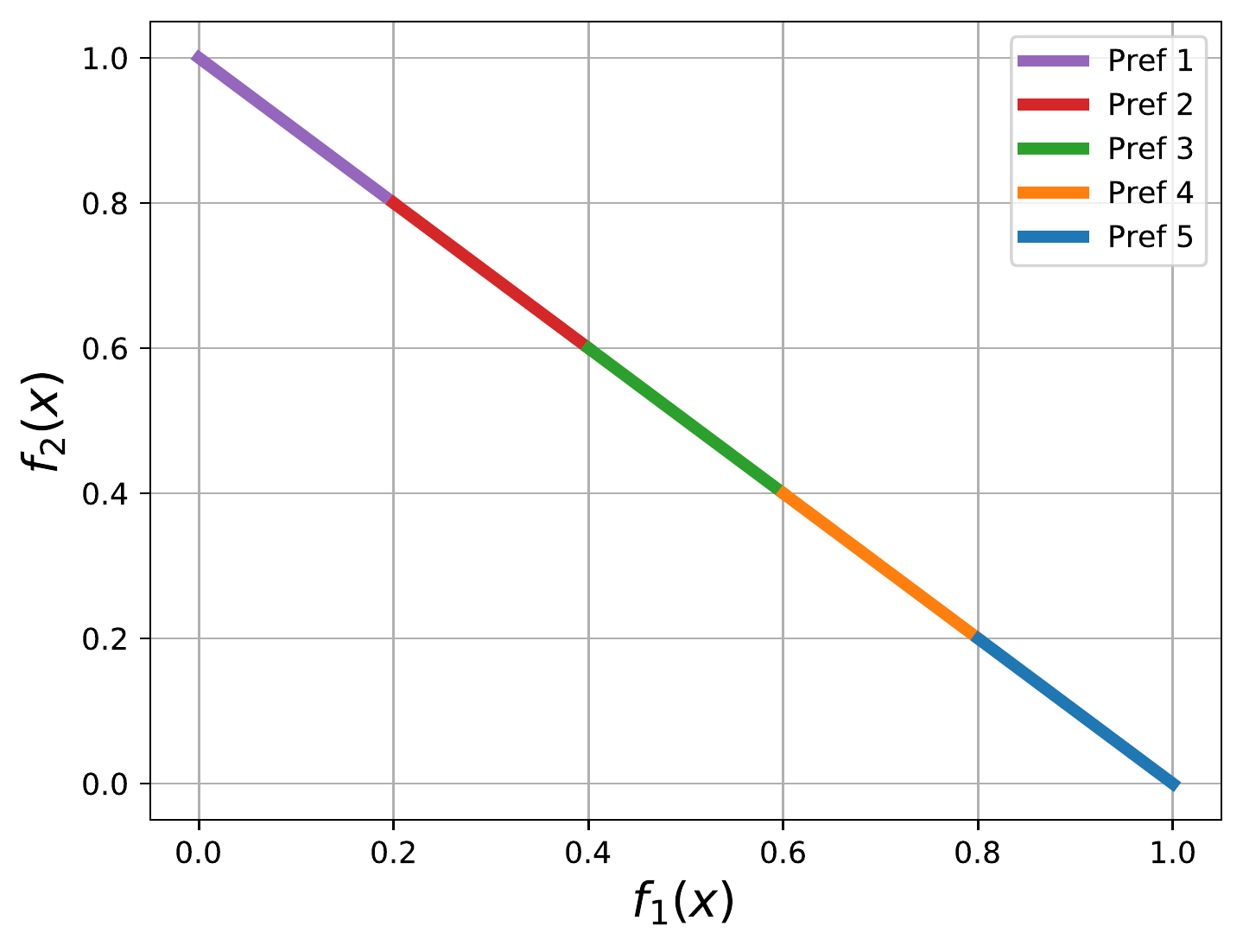}} \hfill
\subfloat[F1]{\includegraphics[width = 0.23\linewidth]{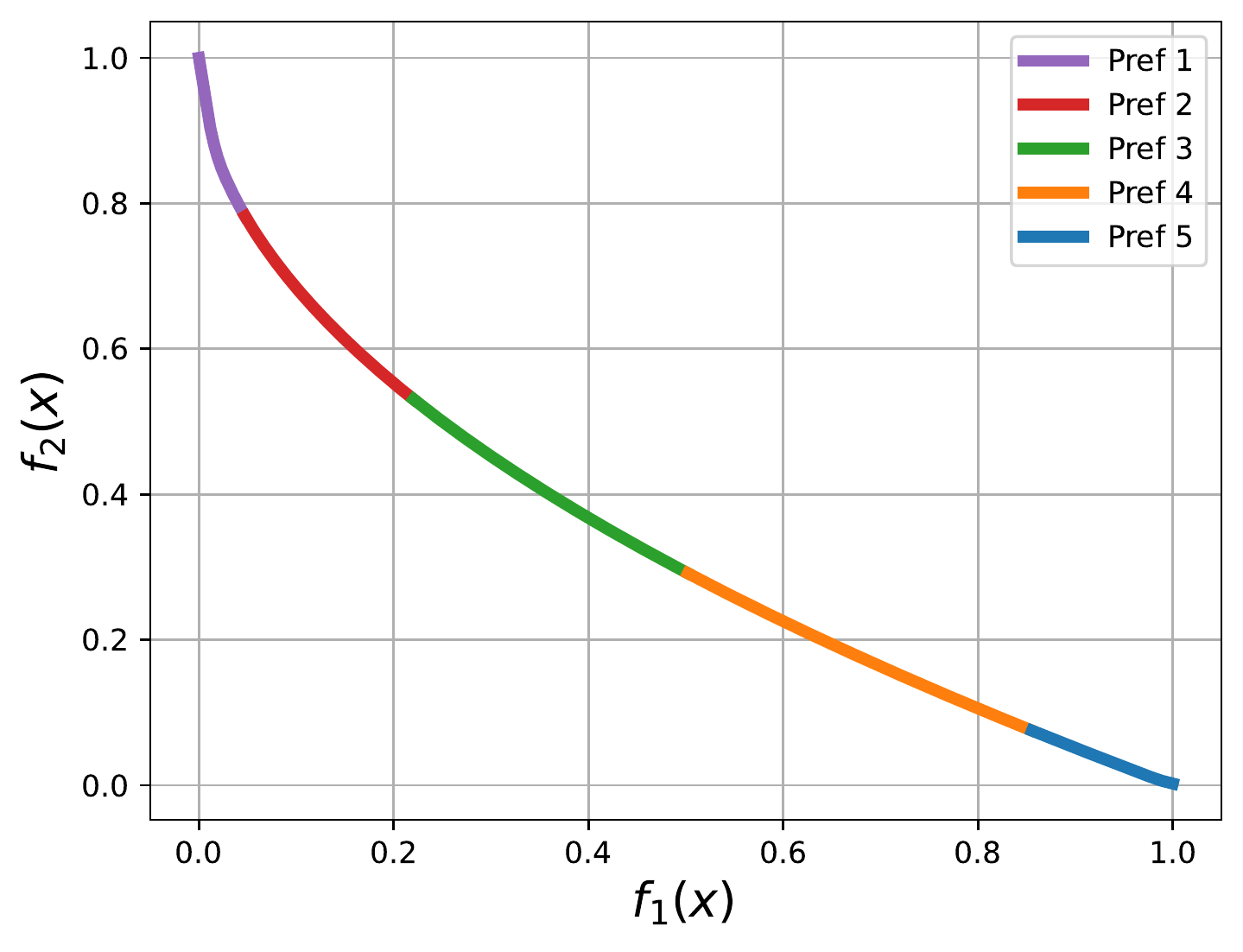}}\hfill
\subfloat[VLMOP2]{\includegraphics[width = 0.23\linewidth]{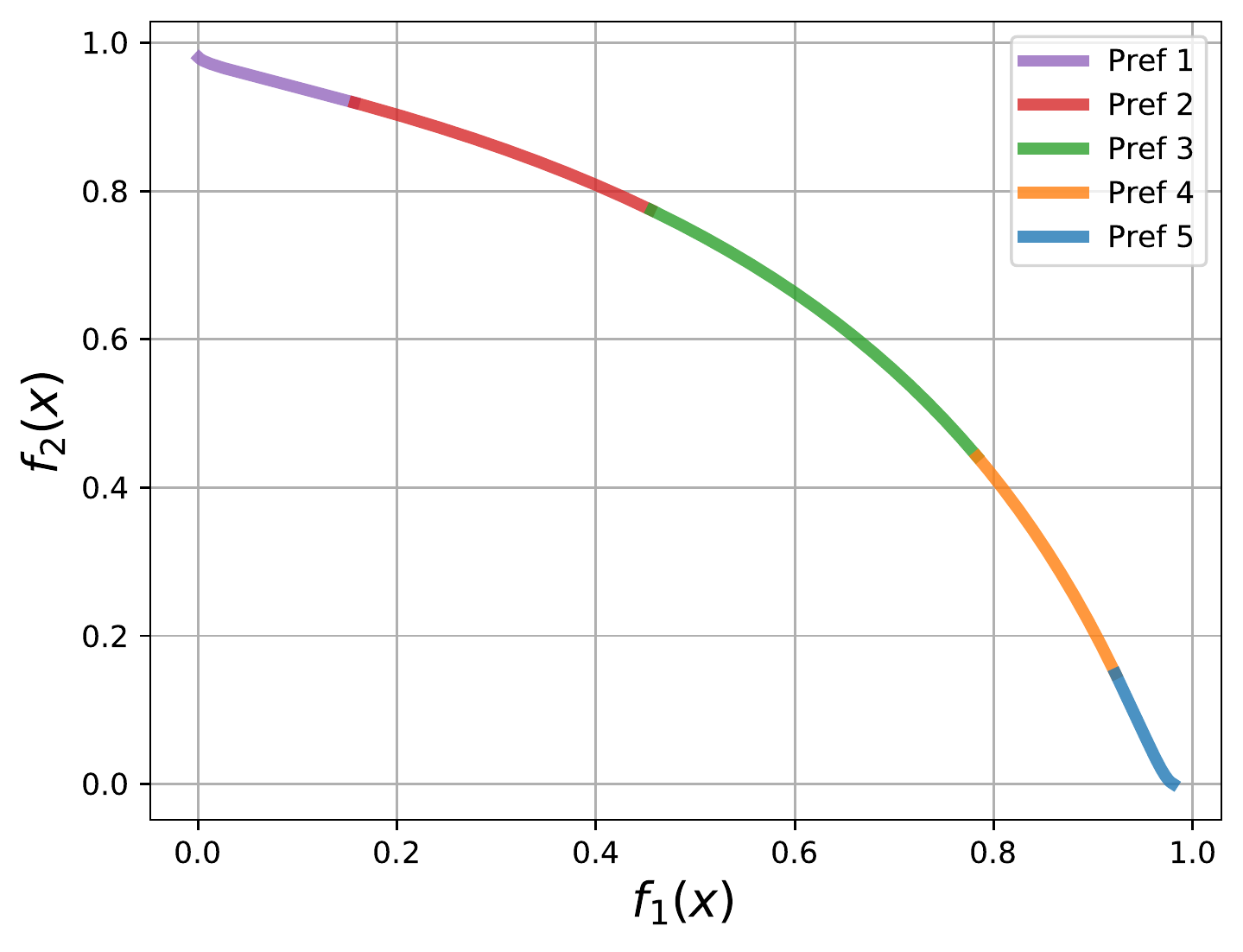}}\hfill
\subfloat[Four Bar Truss]{\includegraphics[width = 0.23\linewidth]{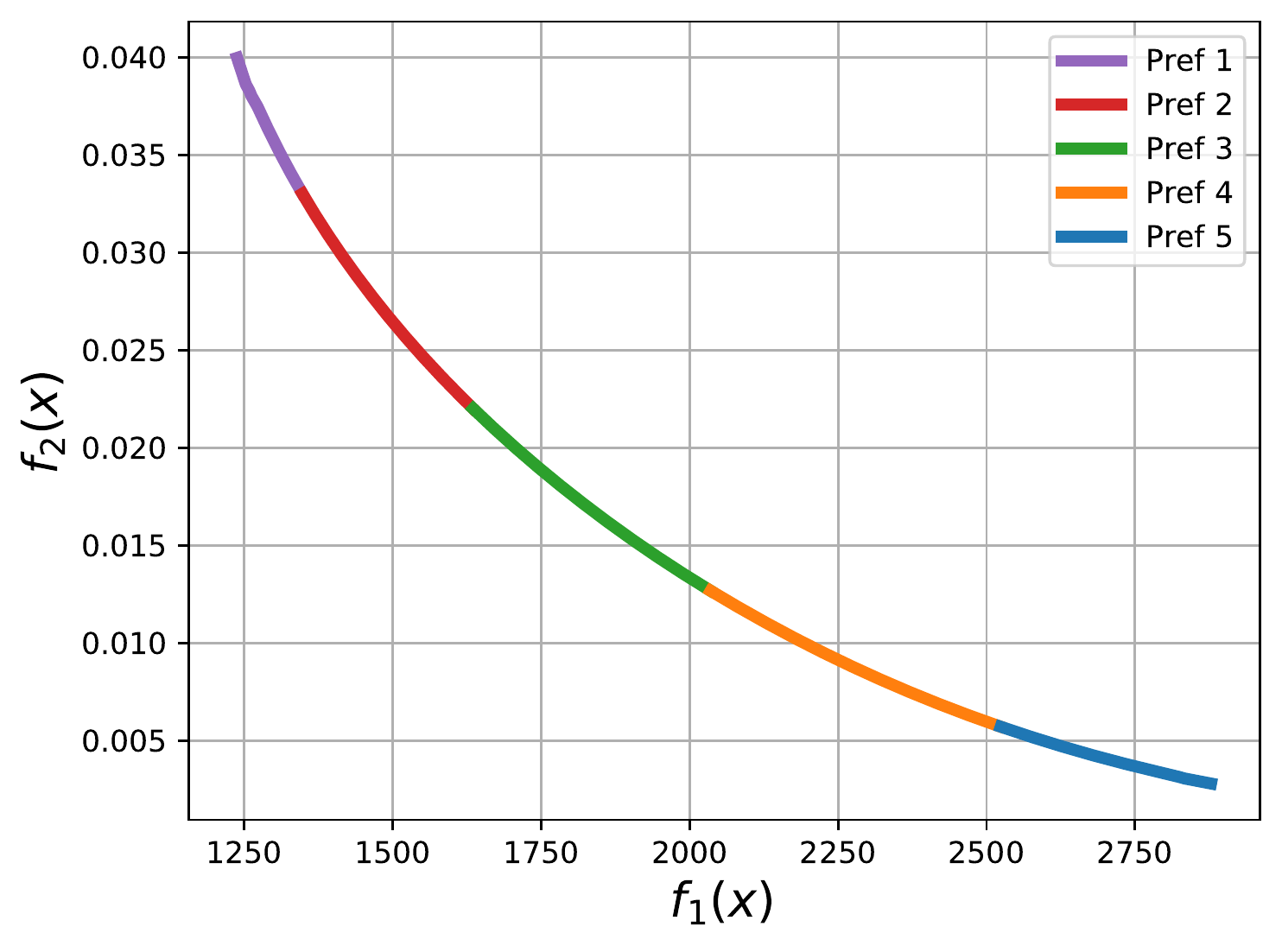}} \\ 
\subfloat[Three Obj. Prefs.]{\includegraphics[width = 0.23\linewidth]{Figures/new_psl_predicted_pref_pref.pdf}} \hfill
\subfloat[DTLZ2]{\includegraphics[width = 0.23\linewidth]{Figures/new_psl_predicted_pref_dtlz2.pdf}}\hfill
\subfloat[Disk Brake]{\includegraphics[width = 0.23\linewidth]{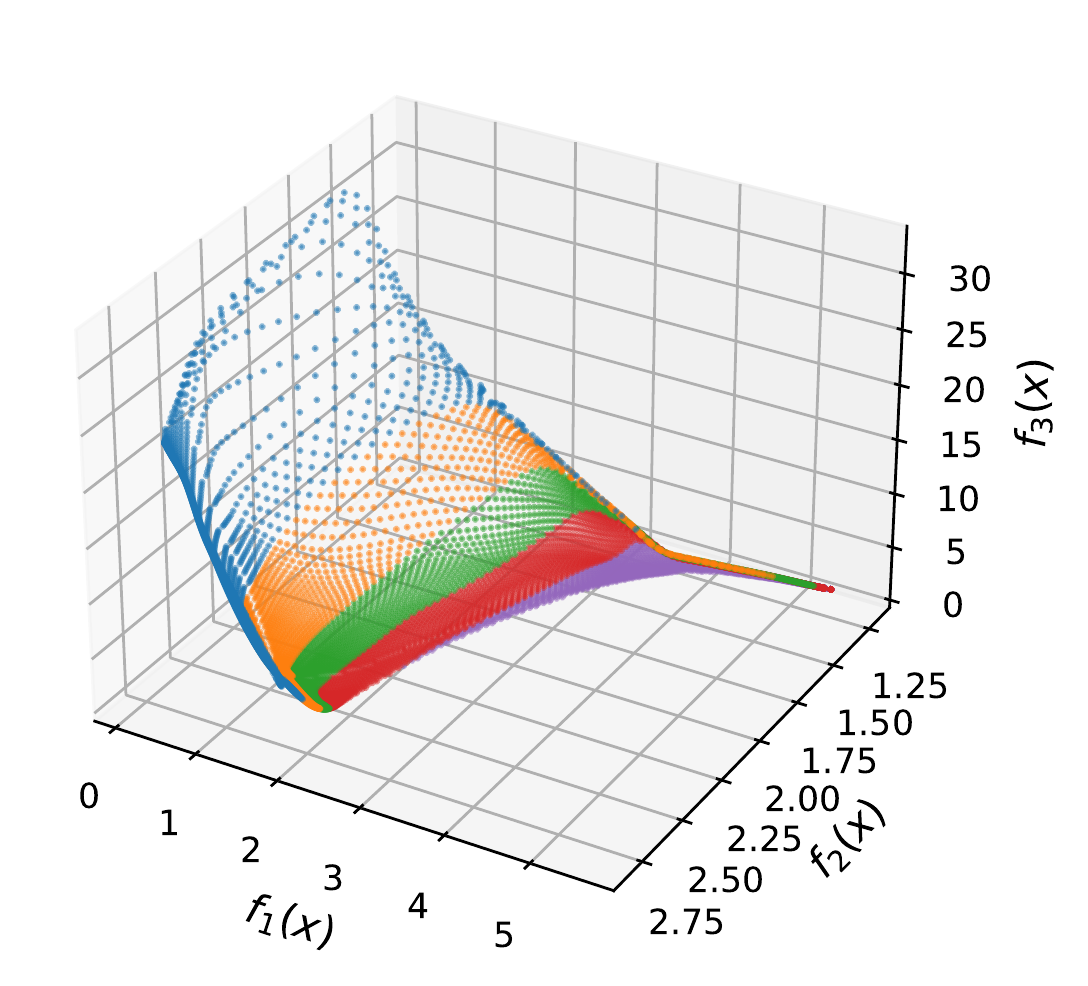}}\hfill
\subfloat[Rocket Injector]{\includegraphics[width = 0.23\linewidth]{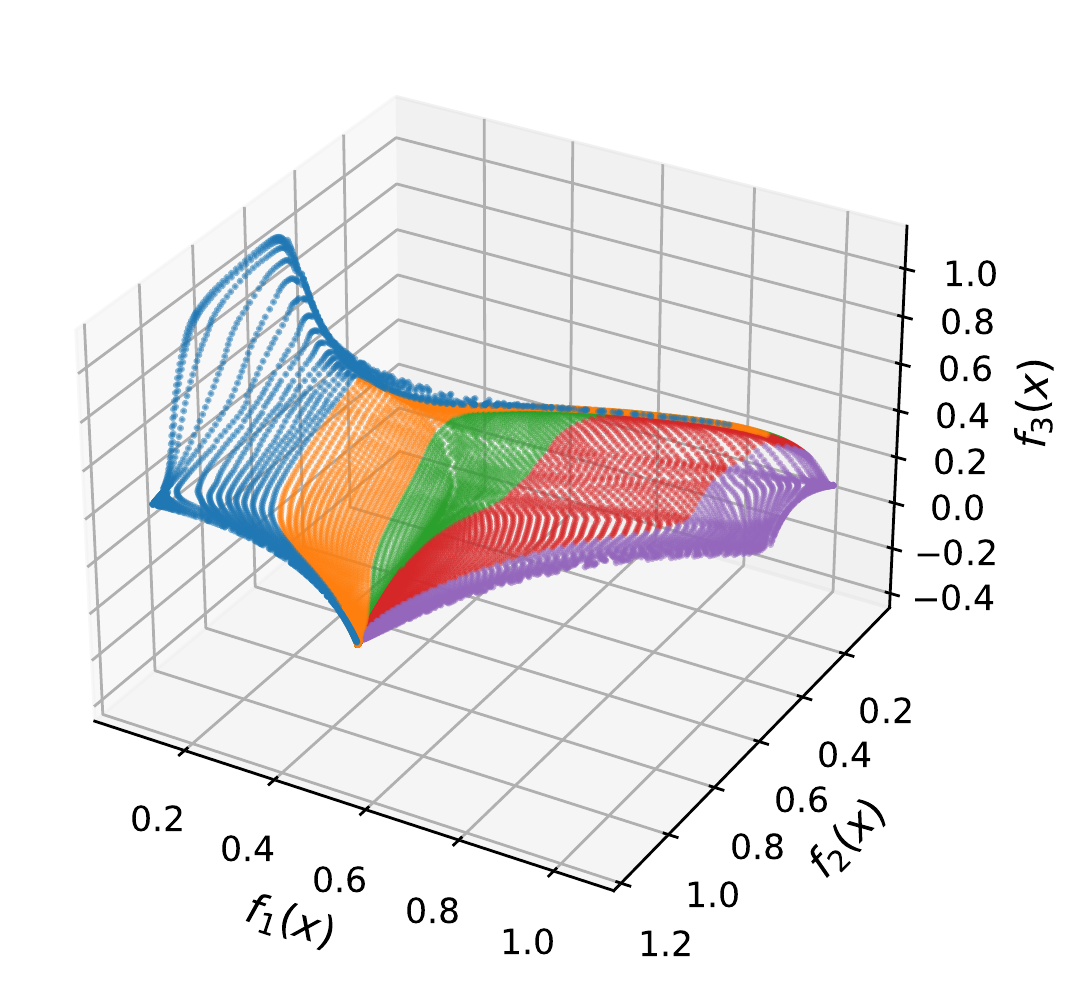}}
\caption{\textbf{Learned Pareto Set with Different Preferences:} Our proposed PSL model allows the decision-makers to flexibly explore the approximate Pareto front to obtain any trade-off solution. Suppose the decision-makers have different sets of preferred trade-offs, our model can successfully generate the corresponding parts of the Pareto fronts. The preference assignment is not fixed, and the decision-makers can adjust it and obtain the corresponding approximate solutions in real time.}
\label{supp_fig_psl_preference}
\end{figure}

Our proposed PSL method enables the decision-makers to easily adjust their preference to explore the whole approximate Pareto set for making decisions. This ability is not supported by other MOBO methods, which is important to flexibly incorporate the decision-maker's preference and prior knowledge into Bayesian optimization~\cite{garnett2022bayesian}.

The decision-making process for real-world applications could be subjective. We follow a similar way with other preference-based MOBO to show our method can successfully generate different trade-off solutions. In Figure~\ref{supp_fig_psl_preference}, suppose the decision-makers have different sets of trade-off preferences, our model can generate the corresponding parts of the Pareto fronts for different problems. We uniformly sample $1,000$ and $10,011$ preferences for the two-objective and three-objective problems respectively, and use the learned Pareto set models to generate the corresponding approximate Pareto solutions. We further divide the preferences into 5 exclusive parts, and label them with the corresponding solutions in different colors. As shown in Figure~\ref{supp_fig_psl_preference}, the corresponding connection from the preferences to the solutions is clear for all problems. 

By directly exploring the approximate Pareto front in an interactive manner, the decision-makers can observe and understand the connection between preferences and corresponding solutions in real-time. This ability could be beneficial for the decision-makers to further adjust and assign their most accurate preferences.

\clearpage 

\subsection{Standard deviation on the approximated Pareto set}
\label{subsec_supp_other_information}

\begin{figure*}[h]
\centering
\vspace{-0.15in}
\subfloat[Std for Obj1.]{\includegraphics[width = 0.33\linewidth]{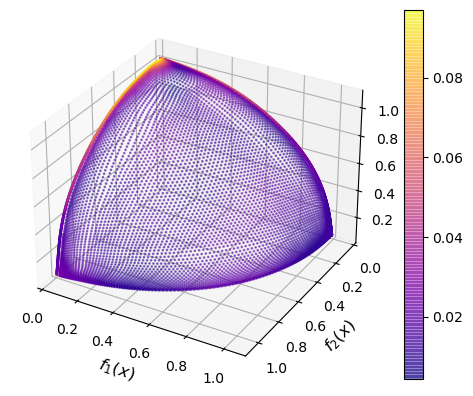}} \hfill
\subfloat[Std for Obj2.]{\includegraphics[width = 0.33\linewidth]{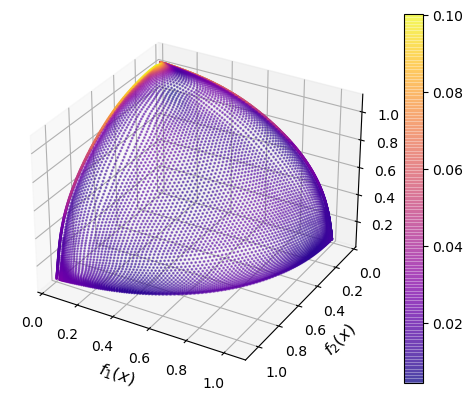}}\hfill
\subfloat[Std for Obj3.]{\includegraphics[width = 0.33\linewidth]{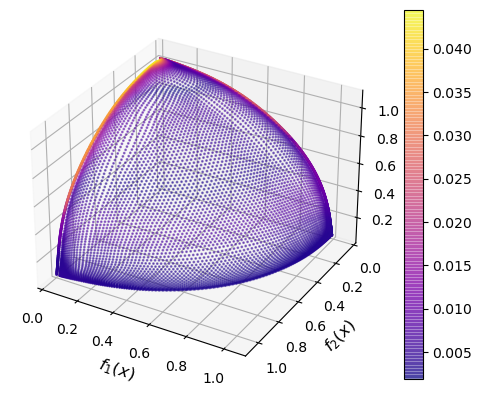}}\\ 
\subfloat[Log Std for Obj1.]{\includegraphics[width = 0.33\linewidth]{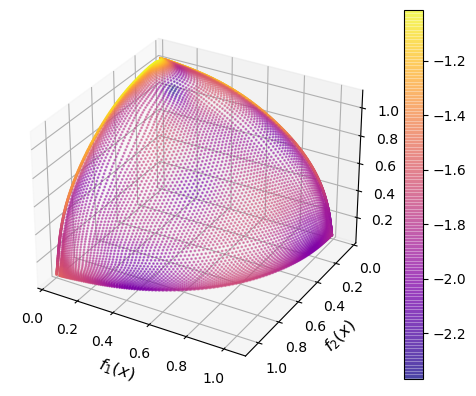}} \hfill
\subfloat[Log Std for Obj2.]{\includegraphics[width = 0.33\linewidth]{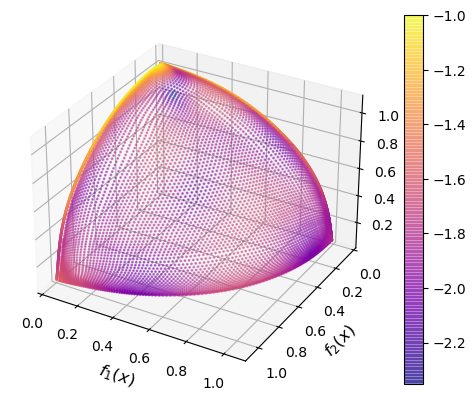}}\hfill
\subfloat[Log Std for Obj3.]{\includegraphics[width = 0.33\linewidth]{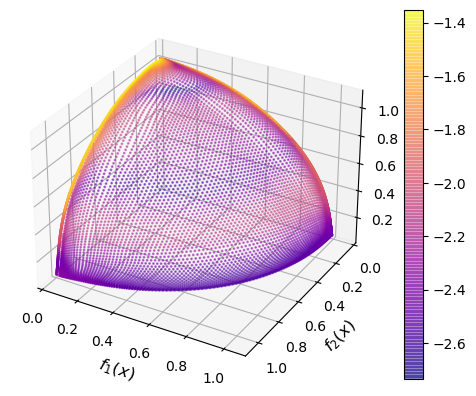}}
\caption{\textbf{Standard deviation and log standard deviation for DTLZ2:} The predicted standard deviations for different objectives on the approximated Pareto front.}
\label{supp_fig_uncertainty}
\end{figure*}

In addition to the predicted mean, our PSL method also provides the uncertainty information (e.g., predicted standard deviation) to support decision-making. Figure~\ref{supp_fig_uncertainty} shows the (log) predicted standard deviations for different objectives on the approximate Pareto front of DTLZ2. According to the results in Figure~\ref{supp_fig_uncertainty}(a)(b)(c), we can observe that the PSL model has low predicted standard deviations for almost all locations on the approximate Pareto front except the top corner and the top left boundary. From the log standard deviations in Figure~\ref{supp_fig_uncertainty}(d)(e)(f), we find that PSL has a higher overall uncertainty level for objective 1/2 than objective 3, where the uncertainty distributions are quite similar for the first two objectives.  
 
\clearpage

\subsection{Performance with different acquisition functions}
\label{subsec_supp_acq}

\begin{figure*}[ht]
\centering
\vspace{-0.15in}
\subfloat[VLMOP1]{\includegraphics[width = 0.33\linewidth]{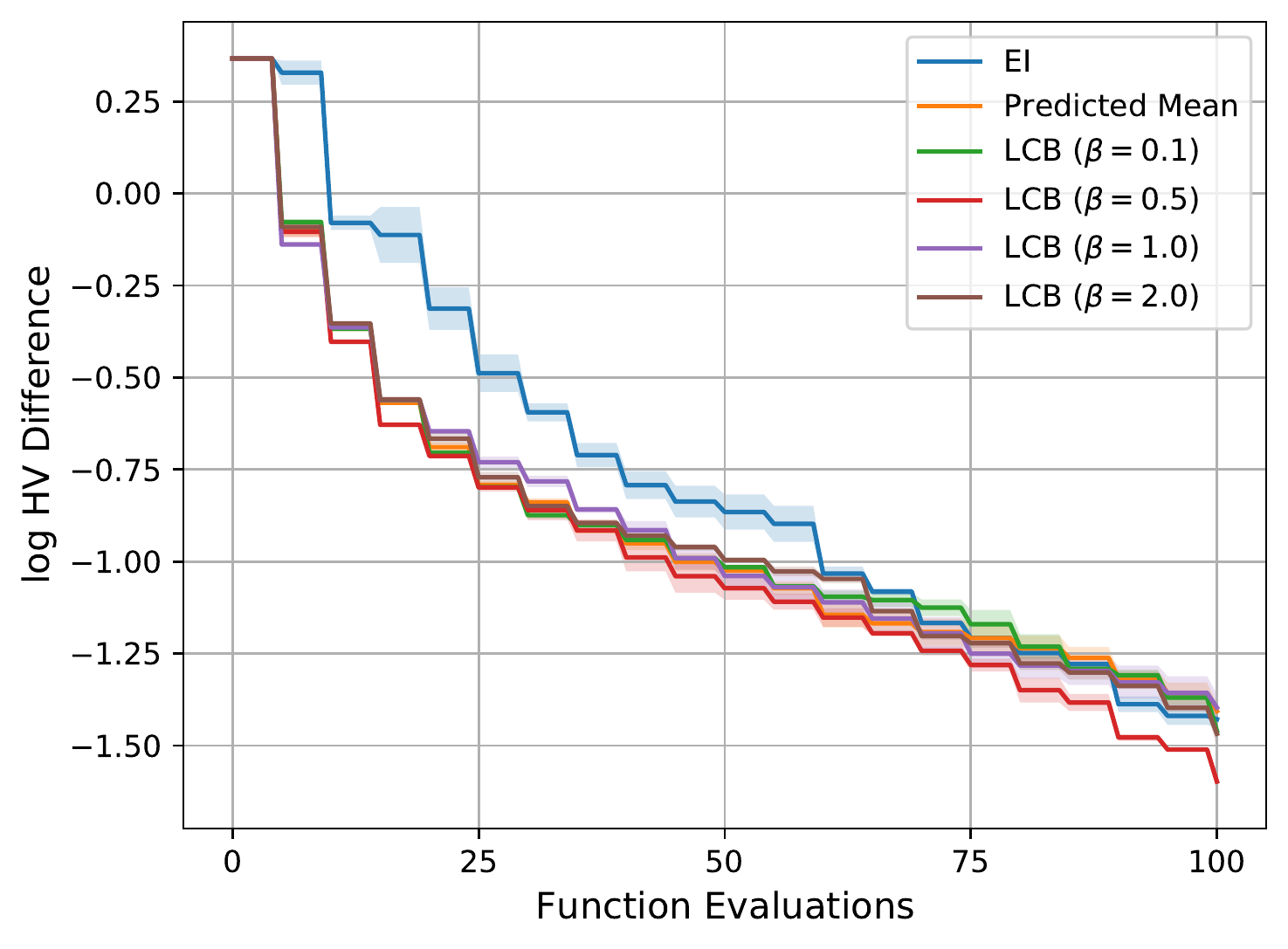}}\hfill
\subfloat[Four Bar Truss]{\includegraphics[width = 0.33\linewidth]{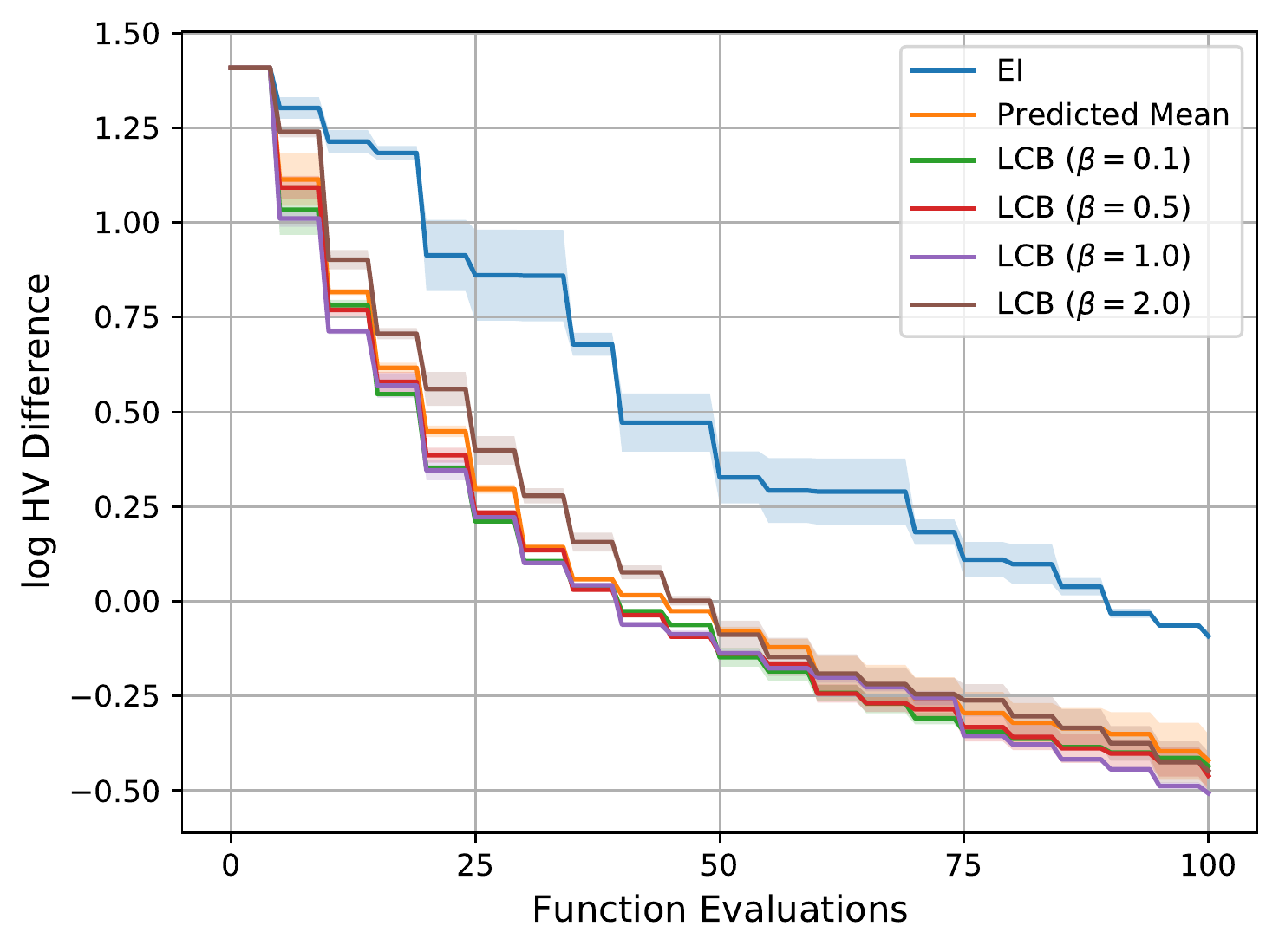}}\hfill
\subfloat[Pressure Vessel]{\includegraphics[width = 0.33\linewidth]{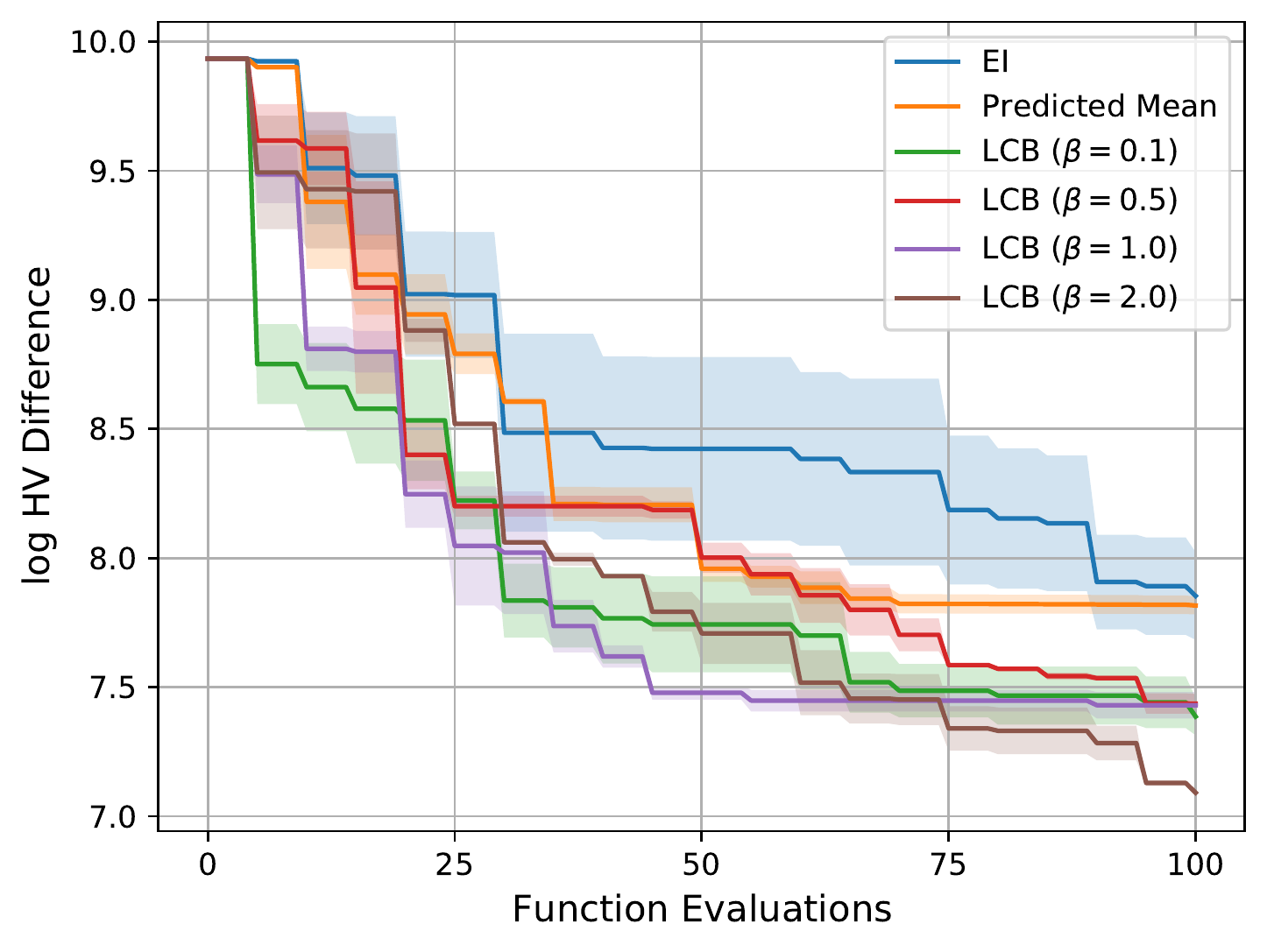}}
\caption{The log hypervolume difference w.r.t. the number of expensive evaluation for PSL with different acquisition functions.}
\label{supp_fig_acq}
\end{figure*}

For the proposed PSL method, we can use different acquisition functions as the surrogate value for the Pareto set learning approach. Here, we compare the performance with $6$ different acquisition functions/settings, namely, the Expected Improvement (EI), Predicted Mean, and the Lower Confidence Bound (LCB) with $\beta = 0.1, 0.5, 1.0$ and $2.0$ on four benchmark and real-world application problems.  

According to the results shown in Figure~\ref{supp_fig_acq}, PSL with all $6$ acquisition functions has reasonably good performance. Although PSL with EI performs slightly worse than the other acquisitions, it can still outperform many other MOBO algorithms with the results shown in Figure~\ref{fig_hv_trend} of the main paper. These results confirm that learning the approximate Pareto set during optimization could be beneficial for MOBO.

Among the other acquisition functions and settings, no single choice can achieve the best performance for all problems. The predicted mean acquisition can be seen as LCB with $\beta = 0$, which is a pure exploitation strategy without considering the uncertainty (e.g., predicted variance) for exploration. It might be surprising that this greedy acquisition can still have a pretty good performance. Indeed, this greedy and purely exploitative approach has been recently studied
and shown to have promising performance for single-objective
BO~\cite{rehbach2020expected,de2021greed}. DGEMO~\cite{lukovic2020diversity} also uses the predicted mean as an acquisition function for MOBO. \citet{garnett2022bayesian} briefly discusses this pure exploitative strategy at the end of Section 7.10 (posterior mean acquisition function and lookahead). One limitation of this greedy approach is that it might be stuck at a non-optimal location with no well-fitted surrogate model due to overexploitation~\cite{garnett2022bayesian}. There are many acquisition policies for Bayesian optimization, but finding the most suitable one for a given problem is still an open question~\cite{garnett2022bayesian}, which is also the case for our PSL method.

\clearpage

\subsection{Results on problems with larger decision space}
\label{subsec_supp_new_results}

\begin{figure*}[ht]
\captionsetup[subfigure]{font=scriptsize,labelfont=scriptsize}
\centering
\subfloat[F1]{\includegraphics[width = 0.33\linewidth]{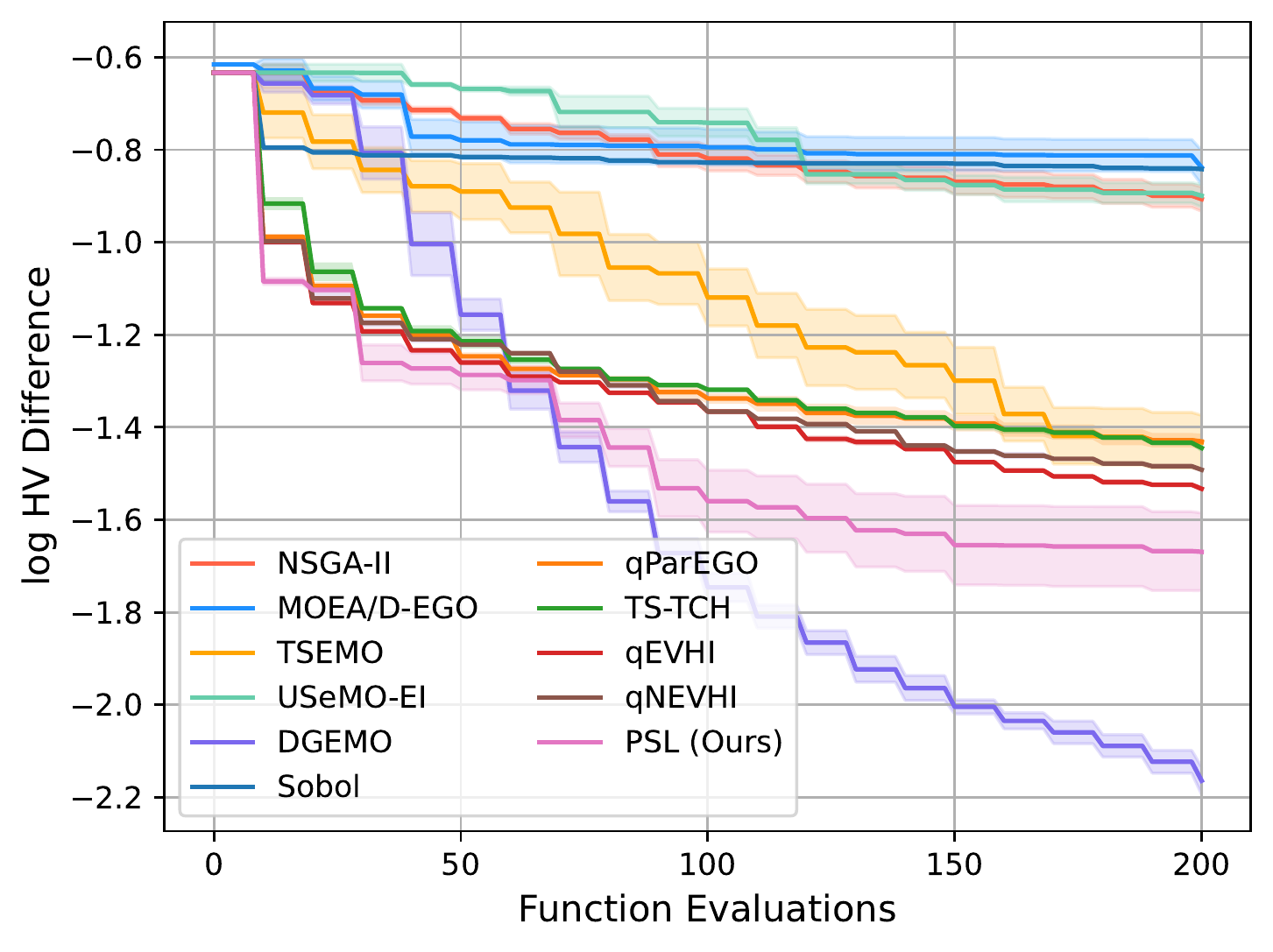}}\hfill
\subfloat[F2]{\includegraphics[width = 0.33\linewidth]{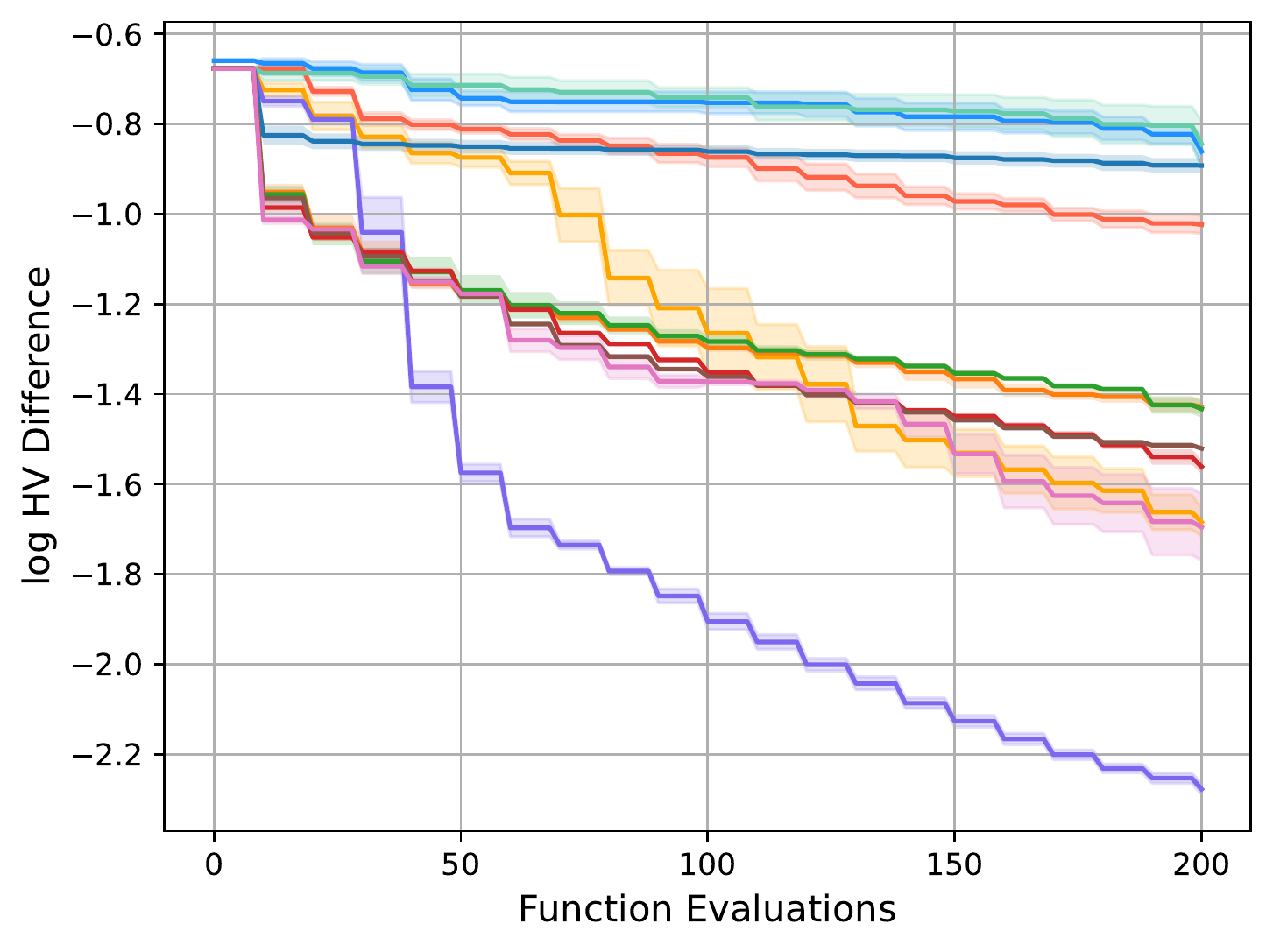}}\hfill
\subfloat[F3]{\includegraphics[width = 0.33\linewidth]{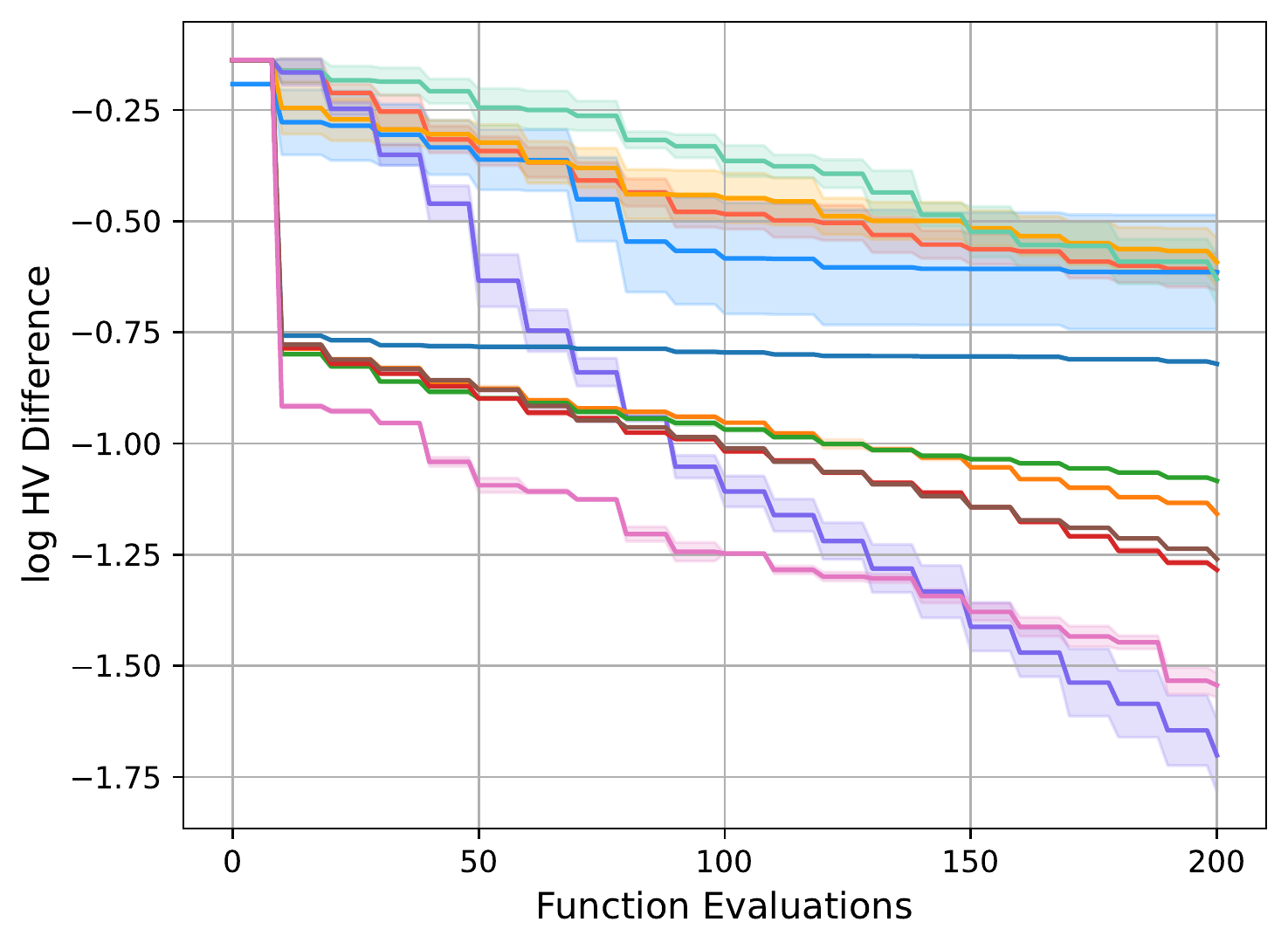}}\\
\subfloat[F4]{\includegraphics[width = 0.33\linewidth]{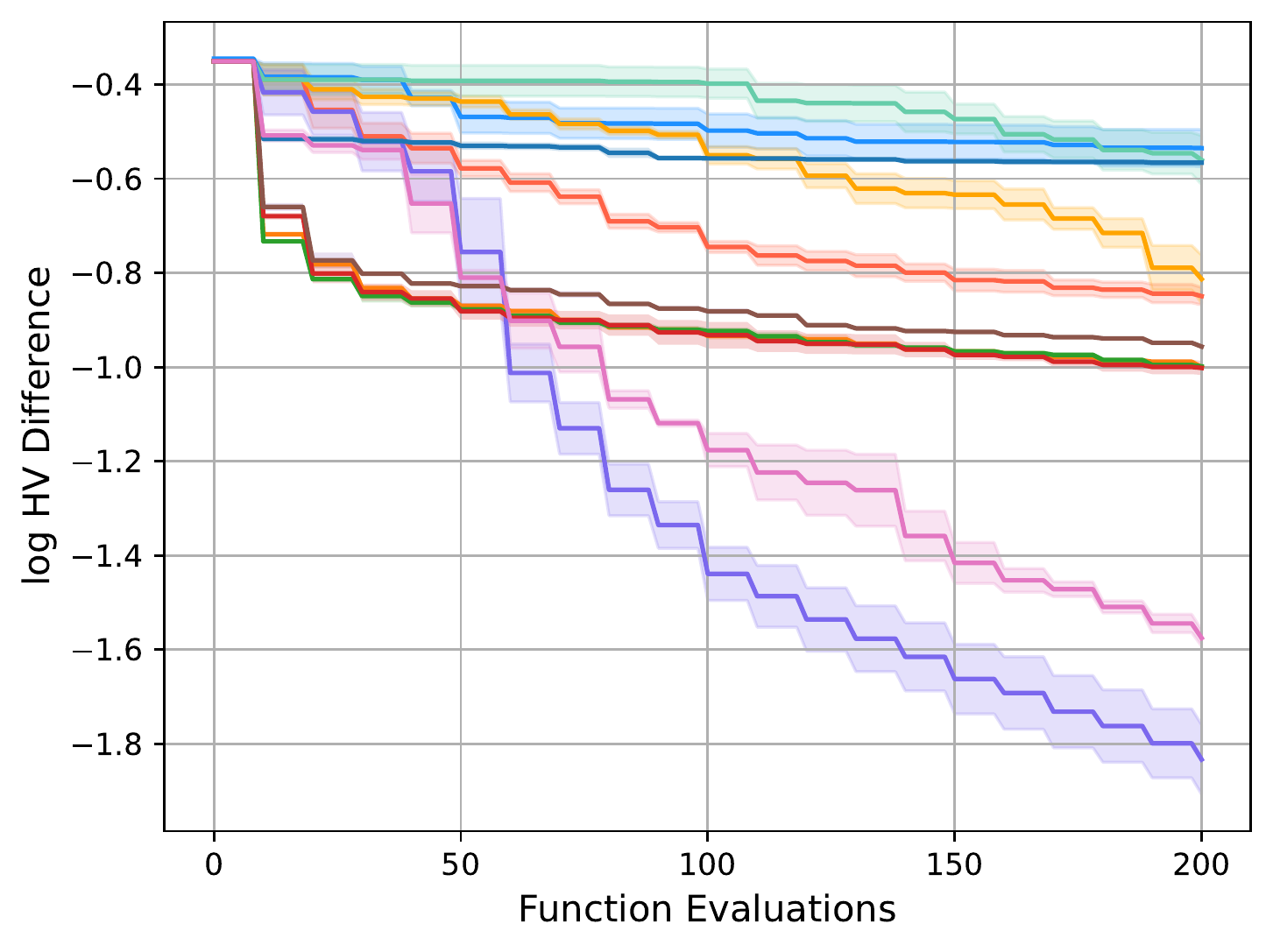}}\hfill
\subfloat[F5]{\includegraphics[width = 0.33\linewidth]{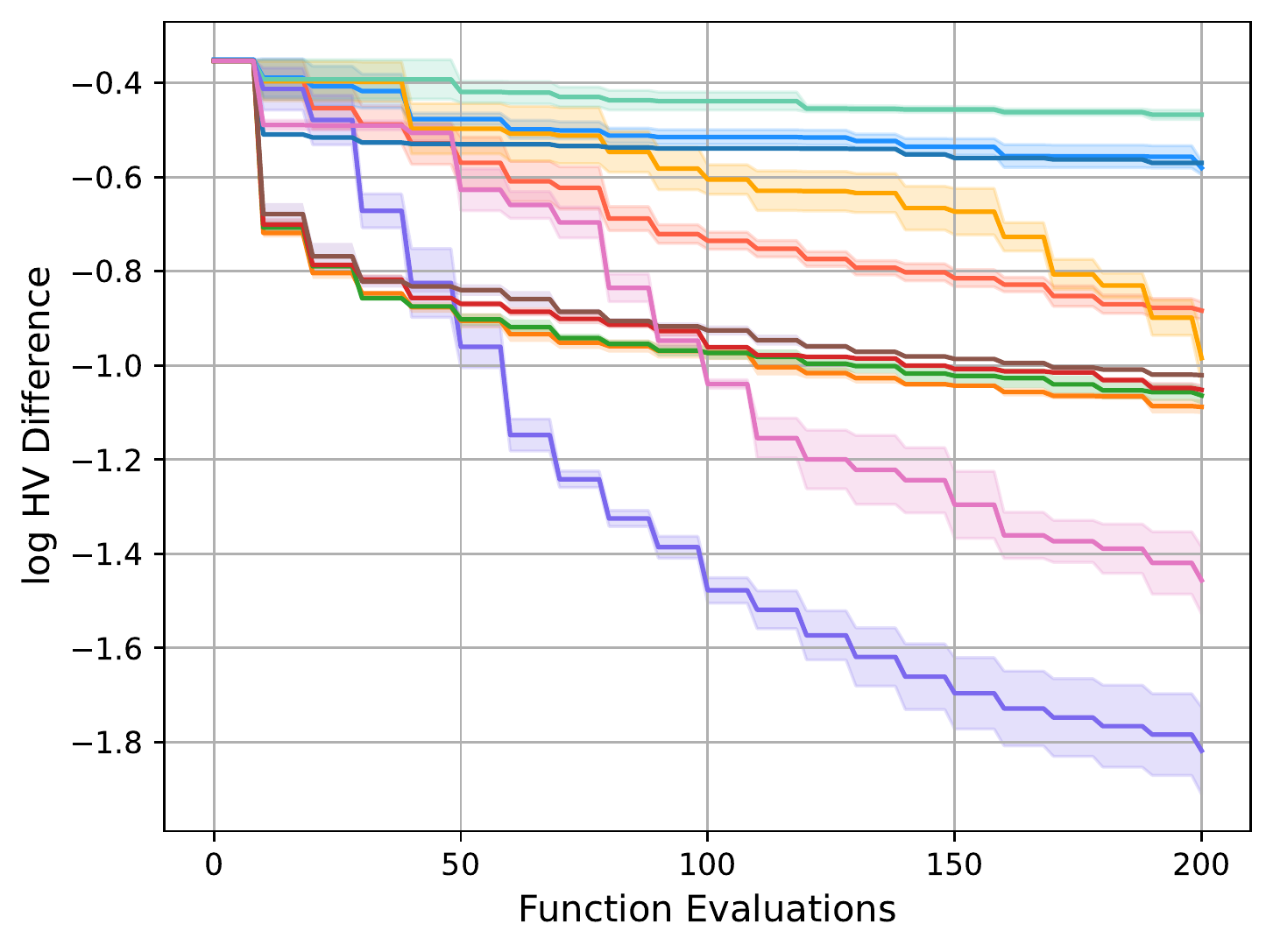}}\hfill
\subfloat[F6]{\includegraphics[width = 0.33\linewidth]{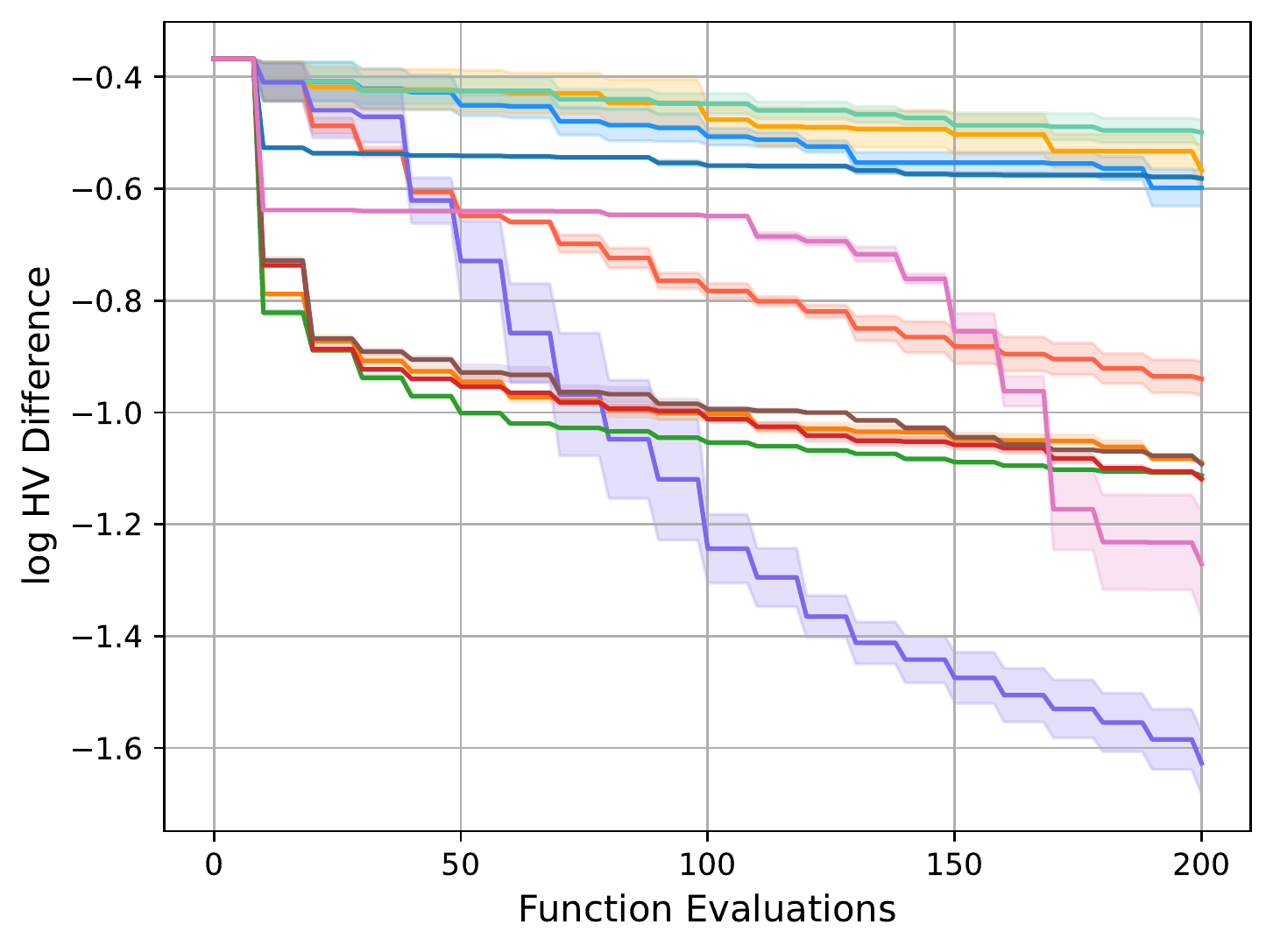}}
\caption{The log hypervolume difference w.r.t. the number of expensive evaluation for different algorithms for F1-F6 problems with $20$ dimensional search space. The solid line is the mean value averaged over $10$ independent runs for each algorithm, and the shaded region is the standard deviation around the mean value. \textbf{The label of each algorithm can be found in subfig(a).}}
\label{fig_hv_trend_new_problems}
\end{figure*}

We conduct experiments on more challenging problems proposed in Appendix~\ref{subsec_supp_benchmark} with a 20-dimensional decision space to further validate our proposed PSL method. The Pareto sets of these problems are all different complicated space curves in the search space. We set the batch size to 10, and run all algorithms with 20 batched evaluations. In other words, the evaluation budget is $200$ for all problems. The experimental results in Figure~\ref{fig_hv_trend_new_problems} show that our proposed PSL method still has a promising performance on these problems.

According to the experimental results, the DGEMO~\cite{lukovic2020diversity} performs the best on these problems with larger decision space. The DGEMO algorithm has a well-designed local search strategy, which fully leverages the first and second order derivatives of the Gaussian process surrogate model to enrich its large candidate set for selection. In addition, it also explicitly encourages exploring diverse solutions during the optimization process. In contrast, for problems with larger decision space, our proposed PSL method could learn an inaccurate Pareto set, which could mislead and slow down the search process. In this work, we focus on the main idea of learning the Pareto set for MOBO problems, and do not aggressively include other powerful methods into our proposed algorithm. Its performance might be further improved by considering the second-order derivatives of GPs and diversity-guided batch selection as in DGEMO~\citep{lukovic2020diversity}.

\subsection{Performance with Different Batch Sizes}
\label{subsec_supp_batchsize}

\begin{figure*}[!ht]
\centering
\vspace{-0.15in}
\subfloat[F1]{\includegraphics[width = 0.33\linewidth]{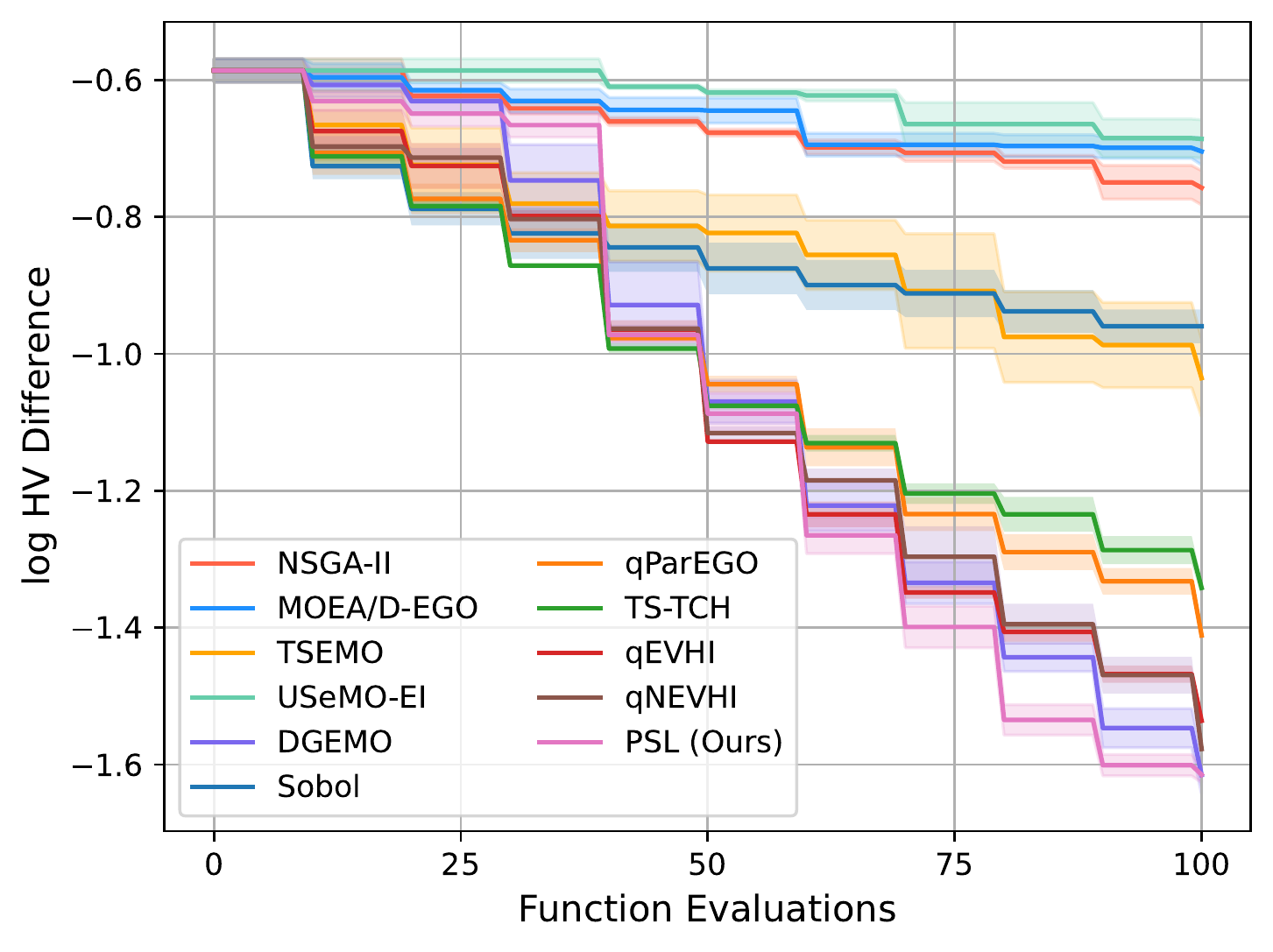}}\hfill
\subfloat[F2]{\includegraphics[width = 0.33\linewidth]{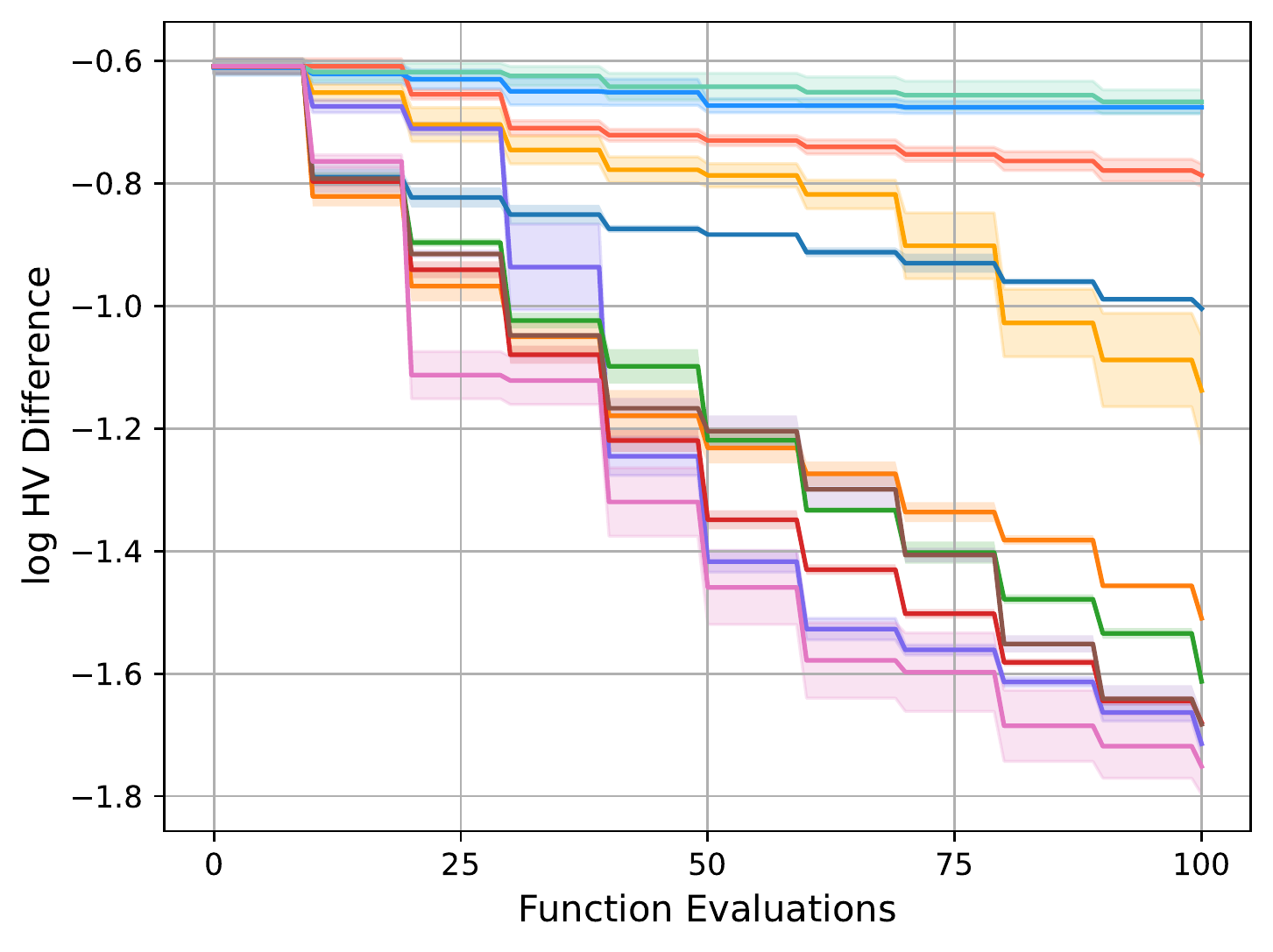}}\hfill
\subfloat[F3]{\includegraphics[width = 0.33\linewidth]{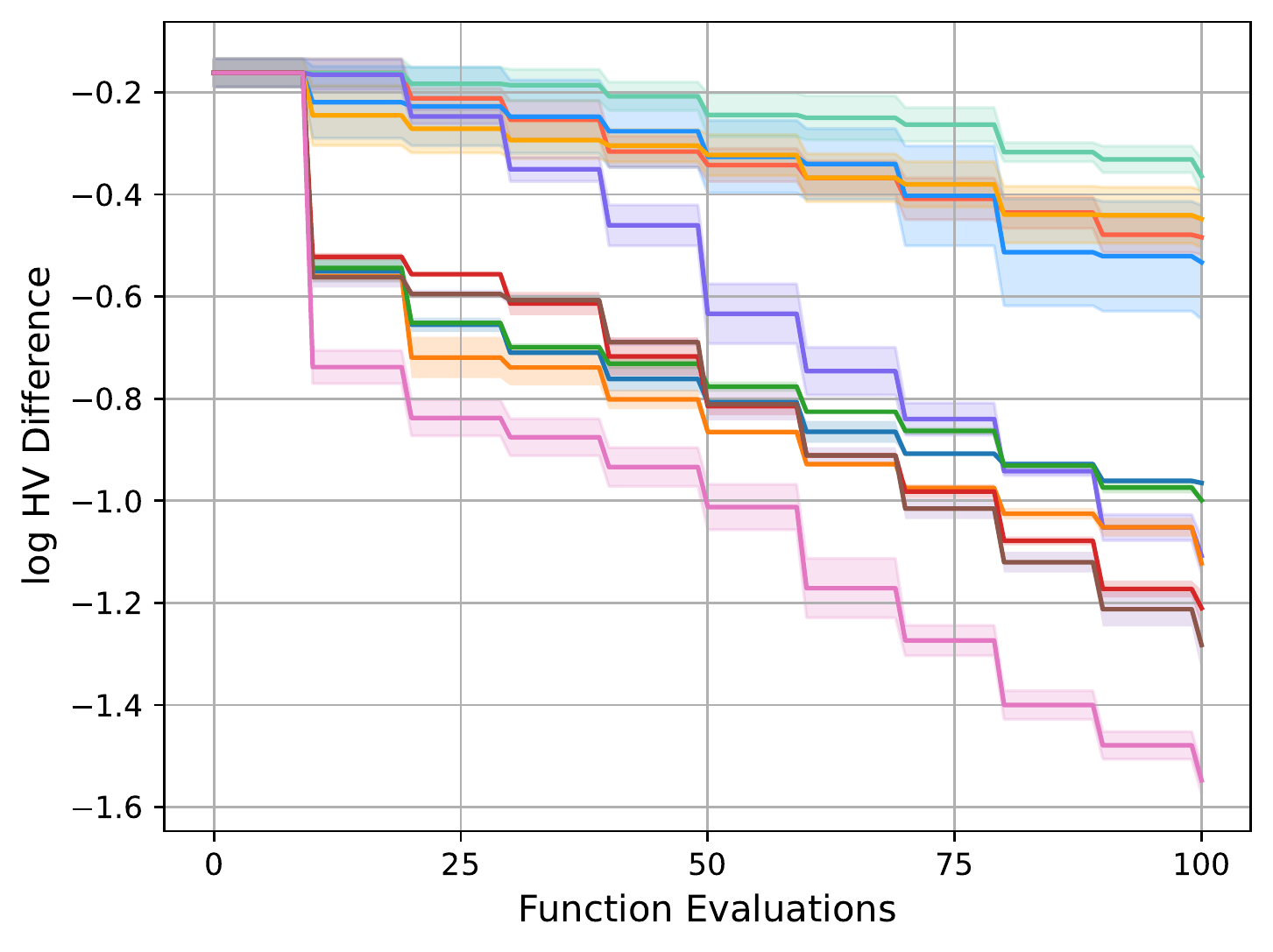}}\\ 
\subfloat[F4]{\includegraphics[width = 0.33\linewidth]{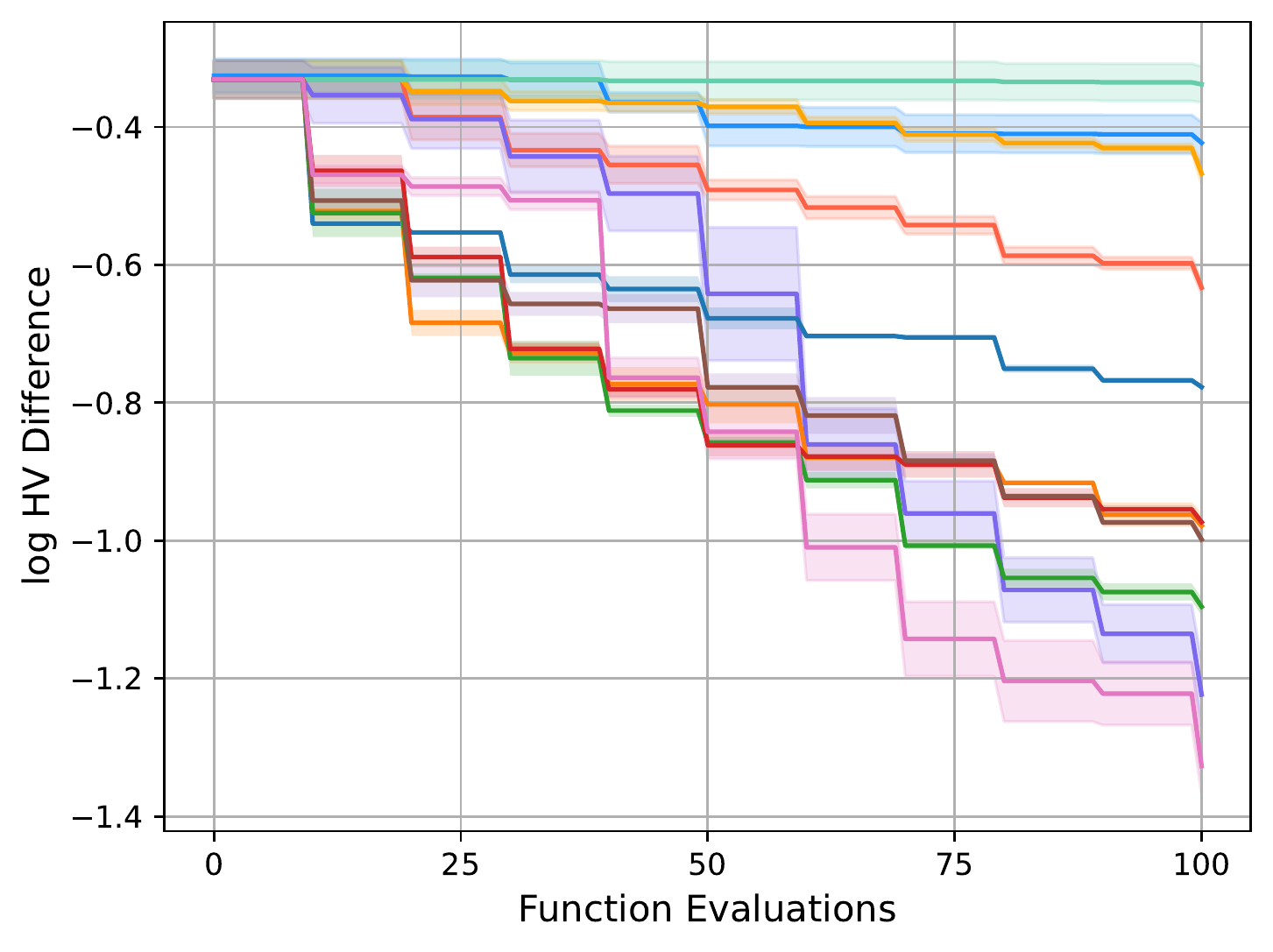}}\hfill
\subfloat[F5]{\includegraphics[width = 0.33\linewidth]{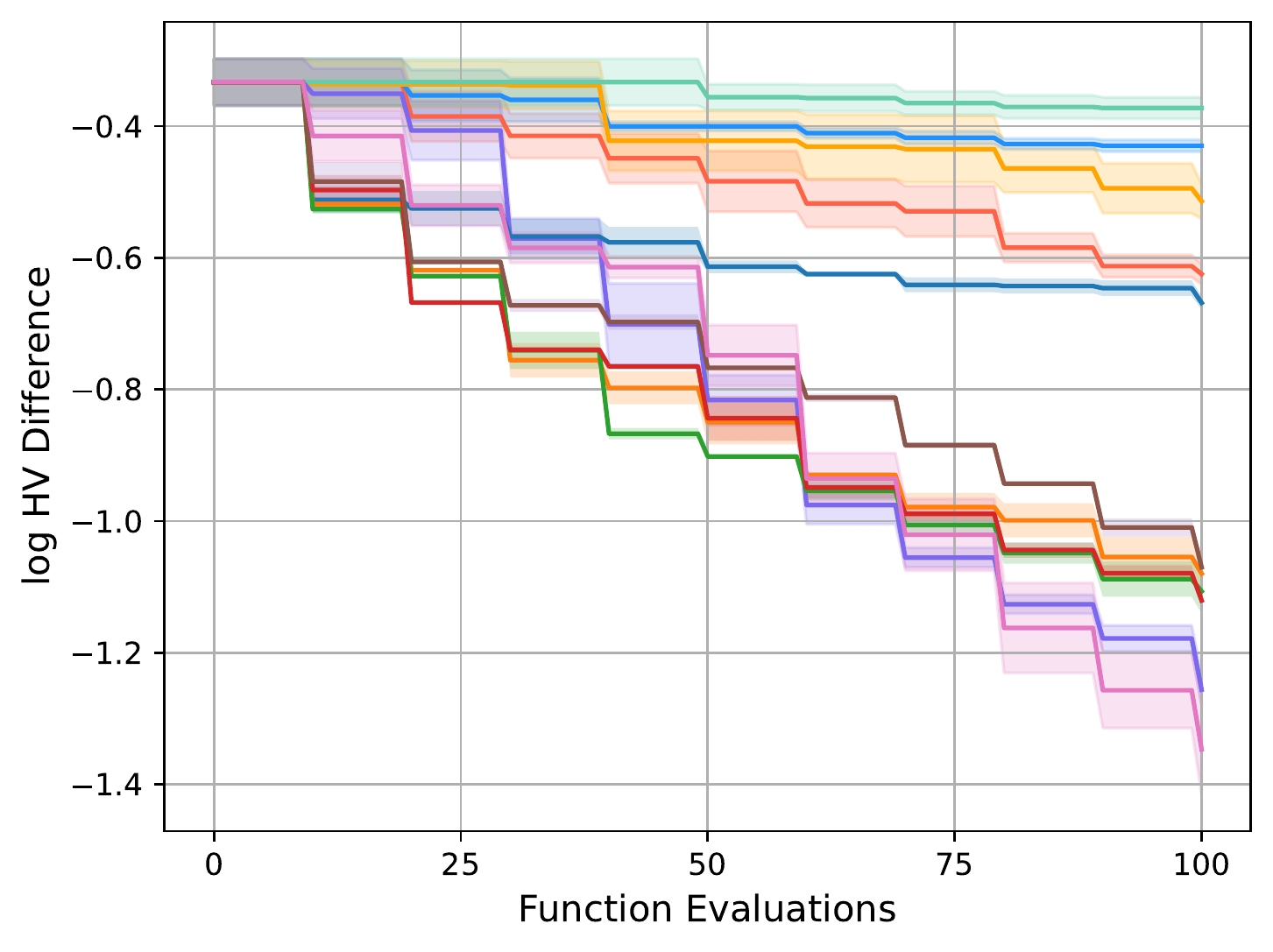}}\hfill
\subfloat[F6]{\includegraphics[width = 0.33\linewidth]{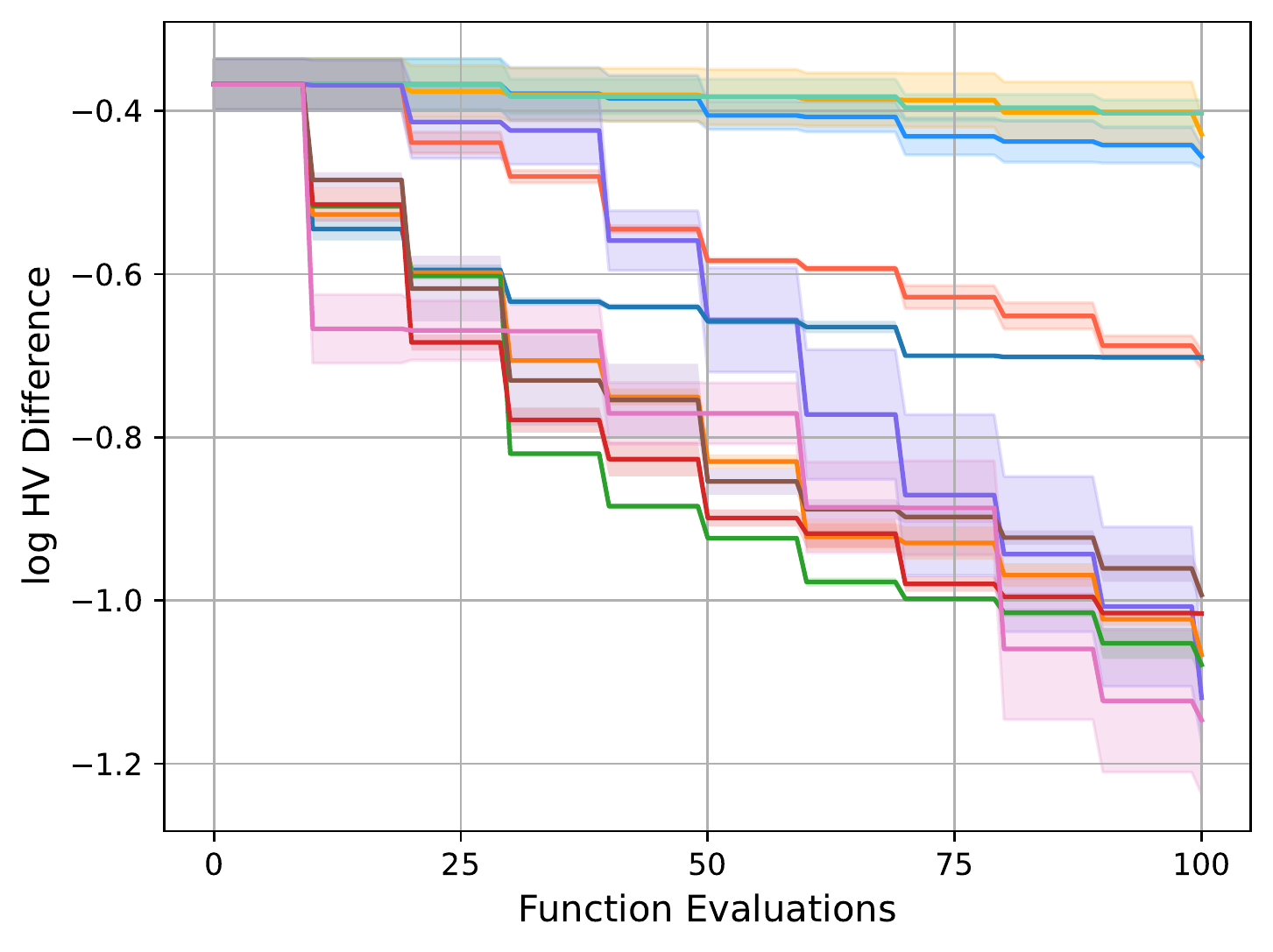}}
\caption{\textbf{Experimental results with Batch Size 10} on the log hypervolume difference w.r.t. the number of expensive evaluation for different algorithms. \textbf{The label of each algorithm can be found in subfig(a).}}
\label{supp_fig_hv_trend_batch10}
\end{figure*}

\begin{figure*}[!ht]
\centering
\vspace{-0.15in}
\subfloat[F1]{\includegraphics[width = 0.33\linewidth]{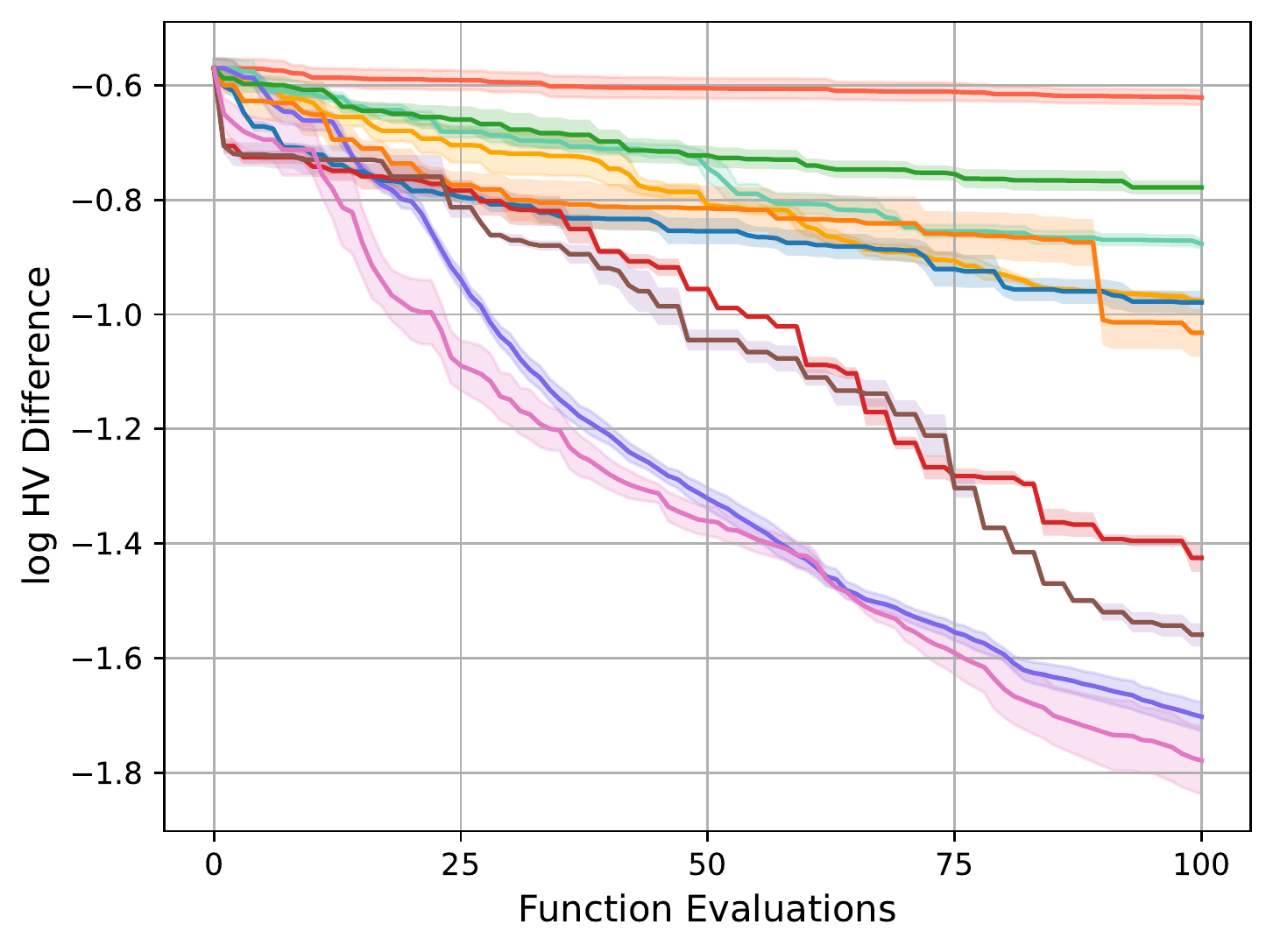}}\hfill
\subfloat[F2]{\includegraphics[width = 0.33\linewidth]{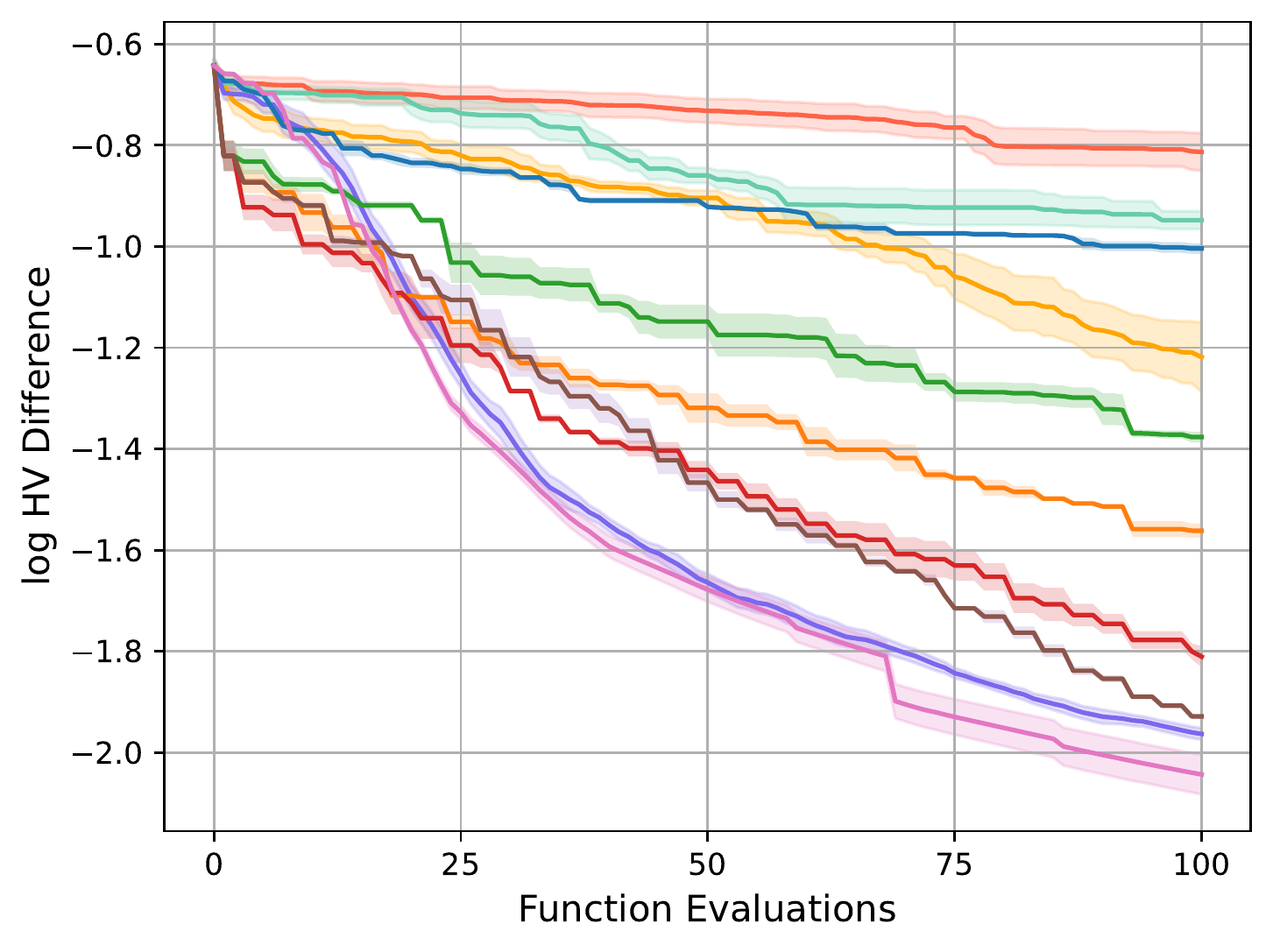}}\hfill
\subfloat[F3]{\includegraphics[width = 0.33\linewidth]{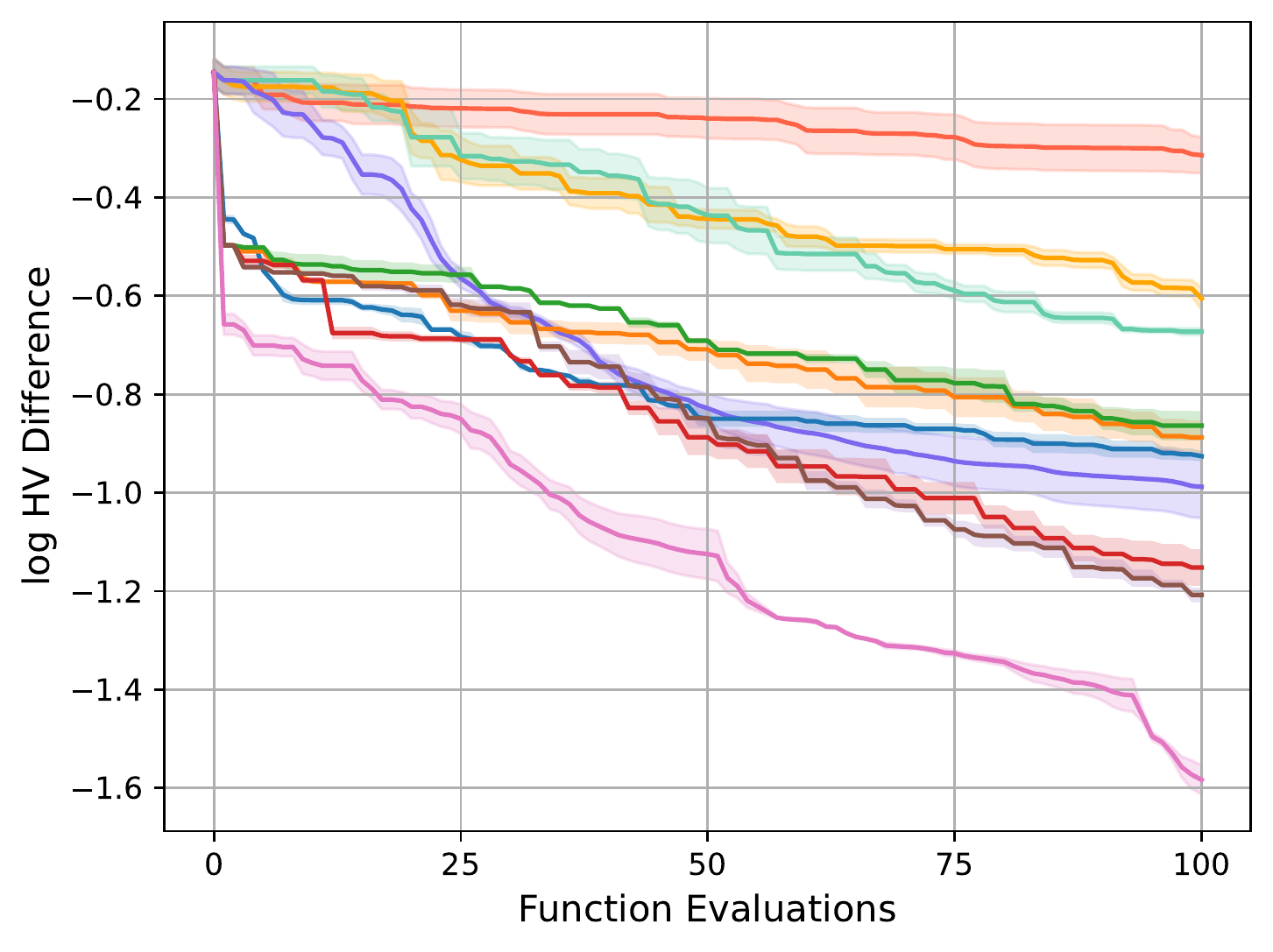}}\\ 
\subfloat[F4]{\includegraphics[width = 0.33\linewidth]{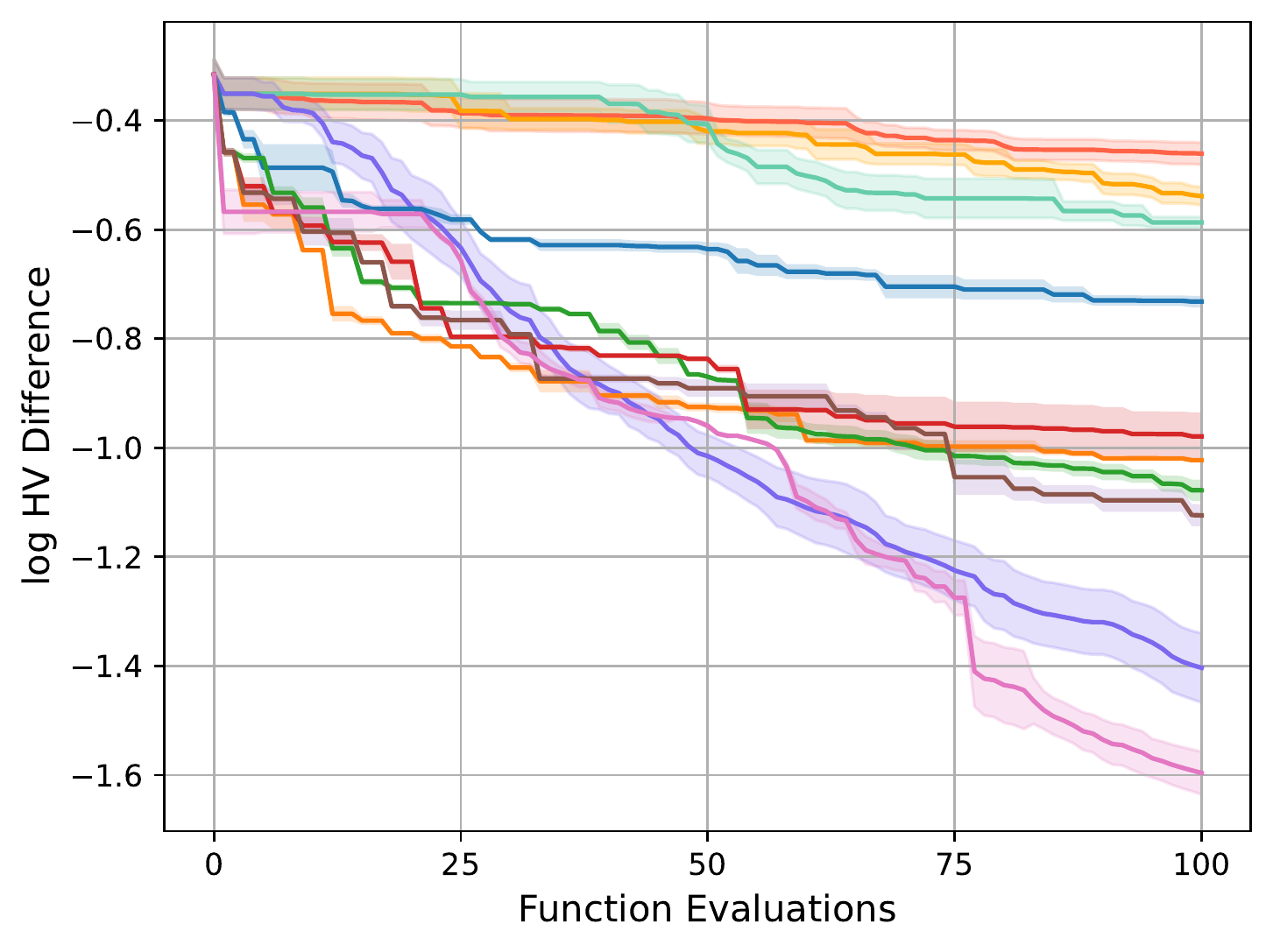}}\hfill
\subfloat[F5]{\includegraphics[width = 0.33\linewidth]{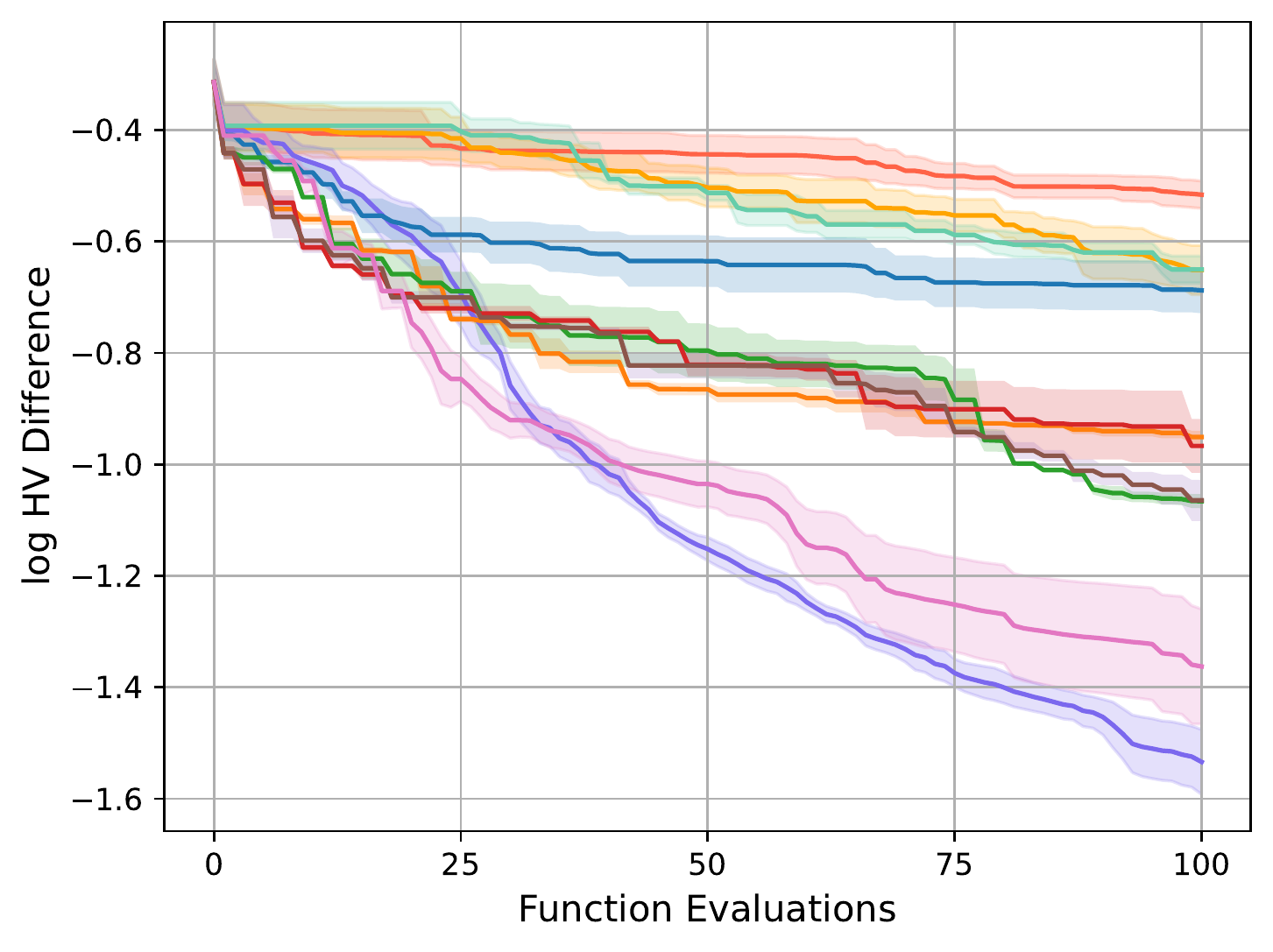}}\hfill
\subfloat[F6]{\includegraphics[width = 0.33\linewidth]{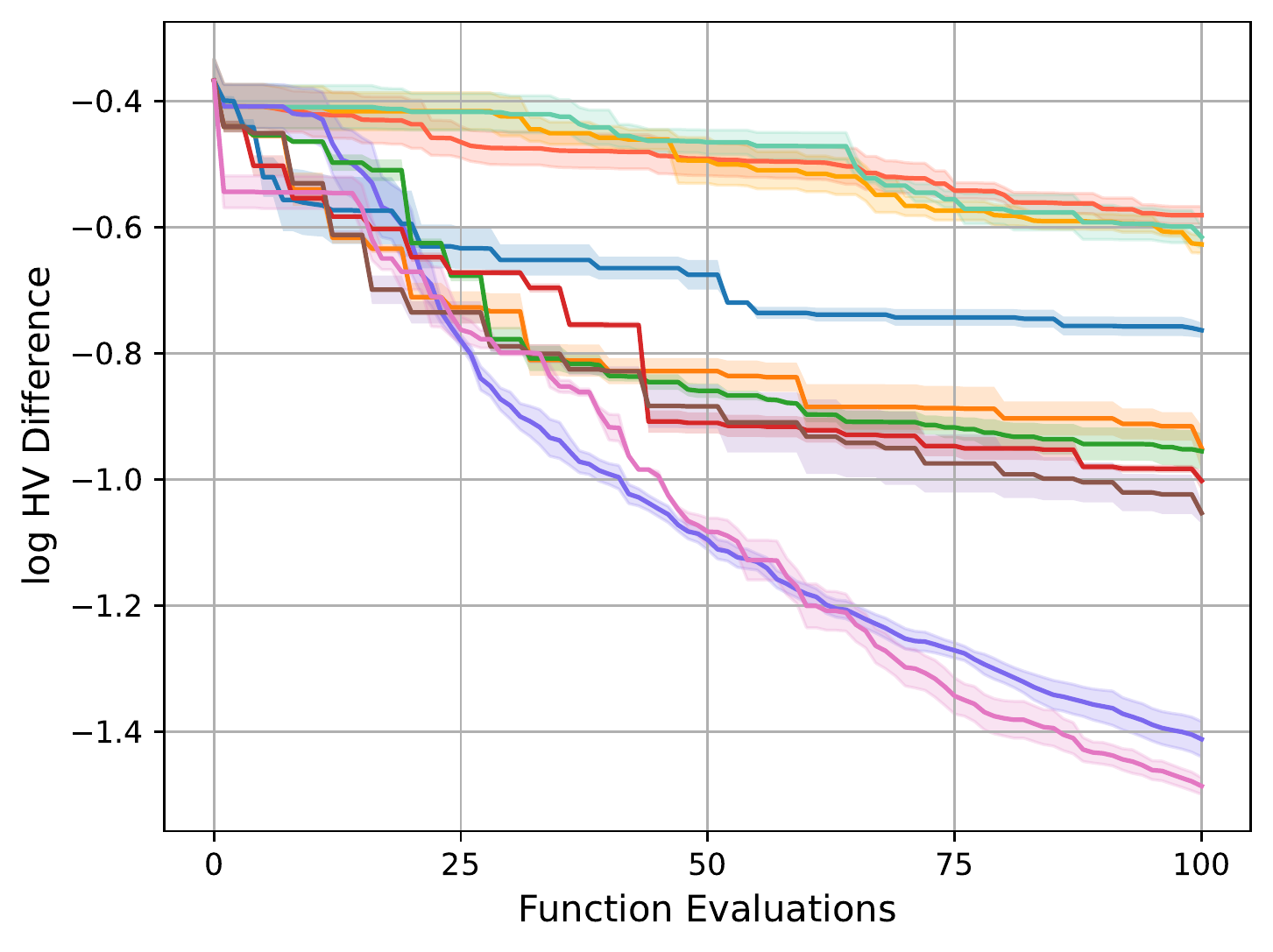}}
\caption{\textbf{Experimental results with Batch Size 1} on the log hypervolume difference w.r.t. the number of expensive evaluation for different algorithms.}
\label{supp_fig_hv_trend_batch1}
\end{figure*}

We conduct additional experiments with different batch size ($B = 10$) as well as sequential evaluation ($B = 1$) for all problems. The total number of evaluations is still $10 + 100 = 110$ for all experiments. In other words, these settings have $10$ and $100$ batched evaluations, respectively. The experimental results are shown in Figure~\ref{supp_fig_hv_trend_batch10} and Figure~\ref{supp_fig_hv_trend_batch1}. According to the results, our proposed PSL method has a better or comparable performance with other MOBO methods in most experiments. It also consistently outperforms the model-free scalarization methods with a large performance gap. These promising performances, along with the efficient run time, confirm that it is worth building the Pareto set model for expensive multi-objective optimization problems. 

\clearpage

\section{Licenses}
\label{sec_supp_licenses}

Table~\ref{table_licence} lists the licenses for the codes and problem suite we used in this work.

\begin{table}[h]
\small
\centering
\caption{List of licenses for the codes and problem suite we used in this work.}
\label{table_licence}
\tabcolsep=0.12cm
\begin{tabular}{llll}
\toprule
Resource                                                 & Type          & Link                                      & License            \\ \midrule
PSL-MOBO (Ours)                                          & Code          & https://github.com/Xi-L/PSL-MOBO          & MIT License        \\
Botorch\citep{balandat2020botorch} & Code          & https://botorch.org/                      & MIT License        \\
DEGMO\citep{lukovic2020diversity} & Code          & https://github.com/yunshengtian/DGEMO     & MIT License        \\
pymoo\citep{pymoo}                & Code          & https://pymoo.org/                        & Apache License 2.0 \\
reproblems\citep{tanabe2020easy}  & Problem Suite & https://ryojitanabe.github.io/reproblems/ & None               \\ \bottomrule
\end{tabular}
\end{table}

\end{document}